\documentclass[dvipsnames]{article}

\usepackage[final]{lychee}
\usepackage[utf8]{inputenc} % allow utf-8 input
\usepackage[T1]{fontenc}    % use 8-bit T1 fonts
\usepackage{url}            % simple URL typesetting
\usepackage{booktabs}       % professional-quality tables
\usepackage{amsfonts}       % blackboard math symbols
\usepackage{nicefrac}       % compact symbols for 1/2, etc.
\usepackage{booktabs} 
\usepackage{multirow}
\usepackage{wrapfig}
\usepackage{amssymb}
\usepackage{epigraph}
\usepackage{makecell}
\usepackage{listings}
\usepackage{subcaption}
\usepackage[labelfont=bf]{caption} % For bold caption labels

\usepackage[ruled,vlined]{algorithm2e}
\usepackage{color, soul}
\usepackage{microtype}      % microtypography
\usepackage{xcolor}         % colors
\usepackage{geometry}
\usepackage{times}
\usepackage{ragged2e}
\usepackage{amsmath}
\usepackage{bm}
\usepackage{latexsym}
\usepackage{graphicx}
\usepackage{float}
\usepackage{longtable}
\usepackage{booktabs} 
\usepackage{multirow}
\usepackage{amssymb}
\usepackage{longtable}
\usepackage{tabu}
\usepackage{bbding}
\usepackage{awesomebox}
\usepackage{bbding}
\usepackage[most,breakable]{tcolorbox}
\usepackage{etoolbox}
\usepackage{fancyhdr}
\usepackage{lipsum}
\usepackage{CJKutf8}
\usepackage{pdfpages}
\usepackage{multicol}
\usepackage{pgffor} 
\usepackage{fontawesome5}
\usepackage{nicematrix} 
\usepackage{paralist}
\usepackage[edges]{forest}
\usepackage{siunitx}
\usepackage{tikz-dependency}
\usepackage{tikz}
\usepackage{bbm}
\usepackage{amssymb}
\usepackage[english]{babel}  % 假设你是英文内容
\usepackage{hyphenat}
\usepackage{siunitx}

\newcommand{\Rmnum}[1]{\expandafter\@slowromancap\romannumeral #1@}

\usepackage{caption} 
\usepackage{chngcntr}
\usepackage[colorlinks, linkcolor=RoyalBlue, anchorcolor=BrickRed, citecolor=RoyalBlue, urlcolor=RoyalBlue]{hyperref}

\usepackage{cleveref}
\crefname{section}{§}{§§}
\Crefname{section}{§}{§§}

\newcommand\refsec[1]{Section~\hyperref[sec:#1]{\ref{sec:#1}}}
\newcommand\refsecs[2]{\hyperref[sec:#1]{§\ref{sec:#1}:~\textsc{#1}}, \hyperref[sec:#2]{§\ref{sec:#2}:~\textsc{#2}}}

\usepackage[tikz]{bclogo}
\definecolor{msftBlue}{RGB}{0,164,239}
\definecolor{msftGreen}{RGB}{127,186,0}
\definecolor{msftYello}{RGB}{255,185,0}
\definecolor{mypurple}{RGB}{138,43,226} 
\definecolor{msftBlack}{RGB}{0,0,0}

\newtcolorbox{myboxnote}[1][]{
  breakable,
  title=#1,
  colback=cyan!0,
  colbacktitle=cyan!0,
  coltitle=black,
  fonttitle=\bfseries,
  bottomrule=0pt,
  toprule=0pt,
  leftrule=1.5pt,
  rightrule=1.5pt,
  titlerule=0pt,
  arc=0pt,
  outer arc=0pt,
  colframe=lightgray,
}
\usepackage{mdframed}

\definecolor{academicblue}{RGB}{54, 95, 145}

\tcbset{
  iclrtakeawaybox/.style={
    width=\linewidth,
    colback=gray!5,                 % 淡灰背景
    colframe=gray!60,              % 灰色边框
    colbacktitle=gray!30,            % 白色标题背景
    coltitle=black,                % 黑色标题字体
    fonttitle=\bfseries,     % 加粗大号标题
    boxrule=0.5pt,                 % 边框粗细
    arc=2pt,                       % 轻微圆角
    sharp corners=south,           % 底部保持直角
    attach boxed title to top left={yshift=-0.1in,xshift=0.15in},
    boxed title style={boxrule=0pt, colframe=white},
    enhanced,
    drop shadow southeast           % 可选阴影（可注释掉）
  }
}
\newtcolorbox{TakeawayBox}[2][]{iclrtakeawaybox,title=#2,#1}

\newenvironment{itemsize*}%
 {\leftmargini=20pt\begin{itemize}%
  \setlength{\itemsep}{3pt}%
  \setlength{\parskip}{0pt}%
  }%
 {\end{itemize}}
\newenvironment{enumerate*}%
 {\begin{enumerate}%
  \setlength{\itemsep}{0pt}%
  \setlength{\parskip}{0pt}}%
 {\end{enumerate}}

\usepackage{fancyhdr} % to change header and footers
\usepackage{blindtext} % to quickly get a full document
\usepackage{ulem}

\usepackage{xparse}
\usepackage{colortbl}

\newcommand{\modelname}{\textbf{Uni-MoE-2.0-Omni}}

\usepackage{svg}

\title{
\vspace{-2em}
\fontsize{16}{19}\selectfont Uni-MoE-2.0-Omni: Scaling Language-Centric Omnimodal Large Model  with Advanced MoE, Training and Data}
\author{
 Yunxin Li\thanks{Contributions are shown in Sec.~\ref{sec:contributors}. $\ddagger$ indicates the corresponding author and project leader.}~, Xinyu Chen, Shenyuan Jiang, Haoyuan Shi, Zhenyu Liu, Xuanyu Zhang, Nanhao Deng, \\  {Zhenran Xu, Yicheng Ma, Meishan Zhang, Baotian Hu$^\ddagger$, Min Zhang$^\ddagger$} \\
Research Institute of Computing and Intelligence\\
{Harbin Institute of Technology, Shenzhen}\\
\href{https://huggingface.co/collections/HIT-TMG/lychee-uni-moe-20}{\includegraphics[height=1em]{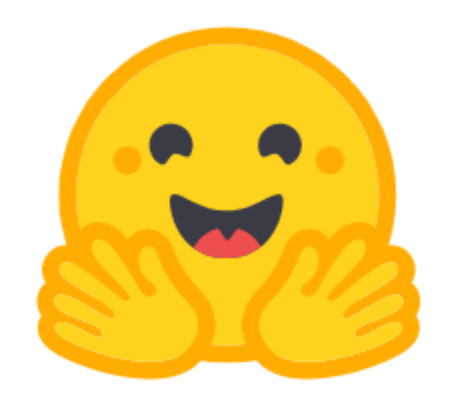} Models}   \href{https://github.com/HITsz-TMG/Uni-MoE}{\includegraphics[height=1em]{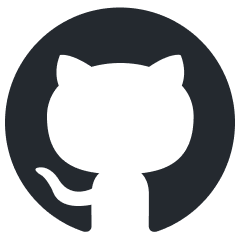} Codes}
\href{https://idealistxy.github.io/Uni-MoE-v2.github.io/}{\includegraphics[height=1em]{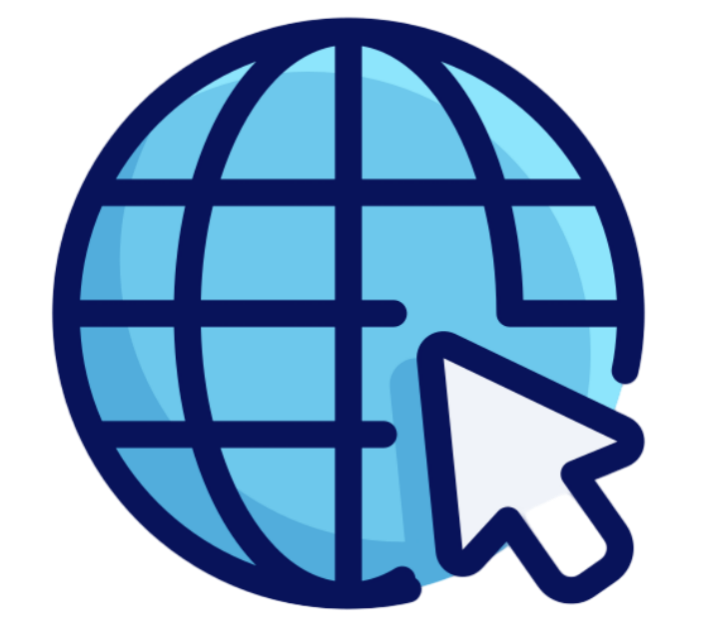} Website}
}

\begin{document}

\maketitle
\vspace{-1em}

\begin{abstract}
We present Uni-MoE 2.0 from the Lychee family. As a fully open-source omnimodal large model (OLM), it substantially advances Lychee's Uni-MoE series in language-centric multimodal understanding, reasoning, and generating. Based on the dense LLM, we build \modelname~from scratch through three core contributions: dynamic-capacity Mixture-of-Experts (MoE) design, a progressive training strategy enhanced with an iterative reinforcement strategy, and a carefully curated multimodal data matching technique. It is capable of omnimodal understanding, as well as generating images, text, and speech.
\textit{Architecturally}, our new MoE framework balances computational efficiency and capability for 10 cross-modal inputs using shared, routed, and null experts, while our Omni-Modality 3D RoPE ensures spatio-temporal cross-modality alignment in the self-attention layer.
\textit{For training}, following cross-modal pretraining, we use a progressive supervised fine-tuning strategy that activates modality-specific experts and is enhanced by balanced data composition and an iterative GSPO-DPO method to stabilise RL training and improve reasoning.
\textit{Data-wise}, the base model, trained on approximately 75B tokens of open-source multimodal data, is equipped with special speech and image generation tokens, allowing it to learn these generative tasks by conditioning its outputs on linguistic cues.
Extensive evaluation across 85 benchmarks demonstrates that our model achieves state-of-the-art or highly competitive performance against leading OLMs, surpassing Qwen2.5-Omni (trained with 1.2T tokens) on over 50 of 76 benchmarks. Key strengths include video understanding (+7\% avg. of 8), omnimodallity understanding (+7\% avg. of 4), and audiovisual reasoning (+4\%). It also advances long-form speech processing (WER reduced by up to 4.2\%) and leads in low-level image processing and controllable generation across 5 metrics.
\end{abstract}
\begin{figure}[htbp]
    \centering
    \includegraphics[width=0.8\linewidth]{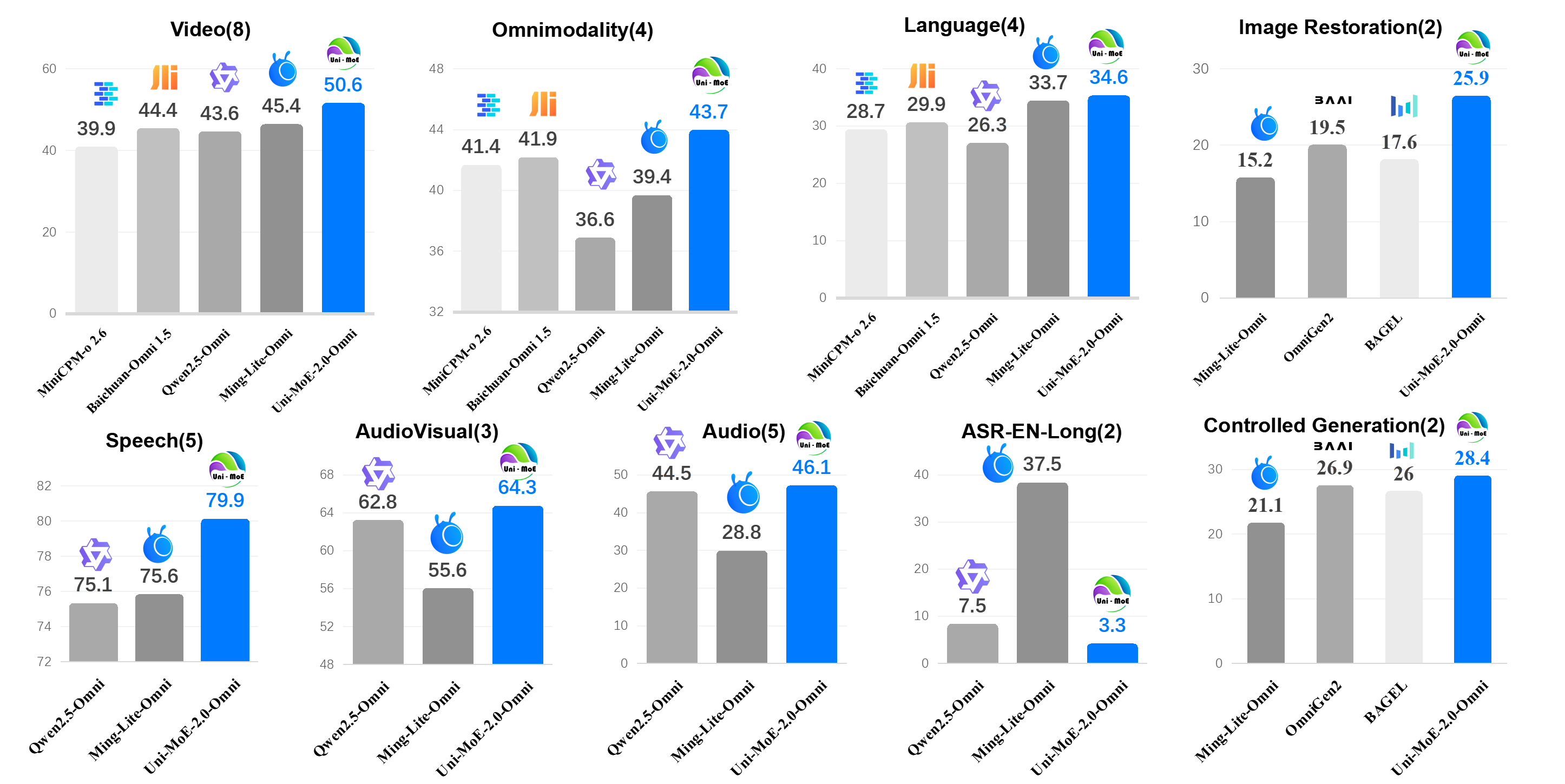}
    \caption{\small The performance of Uni-MoE-2.0-Omni and previous SOTA omnimodal large models.}
    \label{fig:figure_abs}
\end{figure}

\newpage
{
  \hypersetup{linkcolor=RoyalBlue, linktoc=page}
  \tableofcontents
}

% \newpage

\section{Introduction}
\label{sec:introduction}

The pursuit of Artificial Intelligence (AI) has long been guided by a vision of creating systems that can perceive, reason, and interact with the world with human-like breadth and fluency. 
Recently, the field of artificial intelligence has been rapidly advancing towards omnimodal systems, e.g., Qwen-Omni~\citep{qwen_omni}, Ming-Omni~\citep{ming_omni}, Gemini~\citep{gemini25}, and GPT-4o~\citep{gpt4o_system}, which aim to seamlessly integrate and process a diverse array of data types, including text, images, audio, and beyond, within a single, unified large model. 
These omnimodal models represent a pivotal step towards more general and versatile AI, promising models capable of richer understanding and more natural interaction with the complex, multi-sensory world \citep{DBLP:journals/corr/abs-2505-04921,DBLP:journals/corr/abs-2504-21277}. This is not only for pushing the frontiers of AI research but also for enabling a wide range of real-world applications, from advanced human-computer interfaces \citep{DBLP:conf/nips/XieZCLZCHCSLLXZ24,DBLP:journals/corr/abs-2501-12326} and content creation tools \citep{DBLP:conf/nips/SchickDDRLHZCS23,DBLP:journals/corr/abs-2503-23278} to super-intelligence AI assistants that can comprehend and reason across all modalities.

Initial forays, predominantly from well-resourced industrial labs \citep{qwen_omni,ming_omni}, have demonstrated impressive capabilities. However, the architectural and algorithmic path to a genuinely comprehensive omnimodal large model is fraught with complexity. \underline{A central challenge} lies in the inherent tension between two core competencies: deep context understanding and high-fidelity content generation. Many existing systems are architected with a bias, excelling either as powerful context understanding and reasoning (e.g., audio-only Qwen-Omni \citep{qwen_omni}, Baichuan-Omni \citep{DBLP:journals/corr/abs-2501-15368}) that lack generative faculties or as generative powerhouses (e.g., OmniGen \citep{DBLP:journals/corr/abs-2506-18871}, BAGEL \citep{deng2025bagel}, Janus-Pro~\citep{DBLP:journals/corr/abs-2501-17811}, Show-o \citep{DBLP:conf/iclr/XieMBZWLGCYS25}) confined to a narrow set of modalities.
Underlying this issue is the inefficient scaling of model capacity, where simply expanding dense transformers \citep{DBLP:journals/corr/abs-2307-10802,li_unimoe} proves computationally prohibitive and difficult to balance dozens of task types with multiple modal interactions. 
The challenge shows \underline{two critical shortcomings}: a lack of robust and efficient compositional reasoning architecture across multiple modalities, and significant instability when training on heterogeneous data at scale. Hence, the field still lacks a computationally efficient and unified architecture that achieves a synthesis of comprehensive understanding and versatile generation for all modalities.

\textit{This work is anchored in a language-centric perspective to develop an efficient omnimodal large model capable of unified multimodal understanding, reasoning, and generation}. Our approach is predicated on the view that language, serving as a structured representation of the world and a mediator between modalities, provides the cornerstone for effective multimodal integration and transition. This architectural choice is motivated by established advantages of LLMs-based multimodal models \citep{DBLP:journals/corr/abs-2306-13549,DBLP:journals/corr/abs-2505-09777,DBLP:conf/wacv/CuiMCYZLCLYLGLTCZLYMCWZ24}, including great training stability and straightforward scalability to new scenarios.

Specifically, we build \textbf{Uni-MoE-2.0-Omni}, a fully open-source, diverse MoE-based Omnimodal Large Model (OLM). Evolved from the dense Qwen2.5-7B language model, this work demonstrates an efficient pathway to scale a LLM into a powerful and comprehensive OLM.
Through progressive architectural evolution and optimized training strategies, we have successfully transformed the base model into an omnimodal model, achieving the crucial transition from mere multimodal understanding to seamless integration of both understanding and generation.
Our work is founded on \textbf{three key architectural contributions} (Sec. \ref{sec:method}):
\begin{itemsize*}
    \item \textit{Unified Modality Encoding \& Generation}: We design a unified speech encoder that maps diverse audio inputs, including environmental sound, speech, and music, into a shared representation space. For output, a Context-Aware MoE-based TTS module supports dynamic speech synthesis (especially for long speech) and interaction. On the visual side, we employ pre-trained visual encoders to process images and videos, and build a Task-Aware Diffusion Transformer for instruction-guided image generation and editing.
    \item \textit{Deep Cross-Modal Alignment}: To enable deep and efficient fusion of any modality, we introduce an Omni-Modality 3D RoPE mechanism in the self-attention layers. It encodes the temporal-spatial dimensions of speech, image, text, and video tokens, ensuring seamless alignment and interaction across all input types.
    \item \textit{MoE-Driven Cross-Modal Fusion}: We strategically extend the standard MLP layers to MoE layers. This new MoE architecture incorporates three expert types: null experts for inference-time computation skipping, modality-specific routed experts for storing modality knowledge and processing cross-modal information, and small-size shared experts to facilitate universal information exchange. This design enables efficient computation, specialized modality handling, and effective and stable multimodal fusion.
\end{itemsize*}

At training and data recipes,  we introduce \textbf{the following training recipe with the data matching} (Sec. \ref{sec:training_data}) : 
\begin{itemsize*}
    \item  \textit{Progressive Omnimodal Training Optimization}:
To mitigate the instability often encountered when training MoE-based omnimodal large models, we designed a progressive training strategy. This approach sequentially advances through: cross-modal alignment → expert warm-up → MoE fine-tuning and reinforcement learning → generative training. This process efficiently scales dense LLMs into MoE-based omnimodal large models with minimal data requirements, while ensuring convergence stability during iterative reinforcement training (GSPO-DPO) in omnimodal data environments.

\item \textit{Language-Centric Hybrid Training for Multimodal Understanding and Generation}: To bridge the gap between multimodal understanding and generation tasks—often treated separately during training—we propose a hybrid training approach anchored in language generation tasks. By unifying tasks such as image editing, image generation, and speech synthesis within a language generation framework, we break down the inherent barriers between understanding and generation, enabling synergistic enhancement and bidirectional empowerment of both capabilities.
\end{itemsize*}

% \begin{itemsize*}
%     \item  \textit{Language-Centric Cross-Modal Pretraining}: We pretrain the LLM on a corpus of paired modality-to-language data (e.g., image-text, audio-text, video-text). This phase teaches the model to encode and interpret diverse modal inputs by projecting them into a shared semantic space aligned with language.
    
%     \item \textit{Progressive SFT with Modality Experts}: We employ a progressive SFT strategy using modality-specific experts (grouped into audio, vision, and text categories). To enable conditional generation, we introduce special tokens that condition output generation on linguistic cues. This allows the model to learn complex tasks like text-conditioned speech synthesis and image generation directly within the SFT stage.
    
%     \item \textit{Data-Balanced Annealing}: After large-scale SFT, we introduce an annealing phase where we carefully balance the data mixture across all modalities and tasks. This phase uses a reduced learning rate to gently refine the model's performance, ensuring no single modality or task dominates the final model behavior, leading to consistently superior results on omnimodal benchmarks.

%     \item \textit{Iterative Policy Optimization}: To activate long-form reasoning while ensuring training stability for our MoE model, we introduce an iterative GSPO-DPO training method. This approach mainly uses the LLM as a judge to evaluate online rollouts, automatically constructing high-quality preference pairs for offline learning without causing training collapse.

% \end{itemsize*}

The comprehensive evaluation of \modelname~across 85 multimodal benchmarks reveals that Uni-MoE 2.0 further pushes the boundaries of omnimodal intelligence. In comparison to leading omnimodal models, e.g., Qwen2.5-Omni (trained with 1.2T Tokens), Ming-Lite-Omni-1.0/1.5, Baichuan-Omni-1.5, MiniCPM-o 2.6, and other task-specific models, our model mainly achieves superior results in video understanding and reasoning (averaging + 4\% than Ming-Lite-Omni-1.5 on 8 benchmarks), omnimodality comprehension (averaging + 7\% than Qwen2.5-Omni on 4 benchmarks, including popular OmniVideoBench and WorldSense), long speech understanding ($\downarrow$ 3.5\% lower WER than Qwen2.5-Omni on LibriSpeech-clean/other-long) and generation ($\downarrow$ 1\% lower WER on TinyStories-en), and AudioVisual tasks (averaging + 4\% than Qwen2.5-Omni on Speech-Image QA (Reasoning)). Furthermore, \modelname~exhibits competitive performance in most image generation and editing tasks, outperforming strong generative models (e.g., OmniGen and BAGEL) in image editing, controllable generation and low-level image processing tasks, e.g., + 0.5\% than Ming-Lite-Omni on GEdit-Bench and suppressing previous SOTA Qwen-Image and PixWizard on 6 metrics.
We open-source all training and inference codes (GitHub), five model checkpoints (Hugging Face), and data lists (Appendix) to support reproducibility and further research. 

\section{Uni-MoE-2.0-Omni}
\label{sec:method}

\begin{figure}[t]
    \centering
    \includegraphics[width=0.9\linewidth]{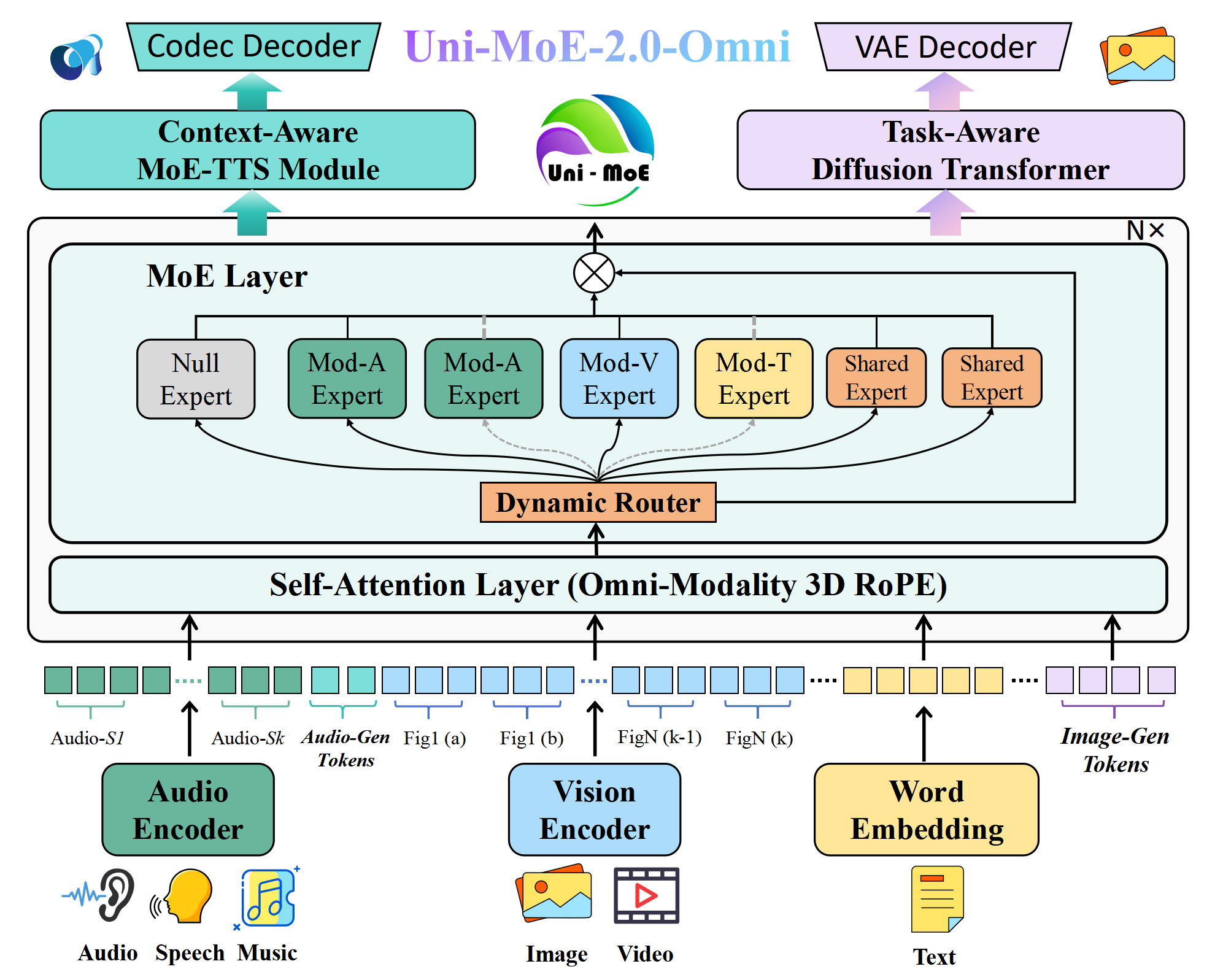}
    \caption{The Uni-MoE-2.0-Omni architecture processes multimodal data through a unified tokenization strategy. Audio is tokenized in 30-second clips, augmented with generation tokens for voice control in the Context-Aware MoE-TTS module, while images are encoded using a sliding window technique. Image Generation Tokens bridge the model to a Task-Aware Diffusion Transformer for end-to-end generation tasks. The model's comprehension is powered by Omni-Modality 3D RoPE, which aligns inputs across time, and a dynamic-capacity MoE layer. This MoE layer dynamically routes information using diverse experts, with stability ensured by null experts (for token skipping) and modality-specific routed experts (A, V, T indicate audio, visual, and textual expert pretrained on corresponding data). In contrast, compact shared experts (only 1/8 size of routed experts) enable efficient cross-modal knowledge transfer.}
    \label{fig:model_arc}
\end{figure}

\subsection{Overview}

Our model processes multimodal inputs—including audio, images, text, and video—through a unified tokenization scheme, as illustrated in Figure \ref{fig:model_arc}. As detailed in Sec. \ref{sec:percpetion}, audio is segmented into 30-second clips, with each clip represented by a sequence of 200 tokens (20 tokens/3s). High-resolution images are encoded using a sliding window method, where each 384×384 patch is independently tokenized. To enhance the model's comprehension of multimodal data (Sec. \ref{sec:mainarchitec}),  we introduce Omni-Modality 3D RoPE, a mechanism that temporally aligns inputs from speech, text, video, and images. Within the MoE layer, we employ diverse experts to enable seamless switching between specialized modules. This is stabilized by null experts to realize a token skipping layer, modal-specific routed experts that ensure balanced training, and enhanced by small-size shared experts (only 1/8 of routed experts) that facilitate cross-modal knowledge transfer. For generation tasks (Sec. \ref{sec:generation}), the model utilizes dedicated control tokens. Audio Generation tokens govern language and role attributes, which are leveraged by context-aware MoE-TTS modules. Similarly, Image Generation tokens, composed of task and content instructions, bridge the foundation model to a task-aware diffusion transformer, enabling end-to-end image generation and editing.

\begin{table}[t]
\centering
\vspace{-1mm}
\begin{tabular}{lccc}
\toprule
\textbf{Module}         & \textbf{Architecture}      & \textbf{Params} \\
\midrule
Audio Encoder  & Whisper-Large-v3 &     637M    \\
Vision Encoder & SigLIP-So400M    &   398M     \\
MoE-LLM & MoE Transformer   & 26B\\
MoE-TTS & MoE Transformer& 1.2BA0.7B\\
Task-DiT & Dense Transformer & 1.5B \\
Codec Decoder & WavTokenizer-large-600-24k-4096 & 442M\\
VAE Decoder & SD-XL & 49M \\ 
\midrule
Shared Expert & MLP & 712M \\
Routed Expert & MLP & 5.7B \\
Activated Expert & 2 Shared Expert + 0\textasciitilde 3 Routed Expert & Min: 1.5B; Max: 18B\\
\bottomrule
\end{tabular}
\caption{\textbf{The architectural design of Uni-MoE-2.0-Omni.}}
\vspace{-10pt}
\end{table}

\subsection{Uni-Perception}
\label{sec:percpetion}

\paragraph{Vision Understanding}

The visual understanding module of Uni-MoE-2.0-Omni employs a unified encoding strategy for both images and videos. The overall architecture consists of two major components: a visual encoder and a mapping network (MLP). 
\begin{itemsize*}
\item \textit{Visual Encoder}. The visual encoder is initialized with SigLIP \citep{zhai2023sigmoid} vision transformer, which transforms the input image or video frames $V$ into visual features $Z_V$. We adopt the output of the last layer of the transformer as the visual features.
\item \textit{Mapping Network}. The mapping network is composed of a two-layer MLP and a 2D average pooling layer. Specifically, the MLP takes the visual features as input and projects them into the representation space of the language model, thereby producing a one-dimensional sequence of visual features $H_V = p(Z_V)$. Subsequently, the average pooling layer performs length compression along both spatial dimensions, enabling more efficient training.
\end{itemsize*}

Through these visual understanding modules, we can uniformly encode Single-Image, Multi-Image, and Video inputs, thereby transforming different visual input paradigms into a unified one-dimensional sequence of visual representations. The specific encoding strategies for each paradigm are described as follows:

\begin{itemsize*}

\item \textit{Single-Image}. For a given single image with arbitrary resolution, we preprocess the input while preserving its aspect ratio. Suppose the original resolution of the input image is $(h, w)$. We traverse a set of candidate resolutions and select a target resolution $(h', w')$ that is closest to the original resolution while requiring minimal padding. The image is then resized so that either its height or width matches the target resolution, while the other dimension is padded with blank pixels to reach the target resolution. Each candidate resolution is constrained such that both its height and width are integer multiples of the vision encoder patch size $p$, ensuring compatibility with the vision encoder input specification. After this preprocessing, the image can be partitioned into $a \times b$ visual patches, where $a = h'/p$ and $b = w'/p$, with each patch of size $p \times p$. Assuming there are $T$ tokens each visual patch, the total number of visual tokens is $ (a \times b) \times T$.

\item \textit{Multi-Image}. The encoding of multiple images follows a similar procedure to that of a single image. For each image, we independently search the most suitable target resolution, resize it accordingly, and then convert it into the corresponding set of visual patches. Assuming there are $n$ images are provided as input, the total number of visual tokens is $\sum^{n}_{i=1}((a_i\times b_i) \times  T)$, where $a_i$ and $b_i$ denote the number of patches along the height and width of the $i$-th image.

\item \textit{Video}. For video data, we adopt the minimum resolution accepted by the vision encoder $(p,p)$ as the target resolution. Each frame of the video is directly resized to this resolution. Regarding frame selection, we uniformly sample the original video at a rate of one frame per second, yielding $f_s$ sampled frames. If the number of sampled frames is less than a predefined lower bound $f_l$ or greater than an upper bound $f_u$, we uniformly resample the video to satisfy these constraints.
Consequently, the final number of frames is determined as $f_n=min(max(f_s, f_l), f_u)$, and the total number of visual tokens is $f_n \times T$.

\end{itemsize*}

This unified encoding of Single-Image, Multi-Image, and Video inputs into a one-dimensional sequence of visual representations enables the transfer of the model’s capability in high-resolution image understanding to the video domain, thereby facilitating faster convergence and improved performance in video understanding.

\paragraph{Speech Understanding}

Speech understanding plays a crucial and indispensable role in speech dialogue models. However, for the speech input of most speech dialogue models, they merely conduct analysis and compilation on semantic information, thus paying relatively less attention to prosodic features such as intonation and timbre information. A model has the potential to comprehensively understand multiple types of audio information in a unified manner through a larger-scale and powerful encoding structure in terms of comprehension ability. This architecture is comprises of two main components:

\begin{itemsize*}
\item \textit{Audio Encoder}. We adopt the Whisper-Large-v3 encoder \citep{whisper} as our speech encoding module and conduct training on a diverse set of audio datasets. Specifically, input audio signals $A$ first undergo resampling to a unified sample rate of 16000 Hz; subsequently, these resampled audio signals are fed into the Whisper encoder, which generates audio features $Z_A$. Notably, the audio features output by this encoder not only encapsulate rich intrinsic information from the input audio but also integrate the generalized speech representation capabilities acquired by the pre-trained Whisper encoder during its prior training phase.
\end{itemsize*}
\begin{itemsize*}
\item \textit{Audio-Language Mapping}.Through comparison, it is found that by using the decoder module of Whisper-Large-v3 as the Qformer and adopting a mapping method of 400 tokens per minute, not only can speech information be efficiently extracted, but also a great deal of paralinguistic semantic information, such as timbre, intonation, and emotion, can be obtained. We utilize all decoder layers of Whisper-Large-v3, with the number of query tokens in the Qformer set to 200 (20 tokens/3s). These types of information are ultimately mapped to the textual dimension through the projection layer, enabling the model to effectively utilise this information for understanding and reasoning.
\end{itemsize*}
Specifically, the workflow of speech understanding is described as follows:
\begin{equation}
    \begin{array}{cc}
         X_Q^{A} = [h_Q^{1}, ..., h_Q^{L}],\vspace{2.0pt}\\
         h_W^{A} = \text{WhisperEncoder}(A), \vspace{2.0pt} \\
         h_S^{A} = \text{MSA}(\text{LN}(X_Q^{A})) + X_Q^{A}, \vspace{2.0pt} \\
         h_C^{A} = \text{MCA}(h_W^{A}, \text{LN}(h_S^{A})) + h_S^{A}, \vspace{2.0pt} \\
         h_l^{A} = \text{MLP}(h_C^{A}),
    \end{array}
\end{equation}
where $h_W^{A}$ is the last hidden states of the pre-trained audio encoder adopted from Whisper-large-v3. $\text{MSA}$ and $h\text{MCA}$ denote the multihead self-attention and cross-attention operations, and $\text{LN}$ is the layer norm function. By leveraging the fixed-length query vectors $X_Q^{A}$, each 30 seconds of audio encoded by the Whisper encoder is mapped to  $L$ audio tokens. $h_C^{A}$ represents the output of the cross-attention module, which is used to distill the main content of the input audio. After going through all layers of the Whisper decoder, we apply a learnable linear layer for projecting the last output into the representation space of LLM.

To process audio longer than Whisper-Large-v3's 30-second limit, we employ a chunking mechanism. Long audio is segmented into consecutive 30-second clips, which are batched and processed by the audio understanding module. \textit{The resulting features are then concatenated along the time dimensions, enabling the understanding of audio of arbitrary length}.

\subsection{Main Architecture}
\label{sec:mainarchitec}

\subsubsection{Omni-Modality 3D RoPE}

Inspired by the M-RoPE design in Qwen2-VL~\citep{Qwen2-VL}, we propose Omni-Modality 3D RoPE to enable efficient positional modeling across text, audio, image, and video modalities. Similar to M-RoPE, we decompose the original rotary embedding into three components corresponding to the temporal, height, and width dimensions. For textual inputs, these components share the same position IDs, rendering Omni-Modality 3D RoPE functionally equivalent to the standard 1D-RoPE. For audio inputs, we align the temporal position IDs with absolute time. Specifically, we define 20 tokens as the minimum temporal unit, which corresponds to a duration of 3 seconds. Instead of adopting the original single-step increment of position IDs, we replace it with the variation rate of absolute time. For image inputs, the temporal IDs of visual tokens remain fixed, while the height and width IDs are assigned according to the spatial locations of tokens within the original image. Unlike M-RoPE,  the height and width dimensions are traversed in a patch-wise manner, such that all tokens within a single vision patch are enumerated before proceeding to the next patch. For video inputs, the temporal IDs of each frame are incremented according to \underline{absolute time}, while the height and width components follow the same ID assignment pattern as images.

As an illustrative example, consider a video input of length 120 seconds accompanied by audio. We place the visual sequence before the audio sequence and sample the video at a rate of one frame every two seconds. Suppose the RoPE ID of the last text token preceding the visual sequence is $(x-1, x-1, x-1)$. Then, the RoPE IDs of the first video frame start from $(x, x, x)$ and increment row by row and column by column up to $(x, x+p, x+p)$, where $p$ denotes the number of tokens along both the height and width of a video frame. The second frame begins with $(x+2\theta, x, x)$ and increases to $(x+2\theta, x+p, x+p)$, where $\theta$ is a specific scaling factor for absolute time. Following this procedure, the final frame in the video sequence has an ID of $(x+118\theta, x+p, x+p)$.
Similarly, assume the RoPE ID of the last token before the audio sequence is $(y-1, y-1, y-1)$. The first audio segment is then assigned the ID $(y, y, y)$, which is repeated 20 times to represent the minimum audio unit. The next audio segment receives the ID $(y+3\theta, y+3\theta, y+3\theta)$. Continuing this process, the final audio segment is assigned the ID $(y+117\theta, y+117\theta, y+117\theta)$. In this manner, temporal alignment between the visual and audio sequences can also be achieved.

% ----------------------------
% Vectors
\def\vzero{{\bm{0}}}
\def\vone{{\bm{1}}}
\def\vmu{{\bm{\mu}}}
\def\vtheta{{\bm{\theta}}}
\def\vpi{{\bm{\pi}}}
\def\vphi{{\bm{\phi}}}
\def\vpsi{{\bm{\psi}}}
\def\valpha{{\bm{\alpha}}}
\def\va{{\bm{a}}}
\def\vb{{\bm{b}}}
\def\vc{{\bm{c}}}
\def\vd{{\bm{d}}}
\def\ve{{\bm{e}}}
\def\vf{{\bm{f}}}
\def\vg{{\bm{g}}}
\def\vh{{\bm{h}}}
\def\vi{{\bm{i}}}
\def\vj{{\bm{j}}}
\def\vk{{\bm{k}}}
\def\vl{{\bm{l}}}
\def\vm{{\bm{m}}}
\def\vn{{\bm{n}}}
\def\vo{{\bm{o}}}
\def\vp{{\bm{p}}}
\def\vq{{\bm{q}}}
\def\vr{{\bm{r}}}
\def\vs{{\bm{s}}}
\def\vt{{\bm{t}}}
\def\vu{{\bm{u}}}
\def\vv{{\bm{v}}}
\def\vw{{\bm{w}}}
\def\vx{{\bm{x}}}
\def\vy{{\bm{y}}}
\def\vz{{\bm{z}}}

% Matrix
\def\mA{{\bm{A}}}
\def\mB{{\bm{B}}}
\def\mC{{\bm{C}}}
\def\mD{{\bm{D}}}
\def\mE{{\bm{E}}}
\def\mF{{\bm{F}}}
\def\mG{{\bm{G}}}
\def\mH{{\bm{H}}}
\def\mI{{\bm{I}}}
\def\mJ{{\bm{J}}}
\def\mK{{\bm{K}}}
\def\mL{{\bm{L}}}
\def\mM{{\bm{M}}}
\def\mN{{\bm{N}}}
\def\mO{{\bm{O}}}
\def\mP{{\bm{P}}}
\def\mQ{{\bm{Q}}}
\def\mR{{\bm{R}}}
\def\mS{{\bm{S}}}
\def\mT{{\bm{T}}}
\def\mU{{\bm{U}}}
\def\mV{{\bm{V}}}
\def\mW{{\bm{W}}}
\def\mX{{\bm{X}}}
\def\mY{{\bm{Y}}}
\def\mZ{{\bm{Z}}}
\def\mBeta{{\bm{\beta}}}
\def\mPhi{{\bm{\Phi}}}
\def\mLambda{{\bm{\Lambda}}}
\def\mSigma{{\bm{\Sigma}}}

\subsubsection{Dynamic-Capacity MoE} 

Previous studies have demonstrated that factual and procedural knowledge is predominantly stored in the feed-forward network (FFN) modules~\citep{DBLP:journals/corr/abs-2503-22941}. The Mixture-of-Experts architecture enables adaptive knowledge retrieval by dynamically activating different FFN modules, which are determined by a router network. Particularly, given $n$ expert parameters $\{w_0, ..., w_{n-1}\}$, the output of a vanilla MoE for inference is:

\begin{equation}
    \vy_i = \sum_{i=0}^{n-1} Gating(\vz)_i \cdot TopK(\vz)_i \cdot Expert(\vx, w_i),
\end{equation}
where $\vz = W_r\vx$, as we use linear network as the router, $Gating(*)$ is a gating function (usually softmax), and $Expert(*)$ is a FNN. $TopK(\vz)$ is the Top-K function, i.e., $TopK(\vz)_i := 1$ if $\vz_i$ is among the TopK coordinates of $\vz$ and $TopK(\vz)_i := 0$ otherwise.

Although achieved notable success, vanilla MoE designs suffer from three key limitations:

\begin{itemsize*}
    \item Discrete expert selection hinders gradient backpropagation. Specifically, the non-differentiable nature of the Top-K function prevents gradients from being effectively propagated from $o_i$ back to $x_i$ in the routing equation, leading to biased optimization directions.
    \item Homogeneous expert types fail to distinguish between domain-specific and general knowledge, and cannot support operations such as selectively forgetting outdated knowledge.
    \item The number of activated experts is fixed, which limits the ability to dynamically adjust the amount of retrieved parametric knowledge according to the varying demands of different tokens.
\end{itemsize*}
  
These limitations prevent MoE from fully exploiting its adaptive activation capability, ultimately resulting in suboptimal performance.
To address these challenges, we propose the Dynamic-Capacity MoE architecture, which incorporates (i) routing gradient estimation to enable differentiable expert selection, (ii) explicit expert role specialization to separately model domain-specific and general knowledge, as well as facilitate knowledge deletion, and (iii) dynamic expert number allocation to adaptively control the amount of parametric knowledge.

\paragraph{Routing Gradient Estimation}

A central difficulty in training vanilla MoE models arises from the non-differentiable Top-$K$ operation used in expert selection, which obstructs gradient propagation to the router and leads to biased optimization. To address this, we migrate the gradient estimation strategy proposed in Grin-MoE~\citep{liu2024gringradientinformedmoe}, which integrates straight-through gradient estimators under the framework of numerical methods for ordinary differential equations (ODEs).  This enables end-to-end optimization of both router and experts under sparse activation constraints, improving the stability of router training and allowing more precise token-to-expert assignments. A formal description is given in Appendix~\ref{sec:gradient_estimation_appendix}.

\begin{algorithm}
\DontPrintSemicolon
\KwIn{Router Output $\vz$, Input Hidden State $\vx$, Routed Expert Count $N_r$, Shared Expert Count $N_s$, Routed Expert Set $\mathcal{E}_r$, Shared Expert Set $\mathcal{E}_s$}
\KwOut{Final MoE output $\vy$}

$\bm{m} \gets \mathrm{ActivateExperts}(\vz)$ \tcc{Binary mask of selected experts per token}
$n^{(tok)} \gets \mathrm{sum}(\bm{m}) - N_s$ \tcc{Number of non-shared experts per token}
$\vw \gets \mathrm{MaskedSoftmax}(\vz, \bm{m})$ \tcc{Routing weights over selected experts}

\For{$k \gets 1$ \textbf{to} $N_r$}{
    $\mathcal{I} \gets \{ i \mid n^{(tok)}_i \ge k \}$ \tcc{Tokens still requiring the $k$-th routed expert}
    $\vx^{(sel)} \gets \mathrm{Gather}(\vx, \mathcal{I})$ \tcc{Hidden states of selected tokens}
    $\vz^{(sel)} \gets \mathrm{Gather}(\vz, \mathcal{I})$ \tcc{Router logits of selected tokens}
    $\vw^{(sel)} \gets \mathrm{Gather}(\vw, \mathcal{I})$ \tcc{Routing weights of selected tokens}
    $e^\star \gets \mathrm{Top1}(\vz^{(sel)})$ \tcc{Highest-probability expert index}
    $w^\star \gets \vw^{(sel)}[e^\star]$ \tcc{Routing weight for chosen expert} 
    $\vo^{(sel)} \gets w^\star \cdot \mathrm{FFN}(\vx^{(sel)}, \mathcal{E}_r[e^\star])$ \tcc{Process tokens with chosen expert}
    $\vo^{est} \gets \mathrm{GradientEstimation}(\vo^{(sel)}, \vz^{(sel)})$ \tcc{\textbf{Full gradient propagation using gradient estimation}}
    $\vy \gets \vy + \mathrm{Scatter}(\vo^{est}, \mathcal{I})$ \tcc{Add weighted outputs to final result}
}

\For{$k \gets 1$ \textbf{to} $N_s$}{
    $w^{(s)} \gets \vw_{\mathrm{shared}}[k]$ \tcc{Routing weight for shared expert}
    $e^{(s)}_k \gets \mathcal{E}_s[k]$ \tcc{$k$-th shared expert index}
    $\vo^{(s)} \gets  w^{(s)} \cdot \mathrm{FFN}(\vx, e^{(s)}_k)$ \tcc{Process all tokens with shared expert} 
    $\vy \gets \vy + \vo^{(s)}$ \tcc{Add weighted shared outputs}
}

\Return{$\vy$}
\caption{Dynamic-Capacity MoE Training Procedure}
\label{algo:dynamic_moe}
\end{algorithm}

\paragraph{Expert Role Specialization}

To further enhance the adaptability of MoE and address the limitation of homogeneous expert types, we explicitly categorize experts into three distinct roles:  

\begin{itemsize*}
    
    \item \textbf{Routed Experts}: These are task-specific experts responsible for modeling domain-specific knowledge. They are dynamically activated according to the proposed dynamic capacity routing strategy.  
    \item \textbf{Shared Experts}: These experts capture general, domain-independent knowledge. Unlike routed experts, shared experts are persistently activated for all tokens, ensuring that common knowledge is always available during inference.  
    \item \textbf{Null Experts}: These are ``empty'' experts whose output is identically zero. They serve as a mechanism for selective forgetting, effectively removing outdated or irrelevant knowledge from the model's output. Null experts are also dynamically activated via the dynamic capacity routing strategy.
\end{itemsize*}

This role specialization enables the model to (i) allocate computational resources according to token-specific knowledge demands, (ii) maintain a persistent general-knowledge backbone via shared experts, and (iii) selectively forget outdated or irrelevant knowledge through null experts, thereby improving both adaptability and controllability of MoE-based architectures.

\paragraph{Dynamic Capacity Routing}

Vanilla MoE applies a fixed number of experts to every token, ignoring variations in token complexity and knowledge demand.  
We address this limitation by introducing a dynamic capacity routing strategy, which determines the number of routed experts for each token based on a Top-P sampling.  
Formally, let the router produce a probability vector over routed experts for token $i$:
\[
\mathbf{p}^{(i)} = [p^{(i)}_1, p^{(i)}_2, \dots, p^{(i)}_{N_r}], \quad \sum_{j=1}^{N_r} p^{(i)}_j = 1,
\]
where $N_r$ is the number of routed experts.  
We sort experts in descending order of $p^{(i)}_j$, obtaining a permutation $\pi^{(i)}$ such that:
\[
p^{(i)}_{\pi^{(i)}(1)} \ge p^{(i)}_{\pi^{(i)}(2)} \ge \dots \ge p^{(i)}_{\pi^{(i)}(N_r)}.
\]
The set of activated routed experts for token $i$ is then:
\[
\mathcal{R}_i = \left\{ \pi^{(i)}(1), \dots, \pi^{(i)}(k_i) \right\}, \quad 
k_i = \min \left\{ k \,\middle|\, \sum_{j=1}^k p^{(i)}_{\pi^{(i)}(j)} \ge P \right\},
\]
where $P$ is the cumulative probability threshold (e.g., $P=0.7$).  

\paragraph{Algorithm}

To provide a holistic view of our method, Algorithm~\ref{algo:dynamic_moe} presents the complete training procedure in pseudocode form.  
For simplicity of presentation, we treat null experts as a special case of routed experts in the pseudocode.  
The algorithm integrates the three proposed components into a unified workflow:  
First, \texttt{ActivateExperts} function determines expert activation, combining the Top‑P sampling strategy for routed experts with the persistent activation constraint for shared experts;  
Second, gradient estimation is applied to routed expert outputs to enable end‑to‑end optimization under sparse activation;  
Finally, after computing routed expert contributions, shared expert outputs are computed and incorporated into the MoE representation, injecting general knowledge into the final output.

\subsection{Uni-Generation}
\label{sec:generation}

\begin{figure}
    \centering
    \includegraphics[width=0.7\linewidth]{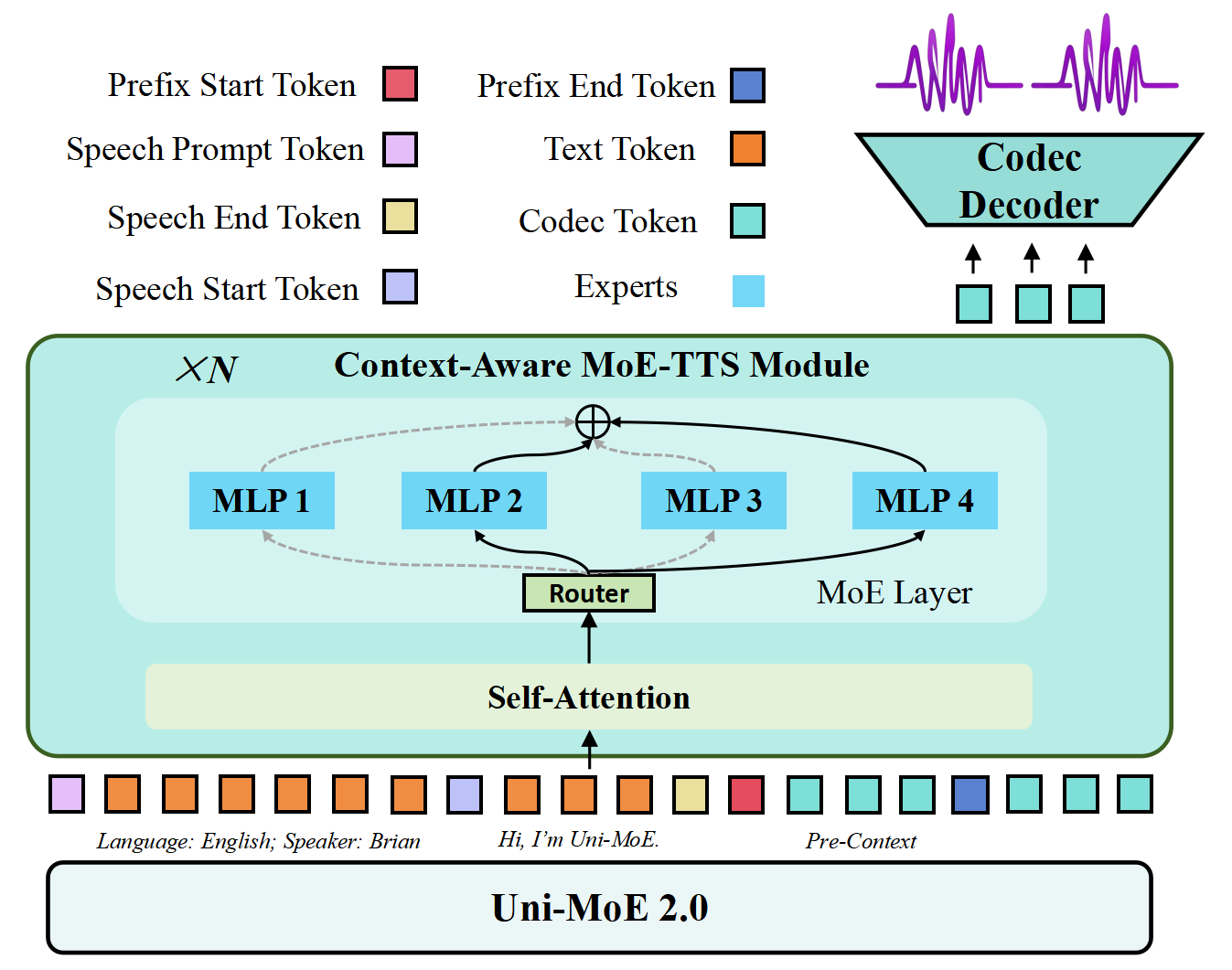}
    \caption{The illustration of Context-Aware MoE-TTS. This figure uses different colored blocks to represent distinct token types, illustrating our long-context streaming decoding method. Furthermore, the Uni-MoE-TTS module will be released separately, featuring three unique and controllable voice styles.}
    \label{fig:model_moe_tts}
\end{figure}
\paragraph{Speech Generation: Context-Aware MoE-TTS} 

Our approach to speech processing employs a dual-strategy architecture to handle input and output efficiently. For speech understanding, we use a continuous encoding method that significantly reduces token consumption (20 tokens/3 seconds). For speech generation, the base model produces both text and control signals. To empower the model with rich audio generation capabilities, including ambient sounds, music, and speech with diverse timbres, emotions, and tones, we integrate the Wavtokenizer \citep{ji2024wavtokenizer} and design a context-aware MoE-TTS module for synthetic voice with three styles. 

As shown in Figure \ref{fig:model_moe_tts}, our MoE-TTS module is built upon a Qwen2.5-0.5B model initialized with a MoE structure. This autoregressive module takes input instructions and text tokens to generate the corresponding speech encodings, which are subsequently decoded into waveform audio by the Wavtokenizer decoder. The training process is divided into two key stages:  First, we pre-train three separate dense models, each on a single-speaker TTS dataset to capture a unique vocal style. Second, this model is transformed into a MoE architecture by replicating its FFN layers to create four experts, which are then fine-tuned on a diverse, multi-style, and multi-speaker dataset. The overall workflow of MoE-TTS is outlined as follows:
\begin{equation}
    \begin{array}{cc}
        s_0 &= [P_1, \ldots, P_n; T_1, \ldots, T_m;], \\
        S_l^{S} &= \text{MSA}(\text{LN}(s_{l - 1})) + s_{l - 1}, \\
        S_l^{M} &= \text{MoE}(\text{LN}(X_l^{s})) + X_l^{s}, \\
        s_l &= \text{LN}(S_l^{M}),
    \end{array}
\end{equation}
where we denote the prompt text, target text content representations to $(P_1, ..., P_n)$ and $(T_1,..., T_m)$ respectively. ``MSA'', ``MoE'' and ``LN'' refer to the multi-head self-attention, mixture of experts and layer normalization. $X_{l}$ shows the output of \textit{l} th block. 

To enhance the performance of MoE-TTS in voice timbre control, we adopt a text-context prompt-guided approach to direct the model to synthesize audio with the specified language and timbre. When the model receives an instruction to generate an audio response, the Uni-MoE base model first produces a special token (i.e. \textit{<speech start>}) indicating the start of speech, . Subsequently, the model generates commands that specify the voice timbre for audio synthesis: fixed timbres include those of Brain, Jenny, and Xiaoxiao, while supported languages cover Chinese and English. Beyond fixed options, the model also enables users to describe desired timbres via natural language; the MoE-TTS module then generates audio with the corresponding timbre based on these descriptive instructions. Upon completion of the timbre prompt, a special token <speech prompt> is generated to signal the end of the prompt. This is followed by the text content that the model needs to transcribe into speech, which is terminated with the special token <speech end>. Through this combination of text-based timbre prompts and target text content, MoE-TTS is guided to perform diverse timbre synthesis.

To synthesize long-form speech, we employ a sentence-splitting strategy during both training and inference. In training, lengthy texts are segmented into short phrases at punctuation marks, which are then batched for efficient processing by MoE-TTS. During inference, we ensure contextual coherence by \textit{using the previously generated speech segment to guide the synthesis of the next}. This split-and-guide approach enables our model to produce coherent and fluent audio clips \underline{exceeding two minutes} in duration\footnote{Context-Aware MoE-TTS are available: \url{https://huggingface.co/HIT-TMG/Uni-MoE-TTS}}.

% \begin{wrapfigure}{r}{0.5\textwidth}
%     \centering
%     \includegraphics[width=0.48\textwidth]{figures/MoE_TTS_model.png}
%     \caption{MoE-TTS.}
%     \label{fig:moe_tts}
% \end{wrapfigure}

\begin{figure}[t]
    \centering
    \includegraphics[width=0.7\linewidth]{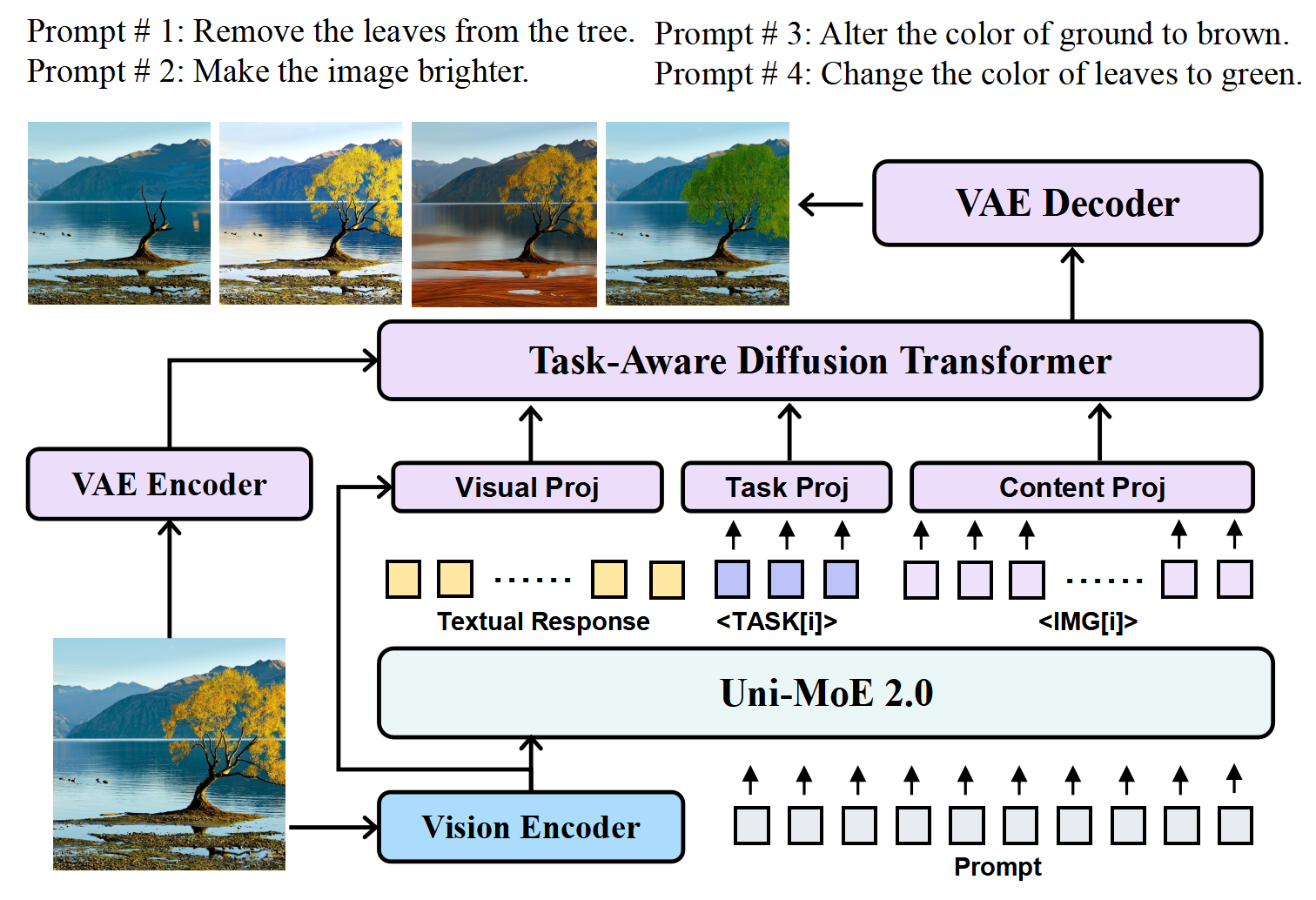}
    \caption{The overview of the Task-aware Diffusion Transformer (Task-DiT). The role of the projection modules is to map external, task-conditioning features into the latent space of the Diffusion Transformer, where they are utilized as context in cross-attention blocks to guide the image generation.}
    \label{fig:task_dit}
\end{figure}

\paragraph{Image Generation: Task-Aware Diffusion Transformer}

% Inspired by recent advancements in unified multimodal models, our work aims to seamlessly integrate image understanding and generation. The primary challenge we address is the inherent performance conflict that arises from unifying these two capabilities. Models that merge representations for both understanding and generation often suffer from a trade-off, where optimizing for one task degrades performance on the other. This is because the feature space optimized for rich semantic comprehension is not naturally aligned with the requirements for high-fidelity visual synthesis. Our architecture is explicitly designed to overcome this limitation by decoupling these functions to preserve the specialized strengths of each domain.

We introduce a Task-Aware Diffusion Transformer (Task-DiT) that bridges the gap between image understanding and generation. Unlike unified models that suffer from performance trade-offs, our framework preserves the strengths of specialized pre-trained models by connecting them via a lightweight, task-aware bridge.
At its core, Task-DiT employs a frozen image generator from PixWizard \citep{lin2024pixwizard} to safeguard its high-fidelity synthesis capabilities. To steer this generator, we introduce a query-based conditioning mechanism. Two distinct sets of learnable tokens are processed by a powerful understanding module:
\begin{itemsize*}
    \item Task Tokens (<TASK[i]>): Encode high-level commands (e.g., text-to-image, editing, low-level image processing) to specify the generative mode.
    \item Image Tokens (<IMG[i]>): Capture the rich semantic essence of the desired output from Uni-MoE-2.0-Omni, forming a compressed scene representation.
\end{itemsize*}

The enriched features from these tokens are then translated for the generator by dedicated, lightweight projectors. A Task Projector modulates the DiT's denoising process based on the command, while a content projector transforms the <IMG[i]> features into a dense conditioning sequence for cross-attention. For image-guided tasks, a Visual Projector aligns source image features encoded by ViT with the DiT's conditioning space.
This design creates a versatile and efficient channel for task-aware instruction, enabling high-quality, multi-modal image generation without the catastrophic interference typical of end-to-end fine-tuned models.

\section{Training and Data Recipes}
\label{sec:training_data}

\subsection{Training Recipe: From LLMs to OLMs}

\paragraph{Alignment with Pretraining}

\begin{figure}[t]
    \centering
    \includegraphics[width=0.92\linewidth]{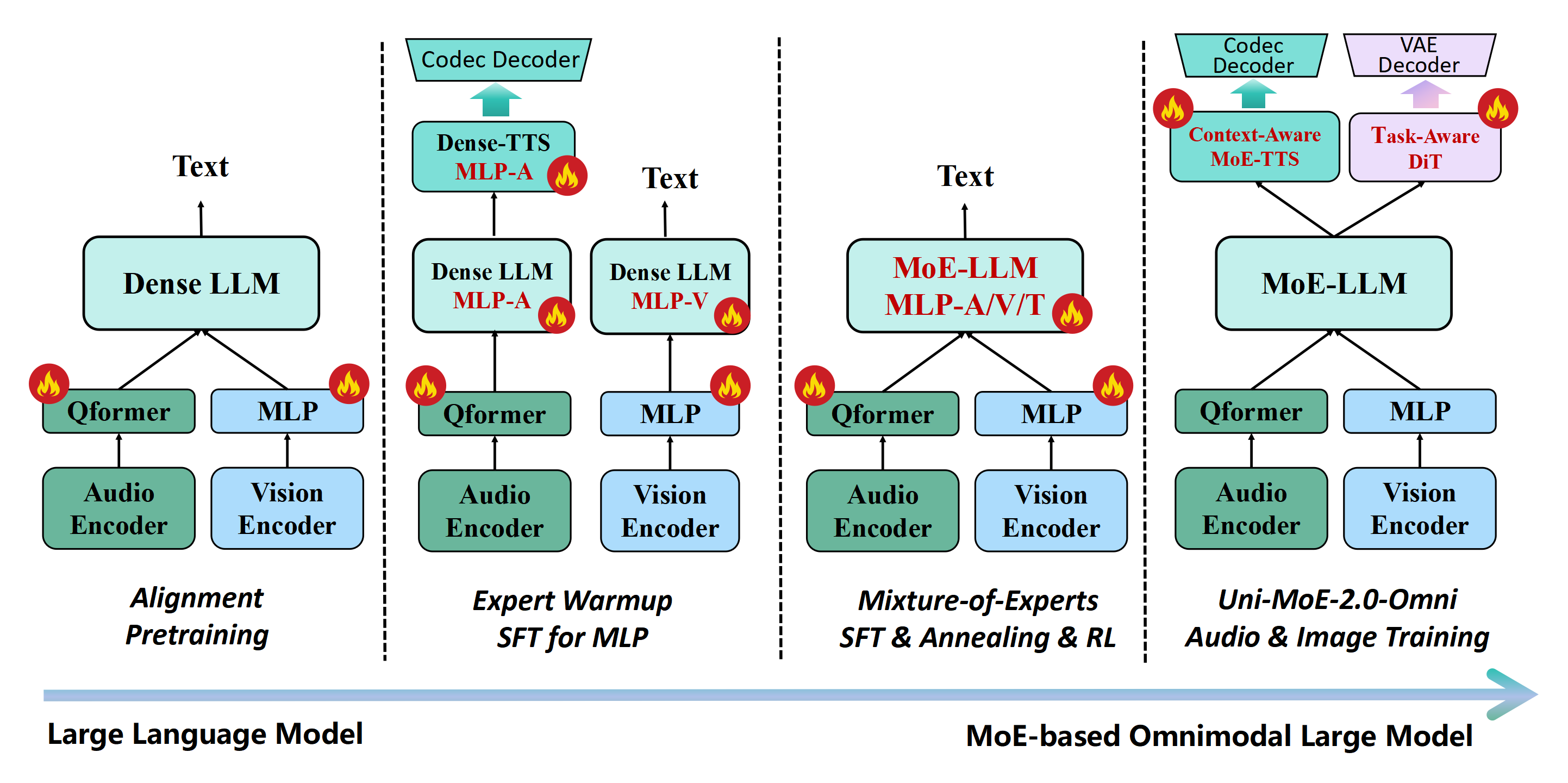}
    \caption{The training recipe for adapting an LLM into an omnimodal large model.}
    \label{fig:training_stage}
\end{figure}

As shown in the leftmost panel of the Figure \ref{fig:training_stage}, the goal of pre-training is to enable the LLM to comprehend multimodal data, such as images, videos, and speech. This is achieved by mapping the representations of these modalities into the LLM's linguistic space, using paired data of multimodal inputs and their corresponding text description.

\paragraph{Supervised Fine-tuning}

In this stage, we collected a large-scale multimodal instruction-following dataset to enable the model to understand and process any cross-modal information. The training comprises two phases:
\begin{itemsize*}
    \item Expert Warmup with Dense Model: We first pre-trained three specialized expert models focusing on mainstream modalities: speech comprehension, speech generation, and visual comprehension. This stage aims to build the model-specific experts for the following stable MoE-based fine-tuning.
    \item Mixed-Data Fine-tuning for MoE-based Model: We then fine-tuned the model on a mixture of all data types using our proposed MoE architecture. The experts in this MoE layer were initialized from the pre-trained experts above, with a final configuration comprising two speech experts, one visual expert, and one null expert. This setup also integrates the inherent language knowledge of the base LLM and two small, shared experts. 
\end{itemsize*}

For generative tasks, the model employs specialized decoding strategies. In speech generation, the Dense LLM outputs both dialogue content and acoustic control signals, which are then rendered into audio by a Dense TTS model. For image tasks, the model first verbalizes its reasoning before generating the final output tokens for tasks like captioning or editing. \textit{This unified approach allows multimodal understanding and generation tasks to be jointly leveraged during instruction tuning, leading to further performance gains.}

\paragraph{Annealing} 

Empirical results indicated that the initial modality-specific expert warmup was insufficient to ensure uniform capability across the diverse spectrum of cross-modal tasks after mixed-data fine-tuning. To further calibrate the model and bridge this performance gap, we conducted a subsequent annealing training phase using a balanced mixture of all data types. Crucially, this annealing process was not confined to the omnimodal foundation model; it was also deployed to fine-tune the specialized sub-modules responsible for image editing/generation and text-to-speech (TTS), promoting stability and proficiency throughout the entire architecture.

\paragraph{Omnimodal Reinforcement Learning}

To develop the Thinking variant of Uni-MoE 2.0, we adopted a training strategy combining cold-start initialization with online reinforcement learning (GSPO) \citep{zheng2025groupsequencepolicyoptimization} and Direct Preference Optimization (DPO) \citep{rafailov2024directpreferenceoptimizationlanguage}. The cold-start phase aims to stimulate the model’s foundational reasoning capabilities across multimodal inputs—including text, images, and video—while online reinforcement learning enhances the model’s autonomous exploration and further refines its reasoning under sparse reward signals. However, during training, we observed limited improvement in the accuracy reward, particularly when processing full-modality data. To address this, we introduced a DPO stage to specifically strengthen the model’s reasoning ability. In this stage, we reuse rollout samples from the online RL phase under an ``LLM as a Judge'' mechanism, selecting those with a logical reasoning process and accurate outcomes as positive examples. Additionally, for samples that yielded zero accuracy during online training, we leverage closed-source commercial models (Gemini-2.5-Flash) as a teacher to generate high-quality reasoning demonstrations, thereby improving the student model, Uni-MoE 2.0. The GSPO-DPO pipeline supports iterative refinement for progressively stronger reasoning performance. A comprehensive treatment of this iterative optimization strategy and associated experimental analysis is provided in VIPO-R1\footnote{Iterative Policy Optimization: \url{https://github.com/HITsz-TMG/VerIPO}} \citep{li2025veripo}. The specific datasets are given in Table \ref{tab:rl_data} of the Appendix.

\paragraph{Generative Training}
The final stage integrates comprehensive omnimodal capabilities by adding speech and image generation. In this phase, the foundational model remains frozen, while the context-aware MoE-TTS and Task-Aware DiT modules are fine-tuned using the corresponding data from the preceding training stages. This approach preserves the model's core omnimodal understanding while efficiently scaling its generative abilities to new modalities. Because the foundation model was pre-trained on data related to image and speech generation, this final training stage converges rapidly, and the new generative modules align effectively with the existing model architecture.

\subsection{Data Recipe}

\subsubsection{Audio-centric Data}

\paragraph{Audio Understanding}

In the pre-training phase of the audio understanding modality, the audio data primarily comprises automatic speech recognition (ASR), audio-caption, and music-caption datasets. Among all pre-training data, ASR datasets constitute the majority, encompassing approximately 15B tokens. In contrast, audio-caption and music-caption datasets are more challenging to acquire than ASR datasets, with a combined total of merely 1B tokens.

During the supervised fine-tuning phase, we reduced the volume of ASR data to 1 billion tokens. To expand the model’s recognition capabilities—originally focused on speech, environmental sounds, and music—toward tasks including speech dialogue, environmental sound question answering, music question answering, and emotion recognition, we collected and constructed a set of relevant datasets for the fine-tuning phase. Furthermore, to activate the model’s capability to generate special tokens required for speech generation, we incorporated data from speech generation tasks, including TTS tasks and speech dialogue tasks, into the training data. In this stage, the total volume of audio data amounts to 5B tokens.

In the annealing stage, we performed sample-level balancing to ensure that the data from all modalities were controlled to be of comparable scale. For the abundant image data, we conducted quality filtering and selected a subset of high-quality samples. For video data, we selected a subset of samples with clear and semantically meaningful audio tracks, which were then expanded into audio–visual unified understanding data. Finally, the image training data amount to approximately 5 billion tokens, while the video training data are expanded to 21 billion tokens.

\begin{table}[t]
\renewcommand{\arraystretch}{0.8}
  \centering
  \resizebox{1.0\textwidth}{!}{%
  \begin{tabular}{lcccc}
    \toprule
    \textbf{Stage} & 
    \makecell{\textbf{Modality} \\ \textbf{Pretraining}} & 
    \makecell{\textbf{Expert} \\ \textbf{Warmup}} & 
    \makecell{\textbf{Omni-Modal} \\ \textbf{Fine-Tuning}} & 
    \makecell{\textbf{Omni-Modal} \\ \textbf{Simulated Annealing}} \\
    \midrule
    \textbf{Data} & \makecell{Image(17M) \\ Video(0.1M) \\ \textit{and} \\ Audio(26M)} 
    & \makecell{Image(11M) \\ Video(2M) \\ \textit{and} \\ Audio(3M)} 
    & \makecell{Image(15M)\\ Video(4M)\\ Audio(9M) \\ Text(4M)} 
    & \makecell{Image(5M)\\ Video(5M)\\ Audio(5M) \\ Text(5M)}
    \\
    \midrule
    \textbf{Tokens} & 13B(i\&v) \& 16B(a) & 19B(i) \& 9B(v) \& 5B(a) & 22B(i) \& 19B(v) \& 8B(a) \& 1B(t) & 5B(i) \& 21B(v) \& 6B(a) \& 4B(t)\\
    \midrule
    \textbf{Training Components} & MLP\&Qformer & MLP\&Qformer\& MLP of LLM & MLP\&Qformer\&MoE & ViT\&MLP\&Qformer\&MoE\\
    \bottomrule
  \end{tabular}
  }
 \caption{Overview of multi-stage training pipeline, detailing data composition, token volume, and trainable components for each stage. The total training volume is about 75B tokens, with datasets shared across stages.}
  \label{tab:training_stage}
\vspace{-10pt}
\end{table}

% \begin{table}[t]
% \renewcommand{\arraystretch}{0.8}
%   \caption{Overview of training stages: data composition, tokens volumes and trainable components}
%   \label{tab:training_stage}
%   \centering
%   \resizebox{1.0\textwidth}{!}{%
%   \begin{tabular}{lcccc}
%     \toprule
%     \textbf{Stage} & 
%     \textbf{Modality Pretrain} & 
%     \textbf{Modality Activation} & 
%     \textbf{Cross-Modal Fine-Tuning} & 
%     \textbf{Cross-Modal Annealing} \\
%     \midrule
%     \textbf{Data} & \makecell{Image Caption \\ Video Caption \\ \textit{or} \\ Audio Caption} 
%     & \makecell{} 
%     & \makecell{Image\\ Video\\ Audio \\ Text} 
%     & \makecell{High Quality \\ All-Modality Data}
%     \\
%     \midrule
%     \textbf{Tokens} & & & 22B + 19B + ? + 1B & 5B + 21B + ? + 4B\\
%     \midrule
%     \textbf{Training Components} & MLP & MLP\&LLM & MLP\&MoE & ViT\&MLP\&MoE\\
%     \bottomrule
%   \end{tabular}
%   }
% \vspace{-10pt}
% \end{table}

\paragraph{Speech Generation}

The training of the MoE-TTS module is divided into three stages. In the first stage, the pre-training activation stage, data with consistent timbre (covering both Chinese and English) is used; this phase primarily aims to activate the capability of the dense model, initialized with the Qwen-0.5B language model, to convert text content of codec tokens into audio content, with a total data volume of 2B tokens. The second stage incorporates a large amount of data with diverse timbres, including two types: data for same-timbre synthesis (where three timbre categories dominate in quantity) and data for stylized custom timbre synthesis (where timbre is controlled via natural language), with a data volume of 5B tokens. In the final stage, training for MoE model activation is conducted using data of the same quantity as the previous stage, resulting in the final version of the MoE-TTS model.

\subsubsection{Vision-centric Data}

\paragraph{Vision-Language Understanding}

During the vision modality pre-training stage, the visual data primarily comprised image–caption and video–caption datasets. The image–caption corpus contained approximately 17 million samples (13 billion tokens), whereas the video–caption data represented a smaller portion, with around 0.1 million samples (0.2 billion tokens).

In the supervised instruction-tuning stage, we curated a large collection of open-source instruction-tuning datasets encompassing a wide range of tasks, including general image understanding, STEM reasoning, document understanding, visual grounding, and video description. At this stage, the image instruction-tuning data consisted of roughly 22 billion tokens, while the video instruction-tuning data accounted for about 19 billion tokens.

During the annealing stage, we applied sample-level balancing to ensure that data from all modalities were maintained at comparable scales. For the abundant image data, we performed quality filtering to retain only a high-quality subset. For the video data, we selected samples with clear and semantically meaningful audio tracks, which were further extended into audio–visual unified understanding datasets. Ultimately, the image training data comprised approximately 5 billion tokens, while the video training data were expanded to 21 billion tokens.

\paragraph{Image Generating and Editing} 

In the task projector alignment stage and the visual projector alignment stage, we utilized data encompassing a variety of task types. This included image generation, image editing, low-level computer vision tasks such as deraining and denoising, and conditional generation like Canny-to-image. In total, this data amounted to approximately 2M samples (1.5B tokens).

In the caption projector alignment stage, we used high-quality image-caption pairs sourced from a diverse range of data types. This collection included factual, human-annotated descriptions of everyday scenes; detailed, context-rich prompts for complex image generation; and elaborate narratives describing the compositional and spatial relationships between elements. This data accounted for a total of 4.2M samples (3.2B tokens).

In the supervised instruction-tuning stage, we curated and filtered a large-scale dataset from multiple open-source dataset collections, amounting to approximately 10.5M samples (8.1B tokens). This stage aimed to unify the model's capabilities under a consistent instruction-following framework. The training data was formatted as instruction-response pairs covering the same core task categories mentioned previously: image generation, image editing, low-level CV enhancements, and conditional image synthesis.

In the annealing stage, we utilized a curated, high-quality dataset of selected samples across various task types. This stage placed particular emphasis on dense captions for image generation, generation processes that incorporate reasoning steps, and various long-tail image editing tasks, such as changing human emotion. This dataset consisted of approximately 3M samples (2.8B tokens). 

\subsubsection{Text-only Data} 
In the Omni-Modal Fine-tuning stage, we introduce pure textual data to further enhance the model’s understanding and reasoning capabilities. For this stage, the textual data primarily consists of text-only instructional data and mathematical question–answer pairs, totalling approximately 1 billion tokens. In the subsequent Omni-Modal Annealing stage, STEM-related data (e.g., codes, math) are incorporated into the training data to further improve model performance, resulting in approximately 4 billion tokens of textual data at this stage.

\section{Experiment}
\label{sec:experiment}

To comprehensively evaluate the capabilities of \modelname, we conducted experiments on 85 benchmarks, including multimodal or omnimodal understanding (image, text, video, cross/tri-modal), audio and music understanding, speech generation, image generation and editing, and low-level image processing tasks. ``Uni-MoE-2.0'' refers to the Uni-MoE-2.0-Omni model without annealing training. ``Uni-MoE-2.0-Base/thinking'' represents the omnimodal understanding and thinking model without speech and image generation abilities.

\begin{table}[t]
    \centering
    \small
    \renewcommand{\arraystretch}{1.4}
    \resizebox{0.99\textwidth}{!}{
    \begin{tabular}{l | c c | c c c c c}
    \toprule
    \multirow{1}{*}{\textbf{Benchmark}} 
        & \textbf{Uni-MoE-2.0}
        & \textbf{Uni-MoE-2.0-Omni} 
        & \textbf{Qwen2.5-Omni}$^*$  
        & \textbf{MiniCPM-o 2.6}$^*$  
        & \textbf{Baichuan-Omni-1.5}$^*$ 
        & \textbf{Ming-Lite-Omni}$^*$  
        & \textbf{Ming-Lite-Omni-1.5}$^*$  \\
    % & \textit{-} & \textit{-} & \textit{7B} & \textit{7B} & \textit{-} & \textit{20B-A3B} & \textit{20B-A3B} \\
    \midrule
    \rowcolor{SkyBlue!15}
    \multicolumn{8}{l}{\textbf{General}} \\
    MMBench-EN~(Dev) & 80.76 & 80.50 & 75.42 & 82.39 & \textbf{82.82} & \underline{82.65} & 82.56 \\
    MMBench-CN~(Dev) & 78.69 & 79.90 & 68.65 & \textbf{80.50} & 68.73 & 78.78 & \underline{80.07} \\
    MMStar & 59.72 & 59.38 & 59.94 & 59.59 & 60.77 & \underline{63.30} & \textbf{63.80} \\
    RealWorldQA & 62.22 & 63.12 & 64.71 & 65.23 & \textbf{66.93} & 64.18 & \underline{66.01} \\
    GQA~(test-dev) & \underline{61.95} & \textbf{62.18} & 49.55 & 59.43 & 58.86 & 61.68 & {61.88} \\
    MME-RealWorld & 53.73 & 53.67 & 49.96 & 46.94 & 52.70 & \underline{58.13} & \textbf{58.55} \\
    CV-Bench & 75.70 & 76.46 & 75.82 & 75.02 & 76.72 & \textbf{79.49} & \underline{79.15} \\
    \midrule
    \rowcolor{SkyBlue!15}
    \multicolumn{8}{l}{\textbf{STEM}} \\
    AI2D & 80.93 & 81.35 & 78.79 & \underline{82.55} & 79.24 & 80.80 & \textbf{82.67} \\
    MMMU~(Val) & 42.67 & 46.67 & 44.44 & 47.33 & 47.11 & \underline{51.78} & \textbf{53.44} \\
    MMMU-Pro~(Standard) & 29.48 & 29.65 & \textbf{36.42} & 30.69 & \underline{33.93} & 33.30 & 32.49 \\ 
    MMMU-Pro~(Vision) & 14.91 & 14.51 & 13.33 & 11.62 & \textbf{25.90} & 12.66 & \underline{15.90} \\
    MathVista~(Testmini) & 60.80 & 61.30 & 56.20 & 66.20 & 59.50 & \textbf{69.50} & \underline{69.00} \\
    MathVision~(Test) & \textbf{40.76} &  \underline{36.61} & 17.14 & 14.21 & 21.13 & 13.29 & 25.20 \\
    LogicVista & 31.47 & 32.81 & 33.93 & \underline{37.05} & 32.59 & \textbf{39.51} & \underline{37.05} \\
    \midrule
    \rowcolor{SkyBlue!15}
    \multicolumn{8}{l}{\textbf{Doc \& OCR}} \\
    DocVQA~(Test) & 79.75 & 79.53 & 88.11 & 82.09 & 89.94 & \underline{92.50} & \textbf{93.55} \\
    ChartQA & 71.64 & 73.04 & 74.80 & 82.80 & 83.28 & \underline{85.56} & \textbf{85.80} \\
    CharXiv~(DQ) & 48.80 & 47.68 & 59.69 & 51.32 & 41.42 & \underline{67.40} & \textbf{68.15} \\
    CharXiv~(RQ) & 23.20 & 24.10 & 25.83 & \underline{28.60} & \textbf{30.70} & 26.00 & 25.83 \\
    SEED-Bench-2-Plus & 64.21 & 64.38 & 68.07 & 65.26 & 66.23 & \textbf{68.47} & \underline{68.25} \\
    \midrule
    \rowcolor{SkyBlue!15}
    \multicolumn{8}{l}{\textbf{Video}} \\
    Video-MME~(w/o sub) & \underline{64.85} & \textbf{66.41} & 59.78 & 60.78 & 59.85 & 62.04 & 62.56 \\
    LongVideoBench~(Val) & \textbf{56.62} & 55.35 & 53.83 & 51.83 & 54.00 & 54.90 & \underline{55.20} \\
    MVBench & \underline{69.33} & \textbf{70.53} & 61.23 & 58.90 & 61.12 & 66.92 & 68.40 \\
    VSI-Bench & \underline{53.87} & \textbf{55.97} & 19.32 & 25.58 & 33.89 & 36.30 & 37.77 \\
    Charades-Sta & 27.73 & \textbf{30.62} & \underline{29.24} & 16.41 & 20.41 & 8.81 & 10.88 \\
    \bottomrule
    \end{tabular}}
    \caption{
        \textbf{Comparison of Uni-MoE-2.0-Omni and variants with other omnimodal models across General, STEM, Doc \& OCR, and Video benchmarks.}
        All results presented in this table are evaluated using the \texttt{lmms-eval} \citep{zhang2024lmmsevalrealitycheckevaluation} to ensure consistency and reproducibility. \textbf{Bold} indicates the highest score, and \underline{underline} indicates the second-highest score for each benchmark.
    }
    \label{tab:results_vision}
    \vspace{-10pt}
\end{table}

\begin{table*}[t]
    \centering
    \scriptsize
    \renewcommand{\arraystretch}{1.0}
    \setlength{\tabcolsep}{3pt} % 减小列间距
    \resizebox{\textwidth}{!}{
    \begin{tabular}{lcccccccccc}
    \toprule
    \textbf{Model} & Video-MME & VideoMMMU & LongVideoBench & MVBench & VSI-Bench & Charades-Sta & TOMATO & EgoSchema & \textbf{Avg.} \\
    \midrule
    \rowcolor{SkyBlue!15}
    \multicolumn{10}{l}{\textbf{Vision-Language Models}} \\
    LLaVA-OneVision-7B & 58.2 & 33.9 & 50.5 & 56.7 & - & - & - & \underline{60.1} & - \\
    LLaVA-Video-7B & 63.3 & 36.1 & 58.2 & 58.6 & - & - & - & 57.3 & - \\
    NVILA-8B & 64.2 & 20.9 & 57.7 & 68.1 & - & - & - & 54.3 & - \\
    VideoLLaMA3-7B & \underline{66.2} & 47.0 & \textbf{59.8} & 69.7 & - & - & - & \textbf{63.3} & - \\
    InternVL2.5-8B* & 64.1 & 46.0 & \underline{58.9} & \textbf{71.9} & 34.6 & 9.5 & 28.0 & 51.2 & 45.5 \\
    Qwen2.5-VL-7B* & 63.0 & \textbf{49.6} & 57.6 & 66.3 & 37.7 & \textbf{40.3} & 22.6 & 58.4 & \underline{49.4} \\
    \midrule
    \rowcolor{SkyBlue!15}
    \multicolumn{10}{l}{\textbf{Omni Models}} \\
    Qwen2.5-Omni* & 59.8 & 43.6 & 55.1 & 61.2 & 19.3 & 29.2 & 25.5 & 53.8 & 43.5 \\
    MiniCPM-o 2.6* & 60.8 & 37.6 & 51.8 & 58.9 & 25.6 & 16.4 & 25.0 & 43.2 & 39.9 \\
    Baichuan-Omni-1.5* & 59.9 & 43.5 & 54.0 & 61.1 & 33.9 & 20.4 & 25.3 & 57.5 & 44.4 \\
    Ming-Lite-Omni* & 62.0 & 48.8 & 54.9 & 66.9 & 36.3 & 8.8 & \underline{28.2} & 57.0 & 45.4 \\
    Ming-Lite-Omni-1.5* & 62.6 & \underline{49.3} & 55.2 & 68.4 & 37.8 & 10.9 & \textbf{34.2} & 54.5 & 46.6 \\
    \midrule
    \rowcolor{SkyBlue!15}
    \multicolumn{10}{l}{\textbf{Our Models}} \\
    Uni-MoE-2.0 & 64.9 & 39.1 & 56.6 & 69.3 & \underline{53.9} & 27.7 & 27.0 & 52.2 & 46.3 \\
    Uni-MoE-2.0-Omni & \textbf{66.4} & 43.6 & 55.4 & \underline{70.5} & \textbf{56.0} & \underline{30.6} & 27.8 & 54.3 & \textbf{50.6} \\
    \bottomrule
    \end{tabular}
    }
    \caption{
        \textbf{Comparison of Uni-MoE-2.0-Omni and variants with other MLLMs across 8 Video benchmarks.}
       * denotes the reproduced results. When evaluating Video-MME, the subtitles are not used.
        \textbf{Bold} indicates the highest score, and \underline{underline} indicates the second-highest score for each benchmark.
    }
    \label{tab:results_video}
\end{table*}

\begin{table}[t]
    \centering
    \small
    \renewcommand{\arraystretch}{1.2}
    \resizebox{\textwidth}{!}{
    \begin{tabular}{l | c c | c c c c c}
    \toprule
    \textbf{Benchmark} & \textbf{Uni-MoE-2.0} & \textbf{Uni-MoE-2.0-Omni} & \textbf{Qwen2.5-Omni*} & \textbf{MiniCPM-o 2.6*} & \textbf{Baichuan-Omni 1.5*} & \textbf{Ming-Lite-Omni*} & \textbf{Ming-Lite-Omni-1.5*} \\
    \midrule
    GPQA Diamond & 30.30 & \textbf{32.83} & 27.27 & 30.81 & 26.77 & 24.24 & \underline{31.31} \\
    GPQA Main & 32.37 & \textbf{33.48} & 23.66 & 29.30 & 25.46 & \underline{32.59} & 31.70 \\
    GPQA Extended & 32.05 & 33.15 & 22.16 & 24.78 & 25.00 & \textbf{34.70} & \underline{34.43} \\
    MMLU-Pro & 36.00 & 38.76 & 32.23 & 29.80 & 42.43 & \underline{42.72} & \textbf{44.75} \\
    \hline
    \textbf{Avg.} & 32.68 & \underline{34.56} & 26.33 & 28.67 & 29.92 & 33.56 & \textbf{35.55} \\
    \bottomrule
    \end{tabular}
    }
    \caption{
        \textbf{Comparison of Uni-MoE-2.0-Omni and other multimodal models on reasoning and general knowledge benchmarks.} All models are tested with direct answers in a zero-shot setting. \textbf{Bold} indicates the highest score, and \underline{underline} indicates the second-highest score for each benchmark.
    }
    \label{tab:results_text}
\end{table}

% \begin{table}[t]
%     \centering
%     \small
%     \renewcommand{\arraystretch}{1.2}
%     \resizebox{0.85\textwidth}{!}{
%     \begin{tabular}{l | c c | c  c  c}
%     \toprule
%     \multirow{2}{*}{\textbf{Benchmark}} & \textbf{Uni-MoE 2.0} & \textbf{Uni-MoE-2.0-Omni} & \textbf{Qwen2.5-Omni} & \textbf{Ming-Lite-Omni} & \textbf{Ming-Lite-Omni-1.5} \\
%     & \textit{-} & \textit{-} & \textit{7B} & \textit{20B-A3B} & \textit{20B-A3B} \\
%     \midrule
%     GPQA Diamond & 30.30 & \textbf{32.83} & 27.27 & 24.24 & 31.31 \\
%     GPQA Main & 32.37 & \textbf{33.48} & 23.66 & 32.59 & 31.70 \\
%     GPQA Extended & 32.05 & 33.15 & 22.16 & \textbf{34.70} & 34.43 \\
%     MMLU-Pro & 36.00 & 38.76 & 32.23 & 42.72 & \textbf{44.75} \\
%     % SimpleQA & 3.17 & 3.03 & 1.73 & 3.93 & 4.07 \\
%     \hline
%     Avg. & 32.68 & \underline{34.56} &  26.33 & 33.56 & \textbf{35.55}\\
%     \bottomrule
%     \end{tabular}
%     }
%     \caption{
%         \textbf{Comparison of Uni-MoE-2.0-Omni and other multimodal models on reasoning and general knowledge benchmarks.} All models are tested with direct answers in a zero-shot setting.
%     }
%     \label{tab:results_text}
% \end{table}

\subsection{Vision-Language Understanding}

\subsubsection{Image Understanding}

As shown in Table~\ref{tab:results_vision}, \textbf{Uni-MoE-2.0-Omni} achieves performance comparable to other Omnimodal Understanding models on both general image understanding and STEM-related image understanding tasks. In particular, Uni-MoE-2.0-Omni attains the highest performance on the GQA (test-dev) and MathVision (Test) benchmarks. For Doc \& OCR tasks, Uni-MoE-2.0-Omni exhibits a certain performance gap compared to the strongest existing Omnimodal Understanding models. We attribute this to the limited availability of Doc \& OCR data during pre-training and the consistently low proportion of such data in subsequent training stages. This observation also, to some extent, highlights the scarcity of publicly available Doc \& OCR datasets in the current open-source ecosystem.

\subsubsection{Video Understanding}

The evaluation results presented in Table~\ref{tab:results_vision} demonstrate that the Uni-MoE-2.0-Omni model achieves SOTA performance across multiple video understanding benchmarks. Specifically, Uni-MoE-2.0-Omni surpasses the previous SOTA model Ming-Lite-Omni-1.5 by approximately 3.85\% on the long video understanding benchmark Video-MME, and by about \textbf{18.20\% on the spatial reasoning benchmark VSI-Bench}.

Furthermore, as shown in Table~\ref{tab:results_video}, Uni-MoE-2.0-Omni demonstrates competitive performance across a diverse set of video understanding and reasoning benchmarks. Notably, our model achieves state-of-the-art results on Video-MME (66.4) and VSI-Bench (56.0), while maintaining strong competitiveness on MVBench (70.5). Compared to other omnimodal models, Uni-MoE-2.0-Omni shows particularly outstanding performance in visual-spatial reasoning (VSI-Bench) and comprehensive video QA (Video-MME), surpassing Qwen2.5-Omni by significant margins of 36.7\% and 6.6\% respectively. These results collectively validate that Uni-MoE-2.0-Omni delivers superior video comprehension capabilities, highlighting its robust generalization and reasoning proficiency in multimodal video understanding.

\subsubsection{Language Capability}

Although the training corpus contains a relatively small amount of text data, as shown in Table \ref{tab:results_text}, our Uni-MoE-2.0-Omni series models demonstrate strong language capabilities on complex reasoning and knowledge benchmarks such as GPQA and MMLU-Pro. Their average performance (32.68–34.56) significantly surpasses that of a comparable competitor, Qwen2.5-Onmi 7B (27.27) and Ming-Lite-Omni-1.5 (31.31), while maintaining competitive results on challenging scientific QA (GPQA Diamond) and comprehensive academic reasoning (MMLU-Pro). These results indicate that Uni-MoE-2.0-Omni not only performs reliably in specialized domain knowledge but also reflects the effectiveness and scalability of its architecture in complex language understanding and reasoning tasks.

\begin{TakeawayBox}{Takeaways: Vision and Language}
Experiments demonstrate Uni-MoE-2.0-Omni's strong multimodal performance. It achieves balanced image understanding, leading on GQA and MathVision, though document OCR lags due to limited data. The model excels particularly in video understanding, setting new SOTA on long-video and spatial reasoning tasks, highlighting its generalization ability. Despite scarce text training data, it outperforms peers on complex reasoning and knowledge benchmarks like GPQA Diamond, validating its scalable and effective architecture for challenging tasks.
\end{TakeawayBox}

\subsection{Audio Understanding and Speech Generation}
\begin{table}[t]
    \centering
    \small
    \renewcommand{\arraystretch}{1.2}
    \resizebox{0.9\textwidth}{!}{
    \begin{tabular}{l | c c | c c c}
    \toprule
    \multirow{1}{*}{\textbf{Benchmark}} & \textbf{Uni-MoE 2.0} & \textbf{Uni-MoE-2.0-Omni} & \textbf{Qwen2-Audio} & \textbf{Qwen2.5-Omni}$^*$  & \textbf{Ming-Lite-Omni}$^*$  \\
    % & \textit{-} & \textit{-} & \textit{7B} & \textit{7B} & \textit{20B-A3B} \\
    \midrule
    \rowcolor{SkyBlue!15}
    \multicolumn{6}{l}{\textbf{Speech Understanding}} \\
    RACE-audio-middle & 90.32 & \underline{89.69} & 28.27 & \textbf{92.76} & 88.30 \\
    RACE-audio-high & \underline{87.62} & 87.19 & 26.95 & \textbf{87.79} & 80.38 \\
    EHSL-short & \underline{88.00} & \textbf{90.00} & 24.00 & 86.00 & 82.00 \\
    EHSL-long & 85.33 & \underline{87.33} & 15.33 & \textbf{90.66} & 83.33 \\
    MELD & \textbf{40.93} & \underline{40.4} & 37.92 & 16.76 & 39.35 \\
    MMAU-speech & 64.69 & \underline{65.00} & - & \textbf{70.97} & 60.88 \\
    MMBench-hint-Speech &\underline{97.98} & \textbf{100}& -& 81.06 &95.34\\
    \textbf{Avg.} & \underline{79.27} & \textbf{79.94} & - & 75.14 & 75.65\\
    \midrule
    \rowcolor{SkyBlue!15}
    \multicolumn{6}{l}{\textbf{Audio Understanding}} \\
    ClothoAQA & 61.76 & \underline{61.83} & 43.45 & \textbf{62.29} & 53.43 \\
    ClothoV1 & \textbf{37.9} & \underline{33.4} & 28.9 & 21.2 & 6.3 \\
    ClothoV2 & \textbf{38.1} & \underline{33.4} & 29.1 & 30.1 & 6.3 \\
    AudioCaps & \underline{33.8} & 33.6 & 40.9 & \textbf{37.1} & 18.5 \\
    MMAU-Sound & 67.17 & \underline{68.06} & - & \textbf{71.97} & 59.3 \\
    \textbf{Avg.} & \textbf{47.74} & \underline{46.05} & - & 44.53 & 28.77\\
    \midrule
    \rowcolor{SkyBlue!15}
    \multicolumn{6}{l}{\textbf{Music Understanding}} \\
    MusicCaps & \underline{23.9} & \textbf{62.4} & 21.8 & 4.00 & 0.5 \\
    MMAU-Music & \underline{59.3} & 56.4 & - & \textbf{65.33} & 52.23 \\
    \bottomrule
    \end{tabular}}
    \caption{
        \textbf{Comparison of Uni-MoE-2.0-Omni and variants with other omnimodal models across Speech/Audio/Music Understanding benchmarks.}
        The accuracy (ACC) metric is employed to assess results of speech understanding, AQA and MMAU; the CIDER is used to evaluate results of all captioning tasks. We found that Ming-Lite-Omni-1.5 often fails to follow instructions and generates off-topic content, making it difficult to evaluate its performance accurately.
    }
    \label{tab:results_audio_x}
\end{table}

\begin{table}[t]
    \centering
    \small
    \renewcommand{\arraystretch}{1.4}
    \resizebox{0.99\textwidth}{!}{
    \begin{tabular}{l | c c | c c c c }
    \toprule
    \multirow{1}{*}{\textbf{Benchmark}} & \textbf{Uni-MoE-2.0} & \textbf{Uni-MoE-2.0-Omni} & \textbf{Qwen2.5-Omni}$^*$ & \textbf{Ming-Lite-Omni}$^*$ & \textbf{Ming-Lite-Omni-1.5}$^*$ & \textbf{Qwen2-Audio} \\
    % & \textit{-} & \textit{-} & \textit{7B} & \textit{20B-A3B} & \textit{20B-A3B}\\
    \midrule
    \rowcolor{SkyBlue!15}
    \multicolumn{7}{l}{\textbf{ASR-EN} $\mathbf{\downarrow}$} \\		
    LibriSpeech-clean & 1.73 & 1.66 & 3.57 & 5.36 & {1.34} & 1.60   \\
    LibriSpeech-other & 3.26 & 3.42 & 7.03 & 9.89 & {2.79} & 3.60 \\
    fleurs-en & 7.78 & 7.72 & 9.74 & 10.16 & 8.07 & {6.90} \\
    mls-en & 5.46 & {5.39} & 6.85 & 9.66 & \textbf{4.04} & 5.40 \\
    CV15-en & \textbf{3.63} & {4.13} & 12.25 & 13.75 & 7.04 & 8.60 \\
    voxpopuli & 10.35 & 9.43 & 9.6 & 10.01 & {7.13} & \textbf{6.84} \\
    LibriSpeech-clean-long & 3.55 & \textbf{2.04} & 7.73 & 43.82 & 61.86 & 11.2   \\
    LibriSpeech-other-long & 6.12 & \textbf{4.2} & 7.98 & 32.2 & 61.49 & 10.3 \\
    Avg. & \underline{5.24} & \textbf{4.75}  & 8.09  &  16.85 & 19.22 & 6.81\\
    \midrule
    \rowcolor{SkyBlue!15}
    \multicolumn{7}{l}{\textbf{ASR-ZH}$\mathbf{\downarrow}$} \\
    Aishell1 & 3.69 & 3.23 & 2.63 & 7.83 & {1.33}  & 1.53 \\
    Aishell2-test-ios & 4.84 & 4.94 & 23.74 & 8.05 & \textbf{2.45} & 2.92 \\
    Aishell2-test-android & 4.84 & 4.84 & 25.14 & 6.19 & \textbf{2.46} & 2.92 \\
    Fleurs-zh & 11.27 & 9.58 & 8.87 & 9.73 & 8.36 & {7.50}  \\
    % Wenetspeech-net & 29.64 & 18.66 & 7.76 & 18.66 & {5.96} & 8.42 & \textbf{5.37}\\
    % Wenetspeech-meeting & 28.93 & 23.25 & 12.61 & 14.9 & {6.86} & 7.16 & \textbf{6.28} \\
    CV15-zh & {3.37} & \textbf{2.97} & 11.01 & 19.08 & 5.96 & 6.90 \\
    Avg. &  5.60 & 5.11 & 14.28 & 10.18 & \textbf{4.12} & \underline{4.35}\\
    \bottomrule
    \end{tabular}}
    \caption{
        \textbf{Comparison of Uni-MoE-2.0-Omni and variants with other omnimodal models across ASR benchmarks.}
        The ASR results presented in this table are evaluated using the word error rate (WER).
        The LibriSpeech-clean/other-long datasets contain speech samples \underline{longer than 3 minutes}.
    }
    \label{tab:results_asr}
\end{table}

\subsubsection{Audio X → Text}
As shown in Table~\ref{tab:results_audio_x} and \ref{tab:results_asr}, Uni-MoE-2.0-Omni demonstrates competitive performance against other omnimodal models and audio large models across Chinese and English ASR, Speech Understanding, Audio Understanding, and Music Understanding benchmarks. 

Uni-MoE-2.0-Omni demonstrates leading performance across diverse audio tasks. It sets a new standard in English Automatic Speech Recognition (ASR), achieving remarkably low Word Error Rates (WER) of 1.73 on LibriSpeech-clean, 3.26 on LibriSpeech-other, and 5.46 on mls-en. The model also delivers strong results in Mandarin ASR, with WERs of 3.69 on Aishell1 and 4.84 on Aishell2. It is worth emphasising that our model possesses strong long-speech understanding capabilities (with an average duration of 3.6 minutes). Specifically, it achieves a low Word Error Rate (WER) of 2.04 and 4.2 on the LibriSpeech-clean-long and LibriSpeech-other-long test sets, two long-audio benchmarks constructed from short-audio synthesis. Beyond transcription, it excels in semantic understanding, attaining high accuracy on spoken question-answering (RACE-audio: 90.32 middle, 87.82 high) and competitive scores on audio captioning tasks (Uni-MoE-2.0-Omni vs. Qwen2.5-Omni: 46.1 vs. 44.5).

However, a performance gap is observed in specialized domains such as Music Understanding when compared to certain benchmarks. We hypothesize that this stems from the relatively lower proportion of high-quality, musically annotated data during pre-training and instruction tuning. This observation not only highlights a specific area for future improvement for our model but also reflects a broader challenge within the multimodal research community: the scarcity of large-scale, open-source music understanding datasets. In summary, Uni-MoE-2.0-Omni proves to be a powerful and versatile model for a wide spectrum of audio-centric tasks, with particularly strong results in speech recognition and general audio comprehension, while its performance in niche domains like music is contingent on the availability of relevant training data.

% However, for some tasks like Music Understanding, there is a performance gap compared to certain stronger omnimodal models. We attribute this to the relatively insufficient specialized music data in the pre-training process and the low proportion of such data in subsequent training stages. This observation also, to some extent, underscores the scarcity of high-quality, publicly available music datasets in the current open-source ecosystem. Overall, Uni-MoE-2.0-Omni showcases remarkable strengths in handling diverse speech and audio-related tasks, especially in ASR and general speech/audio understanding scenarios.

\subsubsection{Text → Speech}

\begin{table*}[t]
    \centering
    \scriptsize
    \renewcommand{\arraystretch}{1.0}
    \setlength{\tabcolsep}{3pt} % 减小列间距
    \resizebox{\textwidth}{!}{
    \begin{tabular}{lcccccccccc}
    \toprule
    \textbf{Model} & LibriTTS-clean & LibriTTS-other & SEED-en & SEED-zh & SEED-hard & TinyStories-en & TinyStories-zh \\
    \midrule
    FireRedTTS & - & - & \underline{1.51} & 3.82 & 17.45 & - & - \\
    MaskGCT & - & - & 2.27 & 2.62 & 10.27 & - & - \\
    E2 TTS & - & - & 1.97 & 2.19 & - & - & - \\
    F5-TTS & - & - & 1.56 & \underline{1.83} & 8.67 & - & - \\
    Llasa & - & - & 1.59 & 2.97 & 11.09 & - & - \\
    CosyVoice & \textbf{3.17} & - & 3.39 & 3.10 & 11.75 & - & - \\
    CosyVoice 2 & - & - & \textbf{1.45} & 2.57 & 6.83 & - & - \\
    GLM-4-Voice & 5.64 & - & 2.91 & {2.10} & - & - & - \\
    Qwen2.5-Omni-7B$^*$ & \underline{5.20} & \textbf{6.68} & {1.73} & \textbf{1.68} & \textbf{2.15} & \underline{6.20} & 8.51 \\
    Ming-Lite-Omni$^*$ & 11.15 & 11.33 & 2.92 & 2.68 & 5.52 & 15.07 & \textbf{4.74} \\
    \midrule
    Dense-TTS & 6.51 & \underline{6.84} & 3.03 & 3.41 & 3.1 & - & - \\
    Uni-MoE-2.0-Omni  & 5.85 & 7.13 & {2.72} & 3.10 & \underline{2.67} & \textbf{5.02} & \underline{7.02} \\
    \bottomrule
    \end{tabular}
    }
    \caption{
        \textbf{Comparison of Uni-MoE-2.0-Omni and variants with other omnimodal models across TTS benchmarks.}
        For English TTS audio results, they are transcribed into text using Whisper. For Chinese TTS audio results, Paraformer is utilized to obtain the transcriptions. Subsequently, these transcriptions are compared with the original text using the WER metric.
        \textbf{Bold} indicates the highest score, and \underline{underline} indicates the second-highest score for each benchmark.
    }
    \label{tab:results_tts}
\end{table*}

As summarised in Table~\ref{tab:results_tts}, we evaluate the text-to-speech (TTS) performance of our model against several omnimodal models on standardized benchmarks. The evaluation methodology involves generating audio from text and then transcribing it back to calculate the Word Error Rate (WER); we employ Whisper for English and Paraformer for Chinese outputs.

The results demonstrate the competitive capability of Uni-MoE-2.0-Omni in Chinese and English speech generation. On the LibriTTS-clean benchmark, it achieves a WER of 5.85, significantly outperforming models like Ming-lite-Omni (11.15) and approaching the performance of the state-of-the-art Qwen2.5-Omni-7B. Furthermore, on the challenging SEED-hard benchmark, Uni-MoE-TTS attains a WER of 2.67, surpassing both Dense-TTS and Ming-Lite-Omni, which underscores its robustness in complex synthesis scenarios.

For long-form speech synthesis, tests on the TinyStory validation sets reveal a nuanced strength profile: Uni-MoE-2.0-Omni excels in English, demonstrating superior prosodic consistency, timbre stability, and linguistic fluency, while Ming-Lite-Omni shows a comparative advantage in Chinese. Overall, our model shows the best comprehensive performance in Chinese and English compared to Ming-Lite-Omni and Qwen2.5-Omni-7B.

In conclusion, Uni-MoE-2.0-Omni proves to be a balanced and effective model for omnimodal TTS, exhibiting particular strength in English synthesis and robust performance across diverse and challenging tasks.

\begin{table*}[t]
    \centering
    \scriptsize
    \renewcommand{\arraystretch}{1.0}
    \setlength{\tabcolsep}{3pt} 
    \resizebox{\textwidth}{!}{
    \begin{tabular}{lcccccccccccc}
    \toprule
    \multirow{2}{*}{\textbf{Model}} & \multicolumn{3}{c|}{LlamaQA} & \multicolumn{3}{c|}{WebQA} & \multicolumn{3}{c|}{BigBench Audio} & \multicolumn{3}{c|}{MultiChallenge Audio}\\
    \cmidrule(lr){2-4} \cmidrule(lr){5-7} \cmidrule(lr){8-10} \cmidrule(lr){11-13}
    & $s \rightarrow t$ & $s \rightarrow s$ & $\Delta_{t-s}$ & $s \rightarrow t$ & $s \rightarrow s$ & $\Delta_{t-s}$  & $s \rightarrow t$ & $s \rightarrow s$ & $\Delta_{t-s}$ & $s \rightarrow t$ & $s \rightarrow s$ & $\Delta_{t-s}$   \\
    \midrule
    Spectron-1B & 21.9 & - & - & 6.1 & - & - & - & - & - & - & - & - \\
    SpeechGPT-7B & 21.6 & - & - & 6.5 & - & - & - & - & - & - & - & - \\
    Freeze-Omni-7B & 72 & - & - & 44.73 & - & - & - & - & - & - & - & - \\
    Moshi-7B & 62.3 & 21.0 & 41.3 & 26.6 & 9.2 & 17.4 & - & - & - & - & - & - \\
    LLaMA-Omni-7B & 67.7 & 49.0 & 18.7 & 33.4 & 23.7 & 9.7 & - & - & - & - & - & - \\
    LLaMA-Omni2-7B & 70.3 & 60.7 & 9.6 & 34.5 & 31.3 & 3.2 & - & - & - & - & - & - \\
    VITA-1.5-7B & 76.7 & - & - & 42.7 & - & - & - & - & - & - & - & - \\
    Stream-Omni-8B & 76.3 & 65.0 & 11.3 & 44.2 & 27.5 & 16.7 & - & - & - & - & - & - \\
    OpenOmni-7B & 74.6 & 67.2 & 7.4 & 44.5 & 28.9 & 15.6 & - & - & - & - & - & - \\
    NExT-Omni-7B & 78.4 & 66.4 & 12 & 45.6 & 28.3 & 17.3 & - & - & - & - & - & - \\
    Step-Audio2-mini-7B & - & - & - & - & - & - & 50.90 & 47.50 & \underline{3.40} & 13.64 & 8.08 & 5.56 \\
    Kimi-Audio-7B & - & - & - & - & - & - & 59.40 & 51.00 & 8.40 & 7.07 & 1.01 & 6.06 \\
    GLM-4-Voice-8B & 74.33 & 65.67 & 8.66 & 45.90 & 43.20 & 2.70 & 44.80 & 42.70 & \textbf{2.10} & 9.09 & 6.06 & 3.03 \\
    Qwen2.5-Omni-7B$^*$  & \underline{77.33} & \textbf{77.33} & 0.00 & \underline{48.28} & \textbf{48.28} & 0.00 & \textbf{58.1} & \textbf{53.8} & 4.3 & \underline{13.13} & \underline{10.61} & \underline{2.52} \\
    Ming-Lite-Omni$^*$  & \textbf{80.33} & 63.66 & 16.67 & \textbf{53.79} & \underline{44.19} & 9.60 & \underline{53.3} & 26.6 & 26.7 & \textbf{27.27} & \textbf{23.23} & 4.04 \\
    \midrule
    Uni-MoE-2.0-Omni  & 75.33 & \underline{75.33} & \textbf{0.00} & 45.13 & 43.95 & \underline{1.18} & 49.2 & \underline{44.7} & 4.5 & 10.61 & 9.6 & \textbf{1.01} \\
    \bottomrule
    \end{tabular}
    }
    \caption{
        \textbf{Comparison of Uni-MoE-2.0-Omni and variants with other omnimodal models across Speech QA benchmarks.}
        We evaluate the model's performance by employing existence matching to assess the ACC of the answers. Some results (models w/o $^*$) are from the MOSS speech and GLM-4-voice, where a lower $\Delta_{t-s}$ value indicates better performance.
    }
    \label{tab:results_s2s}
\end{table*}

\subsubsection{Speech → Speech/Text}
As shown in Table~\ref{tab:results_s2s}, Uni-MoE-2.0-Omni demonstrates competitive performance against other omnimodal models across Speech QA benchmarks, including LlamaQA, WebQA, BigBench Audio, and MultiChallenge Audio. Notably, Uni-MoE-2.0-Omni achieves remarkable results in several key tasks: in the LlamaQA (s→s) subtask, it attains an accuracy of 75.33, ranking second only to the top-performing model, which underscores its strong ability to handle speech-to-speech reasoning in this scenario. In the BigBench Audio (s→s) task, it achieves 44.7, showcasing its robustness in audio-related QA.
For some challenging benchmarks, such as MultiChallenge Audio, Uni-MoE-2.0-Omni still maintains a competitive position, indicating its versatility across diverse speech QA scenarios. While there is a performance gap compared to the strongest models like Ming-Lite-Omni on certain tasks, we attribute this to the relatively limited specialized Knowledge QA data during pre-training and the low proportion of such data in subsequent training stages.

\begin{table*}[t]
    \centering
    \scriptsize
    \renewcommand{\arraystretch}{1.0}
    \setlength{\tabcolsep}{3pt} 
    \resizebox{0.9\textwidth}{!}{
    \begin{tabular}{lccccccc}
    \toprule
    \multirow{3}{*}{\textbf{Model}} & \multicolumn{4}{c|}{\textbf{Speech-Image QA}} & \multicolumn{2}{c}{\textbf{Speech-Video QA}} & \multirow{3}{*}{\textbf{Avg.}} \\
    \cmidrule{2-7}
    & \multicolumn{2}{c|}{A-OK-VQA Speech (Reasoning)} & \multicolumn{2}{c|}{VQAv2 Speech} & \multicolumn{2}{c|}{ActivityNet Speech} \\
    \cmidrule(lr){2-3} \cmidrule(lr){4-5} \cmidrule(lr){6-7}
    & $s,i \rightarrow t$ & $s,i \rightarrow s$ & $s,i \rightarrow t$ & $s,i \rightarrow s$ & $s,v \rightarrow t$ & $s,v \rightarrow s$ \\
    \midrule
    Qwen2.5-Omni-7B* & 58.55 & \underline{51.65} & 76.84 & \underline{71.01} & \underline{60.00} & \textbf{58.91} & \underline{62.83}\\
    Ming-Lite-Omni* & \underline{65.08} & 42.69 & \textbf{81.21} & 30.43 & 58.43 & 0.03 & 55.57 \\
    \midrule
    Uni-MoE-2.0-Omni & \textbf{65.73} & \textbf{52.58} & \underline{78.03} & \textbf{71.15} & \textbf{60.16} & \underline{57.94} & \textbf{64.27}\\
    \bottomrule
    \end{tabular}
    }
    \caption{
        \textbf{Comparison of Uni-MoE-2.0-Omni and variants with other omnimodal models across Speech-Image/Video QA benchmarks.}
        For Speech-Image QA tasks, we evaluate the model's performance using the ACC metric. For Speech-Video QA tasks, we adopt GPT-based evaluation to assess their performance.
    }
    \label{tab:results_mm_s2s}
\end{table*}
\subsubsection{Vision + Speech → Speech/Text}

As shown in Table~\ref{tab:results_mm_s2s}, Uni-MoE-2.0-Omni demonstrates strong and balanced performance across multimodal question-answering tasks, establishing a compelling advantage over comparable models.

In the Speech-Image QA domain, our model achieves top performance on the challenging A-OK-VQA Speech benchmark, leading all competitors in both text (65.73) and speech (52.58) output modalities. This dual dominance highlights Uni-MoE-2.0-Omni's superior capability in integrating visual and auditory information for complex reasoning.

This strength extends to the Speech-Video QA realm. On the ActivityNet benchmark, Uni-MoE-2.0-Omni achieves the highest score (60.16) for text-based answers (s→t), demonstrating excellent comprehension of temporal and visual narratives. More importantly, it maintains a very strong performance for direct speech answers (s→s, 57.94), a task where other models like Ming-Lite-Omni fail catastrophically. This result underscores a key advantage of our model: its unique ability to generate high-quality, contextually accurate responses directly in the speech modality without sacrificing performance, a capability where most other omnimodal models show significant weakness.

Overall, while other models may excel in a single metric, Uni-MoE-2.0-Omni distinguishes itself through its robust and consistent performance across both text and speech output tasks, making it a more versatile and reliable solution for real-world multimodal applications.

\begin{TakeawayBox}{Takeaways: Audio + X →Speech/Text }
Uni-MoE-2.0-Omni demonstrates comprehensive and competitive capabilities across diverse audio and speech tasks, establishing balanced performance in recognition, understanding, and generation while showing particular strength in multimodal integration.
\end{TakeawayBox}

\begin{table*}[t]
    \centering
    \small
    \sisetup{table-align-text-post=false} %
    \resizebox{\textwidth}{!}{%
    \begin{tabular}{
        l
        S[table-format=2.1]
        S[table-format=2.1]
        S[table-format=2.1]
        S[table-format=2.1]
        S[table-format=2.1]
    }
    \toprule
    \textbf{Method} & {WorldSense} & {StreamingBench (Omni)} & {OmniVideoBench} & {OmniBench} & {Avg.} \\
    \midrule
    \rowcolor{SkyBlue!15}
    \multicolumn{6}{l}{\textbf{Vision-Language Models}} \\
    LLaVA-OneVision-7B & 37.7 & 40.8 & {-} & {-} & {-} \\
    LLaVA-Video-7B & 40.2 & 41.7 & {-} & {-} & {-} \\
    Qwen2.5-VL-7B & 38.3 & 45.0 & 29.8 & {-} & {-} \\
    \midrule
    \rowcolor{SkyBlue!15}
    \multicolumn{6}{l}{\textbf{Omni Models}} \\
    Unified-IO-2 XL & 24.7 & {-} & {-} & 38.0 & {-} \\
    Unified-IO-2 XXL & 25.9 & {-} & {-} & 34.0 & {-} \\
    VideoLLaMA 2 & 25.4 & 35.9 & 29.2 & {-} & {-} \\
    Qwen2.5-Omni-7B* & 43.1 & 47.1 & 29.8 & 26.2 & 36.6 \\
    MiniCPM-o 2.6* & 43.2 & \textbf{51.0} & 34.7 & 36.7 & 41.4 \\
    Baichuan-Omni-1.5* & 42.5 & 47.1 & \underline{35.0} & 42.9 & \underline{41.9} \\
    Ming-Lite-Omni* & 42.2 & 38.8 & 33.3 & 43.5 & 39.4 \\
    Ming-Lite-Omni-1.5* & \underline{43.5} & 40.4 & 32.1 & \textbf{47.7} & 40.9 \\
    \midrule
    \rowcolor{SkyBlue!15}
    \multicolumn{6}{l}{\textbf{Our Models}} \\
    Uni-MoE-2.0 & 42.8 & 39.8 & 34.3 & 46.5 & 40.8 \\
    Uni-MoE-2.0-Omni & \textbf{44.7} & \underline{48.1} & \textbf{35.1} & \underline{47.1} & \textbf{43.7} \\
    \bottomrule
    \end{tabular}
    }
    \caption{
        \textbf{Comparison of Uni-MoE-2.0-Omni and other MLLMs across 4 Omnimodal understanding benchmarks.}
         * denotes the reproduced results. \textbf{Bold} indicates the highest score, and \underline{underline} indicates the second-highest score for each benchmark.
    }
    \label{tab:results_omni}
\end{table*}

\subsection{Omnimodality Understanding}

We present in the Table~\ref{tab:results_omni} the performance of the Uni-MoE-2.0-Omni model on four omnimodal benchmarks that require simultaneous understanding of visual and audio information. On WorldSense and OmniVideoBench, which evaluate long video omnimodal comprehension, Uni-MoE-2.0-Omni achieves SOTA performance. On StreamingBench (Omni), which focuses on shorter videos, and OmniBench, which assesses joint image-audio understanding, our model ranks second. Across all evaluation metrics, Uni-MoE-2.0-Omni attains a SOTA overall score of 43.7\%, outperforming the second-best model Baichuan-Omni-1.5 (41.9\%), by approximately 2\%. These results demonstrate that Uni-MoE-2.0-Omni possesses a strong omni-modal understanding capability.

\begin{table*}[t] 
     \centering 
     \normalsize % 使用 \scriptsize 缩小字体 
     \renewcommand{\arraystretch}{1.1} % 调整行高 
     \setlength{\tabcolsep}{4pt} % 调整列间距 
     \resizebox{\textwidth}{!}{ % 确保表格适应页面宽度 
     % l: Model | cc: Gen | cc ccc: Edit 
     \begin{tabular}{l | c c | c c | c c c} 
     \toprule 
     % --- 一级表头 (模型, 分类) --- 
     \multirow{3}{*}{\textbf{Model}} & \multicolumn{2}{c|}{\textbf{Image generation}} & \multicolumn{5}{c}{\textbf{Image edition}} \\ 
     \cmidrule(lr){2-3} \cmidrule(lr){4-8} 
     % --- 二级表头 (Benchmark 名称) --- 
     & \multirow{2}{*}{\textbf{Wise} $\uparrow$ } & \multirow{2}{*}{\textbf{FID} $\downarrow$ } & \multirow{2}{*}{\textbf{GEdit-Bench} $\uparrow$ } & \multirow{2}{*}{\textbf{Emu Edit} $\uparrow$ } & \multicolumn{3}{c}{\textbf{MagicBrush}} \\ 
     \cmidrule(lr){6-8} 
     % --- 三级表头 (MagicBrush 子指标) --- 
     & & & & & \textbf{CLIPimg} $\uparrow$  & \textbf{DINOimg} $\uparrow$  & \textbf{CLIPout} $\uparrow$  \\ 
     %\midrule 
 
     % --- 数据行 (根据新数据填充) --- 
     \midrule 
     \rowcolor{SkyBlue!15} 
     \multicolumn{8}{l}{\textbf{Image Generation Models}} \\ 
     SDXL & 0.48 & 13.63 & - & - & - & - & - \\ 
     PixWizard & 0.43 & 11.99 & 3.20 & 0.039 & 0.907 & 0.811 & 0.298 \\ 
     Qwen-Image & 0.63 & 25.37 & 7.42 & 0.127 & 0.920 & 0.800 & 0.313 \\ 
     \midrule 
     \rowcolor{SkyBlue!15} 
     \multicolumn{8}{l}{\textbf{Omni Models}} \\ 
     JanusPro-7B & 0.41 & 19.82 & - & - & - & - & - \\ 
     Bagel & 0.49 & 25.47 & 6.52 & 0.124 & 0.921 & 0.844 & 0.310 \\ 
     OmniGen & 0.40 & 29.32 & 5.61 & 0.091 & 0.907 & 0.802 & 0.298 \\ 
     OmniGen2 & 0.45 & 31.65 & 6.10 & 0.103 & 0.899 & 0.794 & 0.307 \\ 
     Ming-Lite-Omni & 0.54 & 16.86 & 5.55 & 0.052 & 0.845 & 0.683 & 0.299 \\ 
     Ming-Lite-Omni-1.5 & 0.52 & 32.39 & 6.09 & 0.094 & 0.910 & 0.822 & 0.309 \\ 
     \midrule 
     \rowcolor{SkyBlue!15} 
     \multicolumn{8}{l}{\textbf{Our Models}} \\ 
     Uni-MoE-2.0-Omni & 0.44 & 18.04 & 6.02 & 0.076 & 0.789 & 0.590 & 0.288 \\ 
     Uni-MoE-2.0-Image & 0.46 & 18.95 & 6.00 & 0.080 & 0.854 & 0.714 & 0.293 \\ 
     
     \bottomrule 
     \end{tabular}} 
     \caption{ 
         \textbf{Comparison of models across Image Generation and Image Edition benchmarks.} % 您可以修改此标题 
     } 
     \label{tab:results_visualgen} % 您可以修改此标签 
 \end{table*}
\begin{table*}[t]
    \centering
    \small % 使用 \small 字体
    \renewcommand{\arraystretch}{1.1} % 调整行高
    \setlength{\tabcolsep}{4pt} % 调整列间距
    \resizebox{\textwidth}{!}{ % <-- 已移除, 因为表格变窄了
    
    % l: Model | cccccc: CondGen | cccc: LowLevel
    \begin{tabular}{l | c c c c c c | c c c c}
    \toprule
    % --- L1 Headers (Categories) ---
    \multirow{3}{*}{\textbf{Model}} & \multicolumn{6}{c|}{\textbf{Controllable Generation}} & \multicolumn{4}{c}{\textbf{Low-Level Image Restoration}} \\
    \cmidrule(lr){2-7} \cmidrule(lr){8-11}
    
    % --- L2 Headers (Benchmarks) ---
    & \multicolumn{3}{c|}{\textbf{Canny-to-Image}} & \multicolumn{3}{c|}{\textbf{Depth-to-Image}}
    & \multicolumn{2}{c|}{\textbf{Derain}} & \multicolumn{2}{c}{\textbf{Denoise}} \\
    \cmidrule(lr){2-4} \cmidrule(lr){5-7} \cmidrule(lr){8-9} \cmidrule(lr){10-11}

    % --- L3 Headers (Sub-metrics) ---
    & \textbf{F1-Score} $\uparrow$ & \textbf{FID} $\downarrow$ & \textbf{CLIP-S} $\uparrow$ & \textbf{RMSE} $\downarrow$ & \textbf{FID} $\downarrow$ & \textbf{CLIP-S} $\uparrow$ 
    & \textbf{PSNR} $\uparrow$ & \textbf{SSIM} $\uparrow$ & \textbf{PSNR} $\uparrow$ & \textbf{SSIM} $\uparrow$ \\
    %\midrule

    % --- 数据行 (已过滤) ---
    \midrule
    \rowcolor{SkyBlue!15}
    \multicolumn{11}{l}{\textbf{Image Generation Models}} \\
    PixWizard & 0.24 & \underline{18.32} & 28.88 & 42.61 & 23.41 & 27.59 & 24.62 & 0.77 & \textbf{27.75} & \textbf{0.81} \\
    Qwen-Image & \textbf{0.47} & 37.59 & 27.45 & 51.23 & 27.54 & 24.81 & 26.37 & 0.80 & 22.19 & 0.46 \\
    \midrule
    \rowcolor{SkyBlue!15}
    \multicolumn{11}{l}{\textbf{Omni Models}} \\
    Bagel & 0.17 & 130.44 & 24.36 & 64.85 & 39.40 & 27.53 & 17.14 & 0.59 & 18.14 & 0.25 \\
    OmniGen & {0.42} & 32.15 & 29.25 & \textbf{32.09} & \textbf{16.98} & \textbf{29.81} & 13.57 & 0.22 & 17.05 & 0.19 \\
    OmniGen2 & 0.16 & 45.67 & 27.35 & 59.57 & 52.30 & 26.47 & 22.22 & 0.77 & 16.78 & 0.41 \\
    Ming-Lite-Omni & 0.17 & 154.95 & 21.83 & 85.71 & 126.31 & 20.33 & 11.63 & 0.21 & 17.61 & 0.51 \\
    Ming-Lite-Omni-1.5 & 0.20 & 187.42 & 21.67 & 87.53 & 134.01 & 20.78 & 16.01 & 0.38 & 14.35 & 0.13 \\
    \midrule
    \rowcolor{SkyBlue!15}
    \multicolumn{11}{l}{\textbf{Our Models}} \\
    Uni-MoE-2.0-Omni & 0.24 & 20.23 & 28.58 & 42.41 & 27.45 & 27.00 & 25.41 & \textbf{0.82} & 25.70 & 0.48 \\
    Uni-MoE-2.0-Image & 0.24 & \textbf{18.23} & \textbf{29.25} & 44.23 & \underline{21.91} & \underline{27.60} & \textbf{25.69} & \textbf{0.82} & \underline{26.01} & 0.47 \\
    \bottomrule
    \end{tabular}} % <-- 已移除
    \caption{
        \textbf{Comparison of models across Controllable Generation and Low-Level Image Restoration benchmarks.} % <-- 标题已更新
    }
    \label{tab:results_visualgen2} % 您可以修改此标签
\end{table*}
\subsection{Image Generation and Edition}
As shown in Table~\ref{tab:results_visualgen} and Table~\ref{tab:results_visualgen2}, our Uni-MoE-2.0-Omni models demonstrate strong and versatile performance. They particularly excel in controllable generation and low-level image restoration, significantly outperforming other omni-models. While specialized generators may lead in pure generation, our models remain highly competitive in complex editing and conditional tasks, showcasing their versatility.

\textbf{Image Generation}
In pure image generation, our models are competitive. Notably, Uni-MoE-2.0-Omni (0.44) surpasses the original PixWizard (0.43) on the Wise benchmark. While specialized models like Qwen-Image lead on the Wise, our FID score (18.04) is highly competitive, outperforming several omnimodalmodels. For instance, it is 9.0\% lower (better) than JanusPro-7B (19.82) and 29.2\% lower than Bagel (25.47).

\begin{figure}[t]
    \centering
    \includegraphics[width=0.9\linewidth]{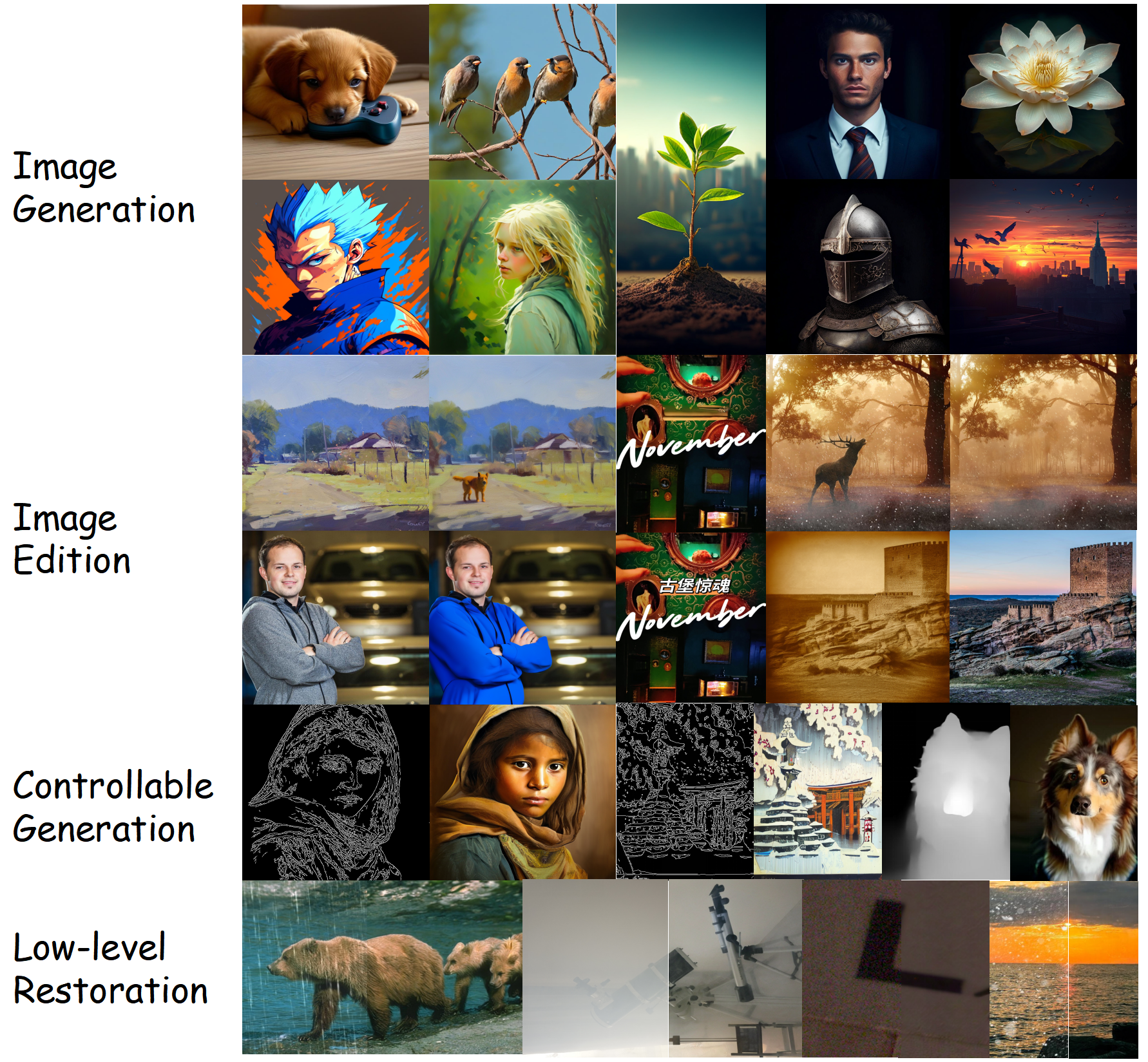}
    \caption{The cases of Image Generation, Image Edition, Controllable Generation, and Low-Level Image Restoration.}
    \label{fig:cases_image_gen}
\end{figure}

\textbf{Image Edition}
Our models show particularly robust performance in image editing. On GEdit-Bench, Uni-MoE-2.0-Omni achieves a score of 6.02, which is 88.1\% higher (better) than the original PixWizard (3.20). On the Emu Edit benchmark, our model's score of 0.076 is also 94.8\% higher than PixWizard (0.039), demonstrating strong performance in instruction-following edits. While state-of-the-art models like Qwen-Image (7.42 GEdiT, 0.127 Emu) still lead, our model is highly competitive. Furthermore, the \mbox{Uni-MoE-2.0-Image} model (MoE and DiT training version) demonstrates even stronger capabilities on complex tasks, especially on the MagicBrush benchmark, where it significantly improves upon the base model's CLIPImg score (0.854 vs. 0.789).

\textbf{Controllable Generation}
This is a particularly strong area for our models. For Canny-to-Image tasks, our \mbox{Uni-MoE-2.0-Image} model achieves a dramatically better FID score (18.23) compared to both the specialized Qwen-Image (37.59) and the omni-model OmniGen2 (45.67). While its F1-Score (0.24) is lower than Qwen-Image (0.47), it surpasses OmniGen2 (0.16). For Depth-to-Image tasks, our model (27.45 FID, 42.41 RMSE) outperforms both Qwen-Image (27.54 FID, 51.23 RMSE) and OmniGen2 (52.30 FID, 59.57 RMSE) on the key metrics of FID and RMSE, demonstrating superior controllable generation.

\textbf{Low-Level Image Restoration}
In low-level image restoration tasks, our \mbox{Uni-MoE-2.0-Omni} model shows superior performance. For Derain tasks, its PSNR (25.41) is highly competitive with the specialized Qwen-Image (26.37) and significantly better than OmniGen2 (22.22). Notably, its SSIM (0.82) is superior to both Qwen-Image (0.80) and OmniGen2 (0.77). For Denoise tasks, our model (25.70) outperforms Qwen-Image (22.19) by 15.8\% and OmniGen2 (16.78) by 53.1\% in terms of PSNR.

\begin{TakeawayBox}{Takeaways: Image Generating and Editing}
While highly competitive in pure image generation, our model's primary strength lies in its versatility and control. It demonstrates state-of-the-art performance in conditional tasks like depth-to-image generation and excels in low-level restoration like image denoising, outperforming strong specialized models and establishing a new benchmark for omnimodal capabilities.
\end{TakeawayBox}

\subsection{MoE Analysis}

\begin{figure*}[htbp]
\centering % 将整个子图网格居中

% --- 第一行 ---
\begin{subfigure}[b]{0.24\textwidth}
    \centering
    \includegraphics[width=\textwidth]{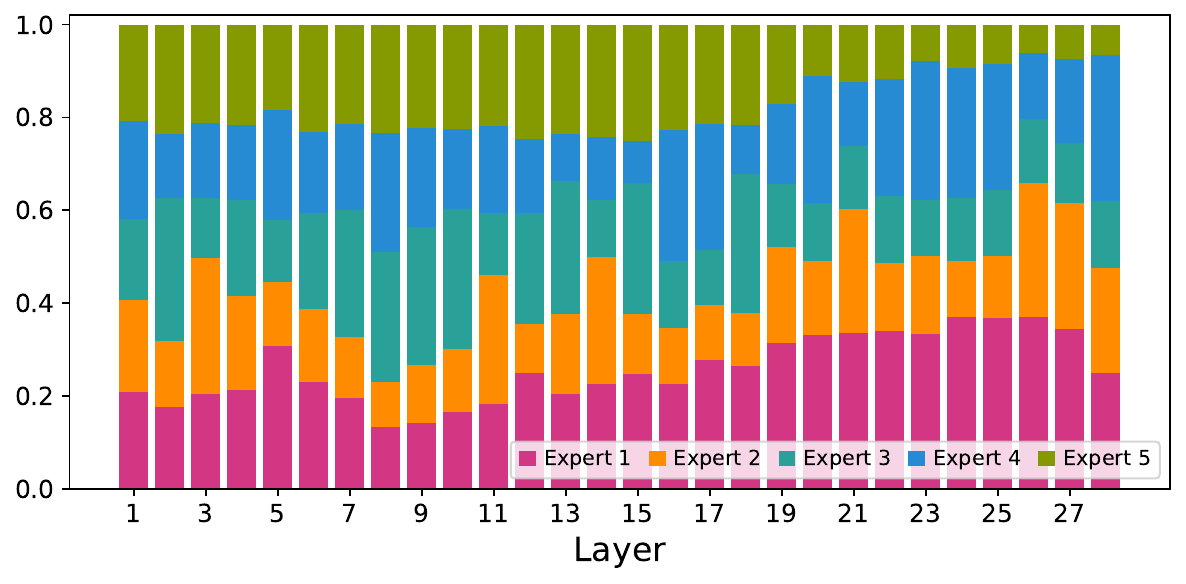}
    \caption{\textbf{Overall}}
    \label{fig:sub_overall}
\end{subfigure}
\begin{subfigure}[b]{0.24\textwidth}
    \centering
    \includegraphics[width=\textwidth]{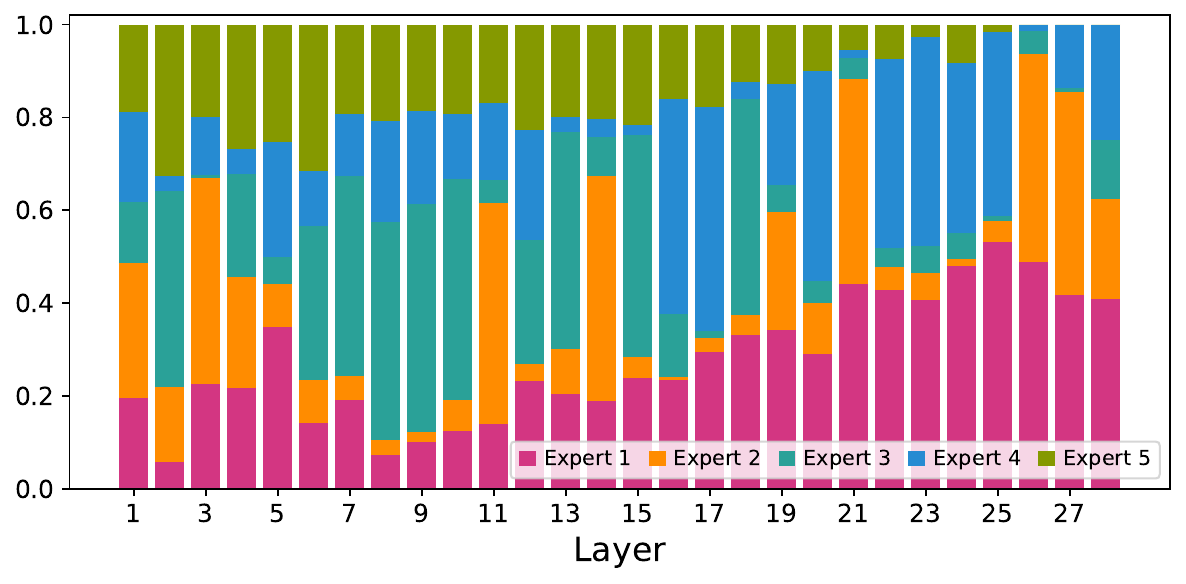}
    \caption{Image Understanding}
    \label{fig:sub_image}
\end{subfigure}
\begin{subfigure}[b]{0.24\textwidth}
    \centering
    \includegraphics[width=\textwidth]{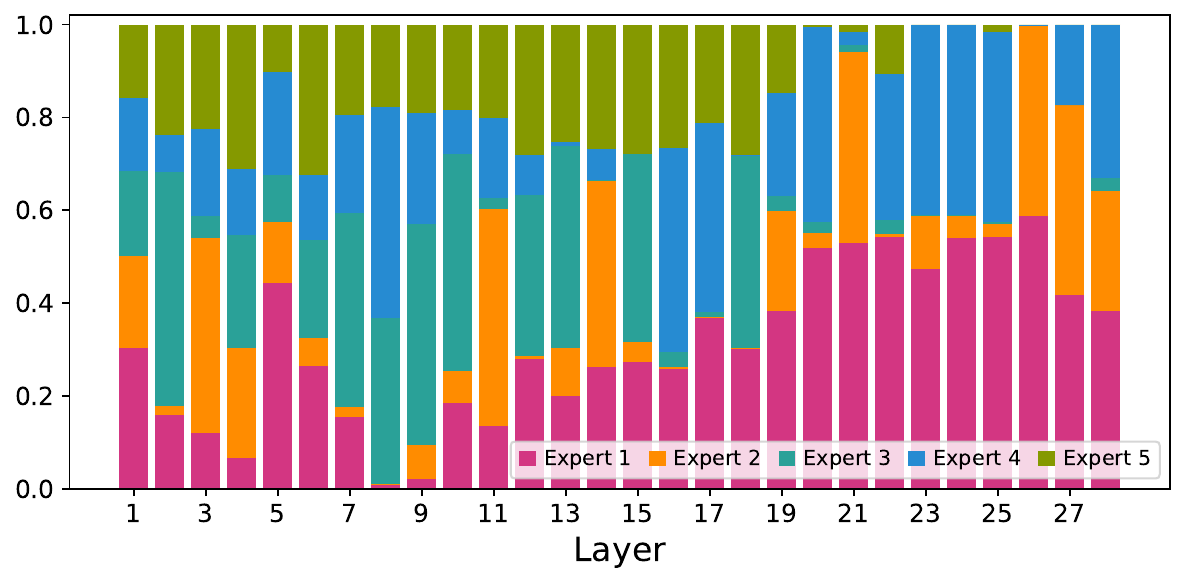}
    \caption{Video Understanding}
    \label{fig:sub_video}
\end{subfigure}
\begin{subfigure}[b]{0.24\textwidth}
    \centering
    \includegraphics[width=\textwidth]{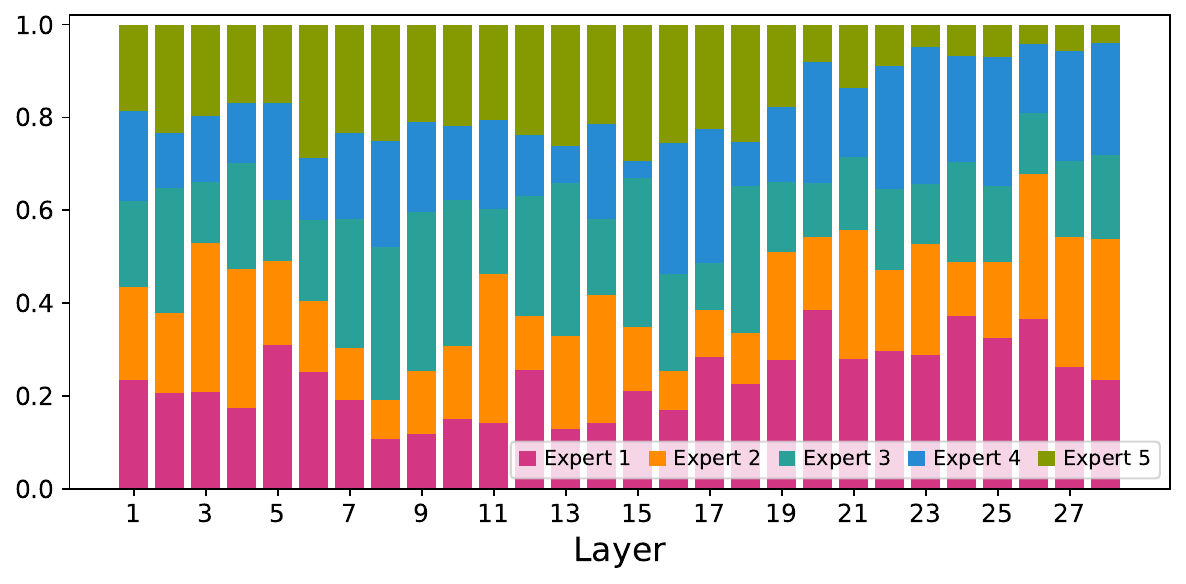}
    \caption{Audio Understanding}
    \label{fig:sub_audio}
\end{subfigure}

\vspace{0.5cm} % 在两行之间添加一些垂直间距

% --- 第二行 ---
\begin{subfigure}[b]{0.24\textwidth}
    \centering
    \includegraphics[width=\textwidth]{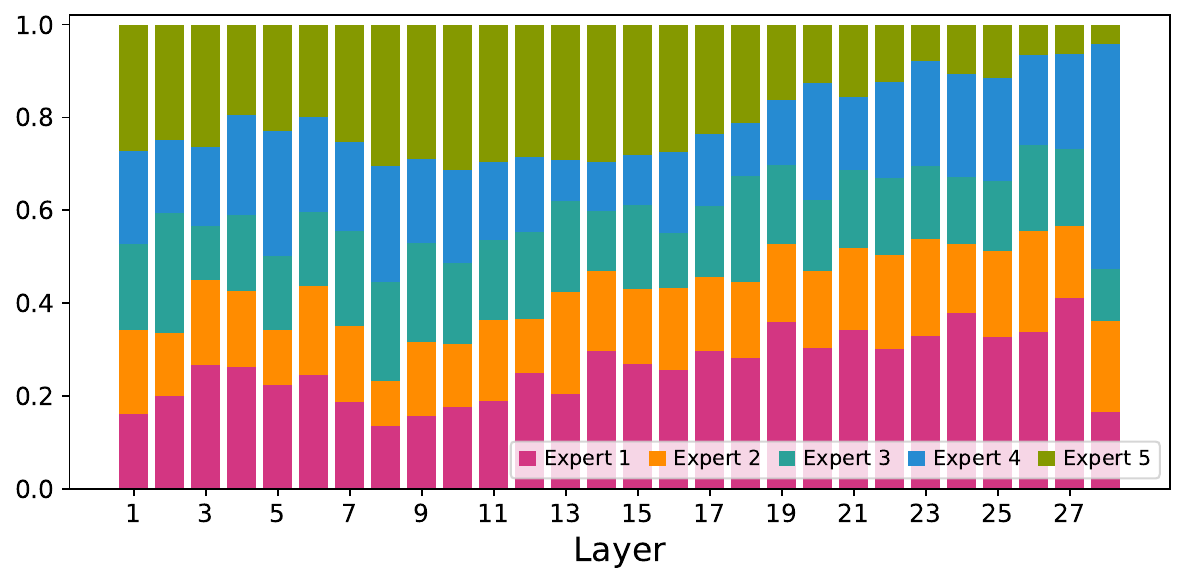}
    \caption{Omni Understanding}
    \label{fig:sub_omni}
\end{subfigure}
\begin{subfigure}[b]{0.24\textwidth}
    \centering
    \includegraphics[width=\textwidth]{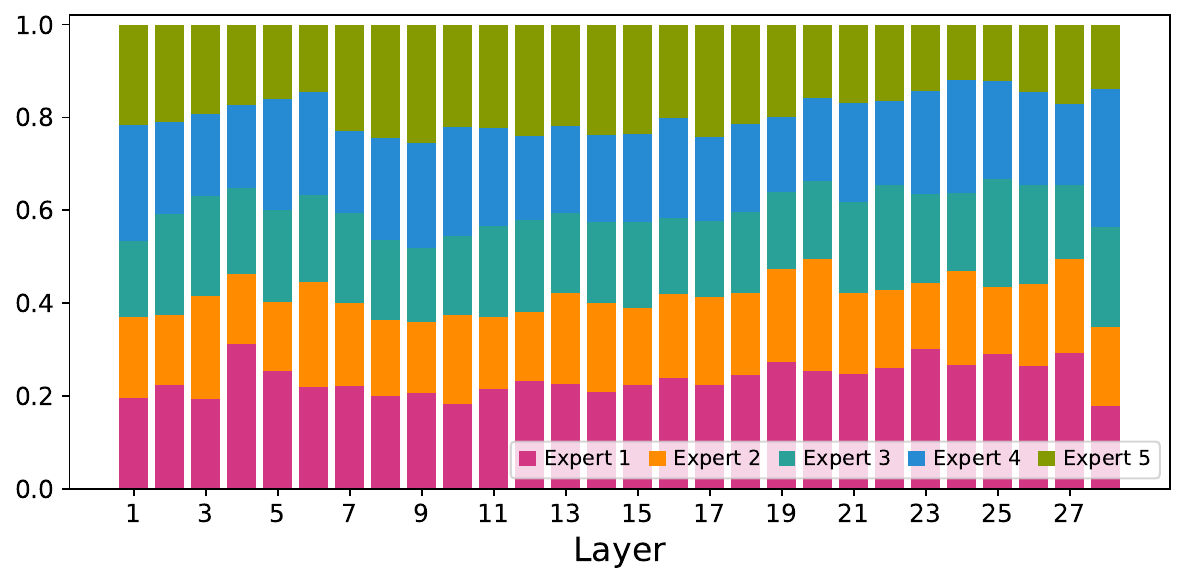}
    \caption{Image Generation.}
    \label{fig:sub_image_gen}
\end{subfigure}
\begin{subfigure}[b]{0.24\textwidth}
    \centering
    \includegraphics[width=\textwidth]{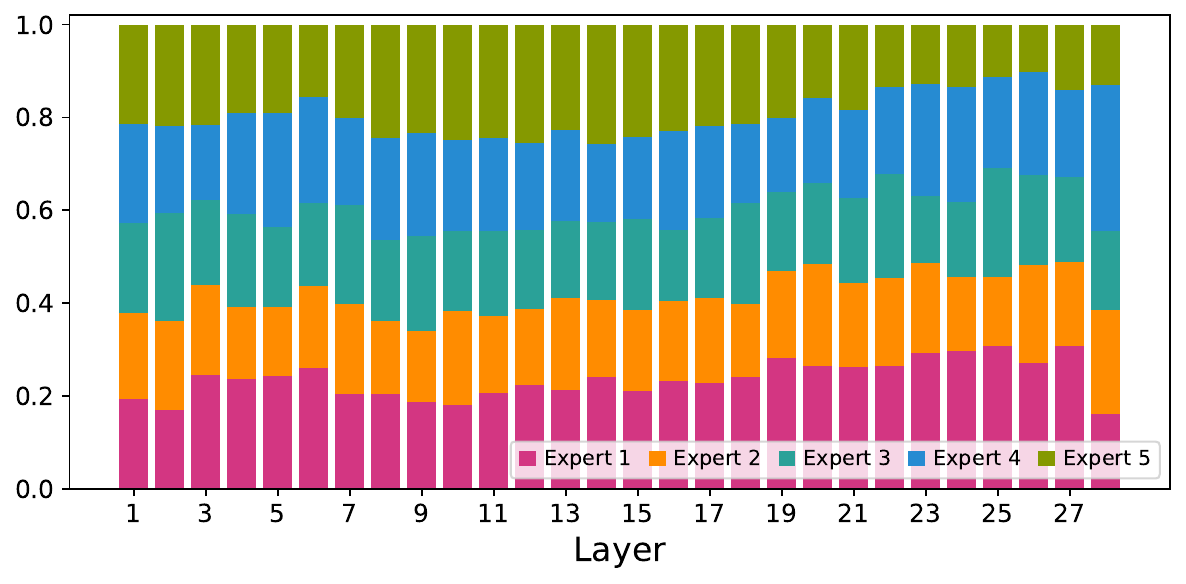}
    \caption{Image Edit}
    \label{fig:sub_image_edit}
\end{subfigure}
\begin{subfigure}[b]{0.24\textwidth}
    \centering
    \includegraphics[width=\textwidth]{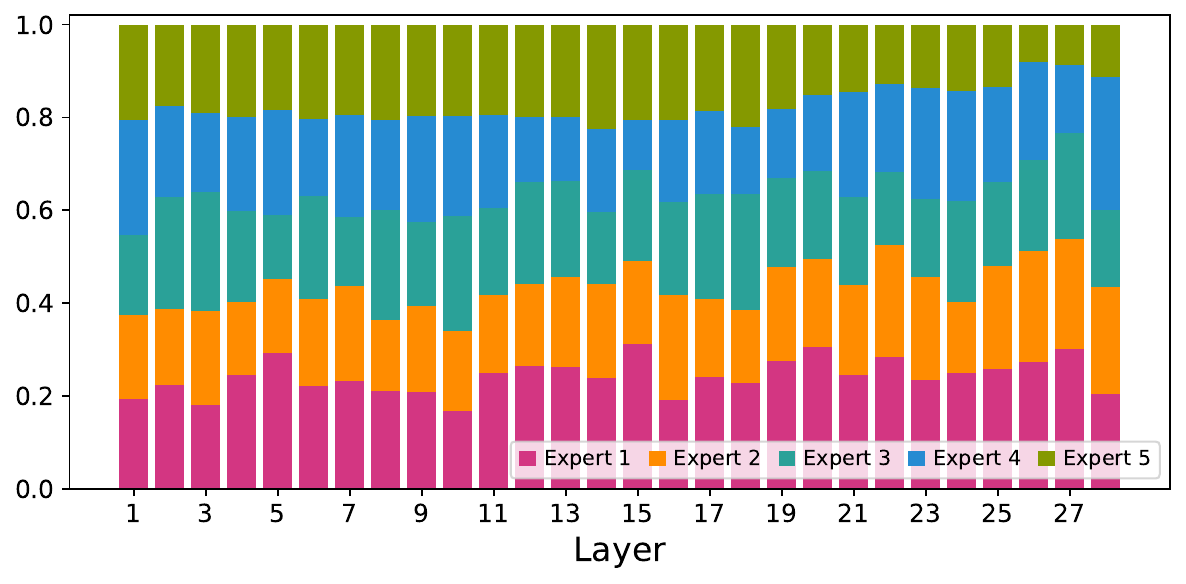}
    \caption{Audio Generation.}
    \label{fig:sub_audio_gen}
\end{subfigure}

% 整个图的总标题和标签
\caption{Analysis of expert activation proportion of Uni-MoE-2.0-Omni across different subtasks. The five experts shown in this figure are four routed experts (E1-E4, colored) and the null expert (E5, green). The vertical axis represents the proportion of tokens assigned to each expert at layer $i$ for the current task.}
\label{fig:expert_activation}
\end{figure*}
\begin{figure*}[htbp]
\centering % 将整个子图网格居中

% --- 第一行 ---
\begin{subfigure}[b]{0.24\textwidth}
    \centering
    \includegraphics[width=\textwidth]{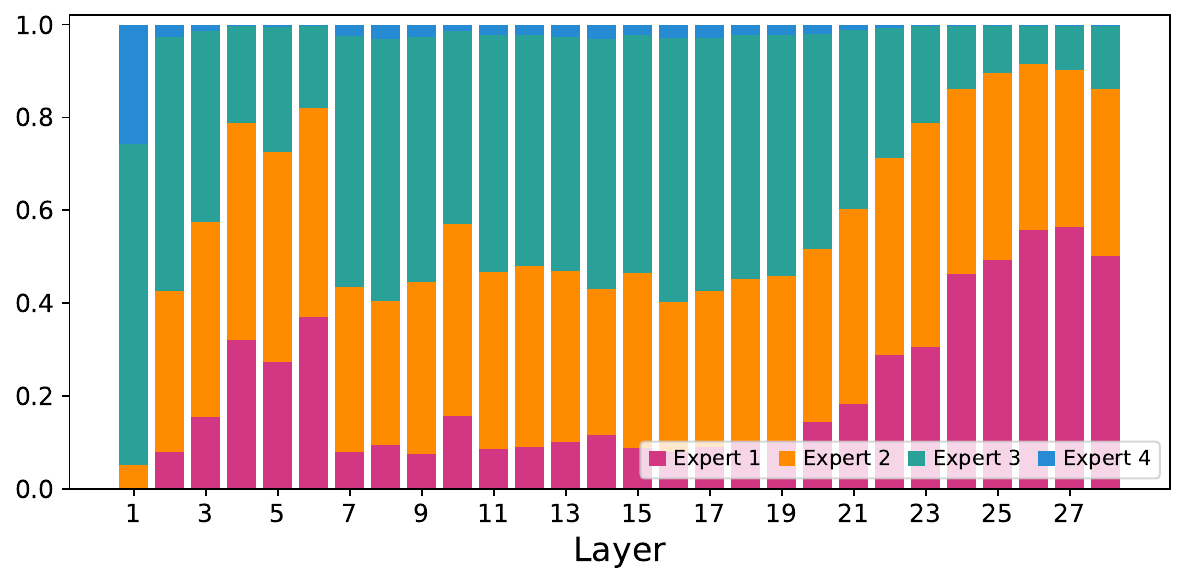}
    \caption{\textbf{Overall}}
    \label{fig:sub_overall_topk}
\end{subfigure}
\begin{subfigure}[b]{0.24\textwidth}
    \centering
    \includegraphics[width=\textwidth]{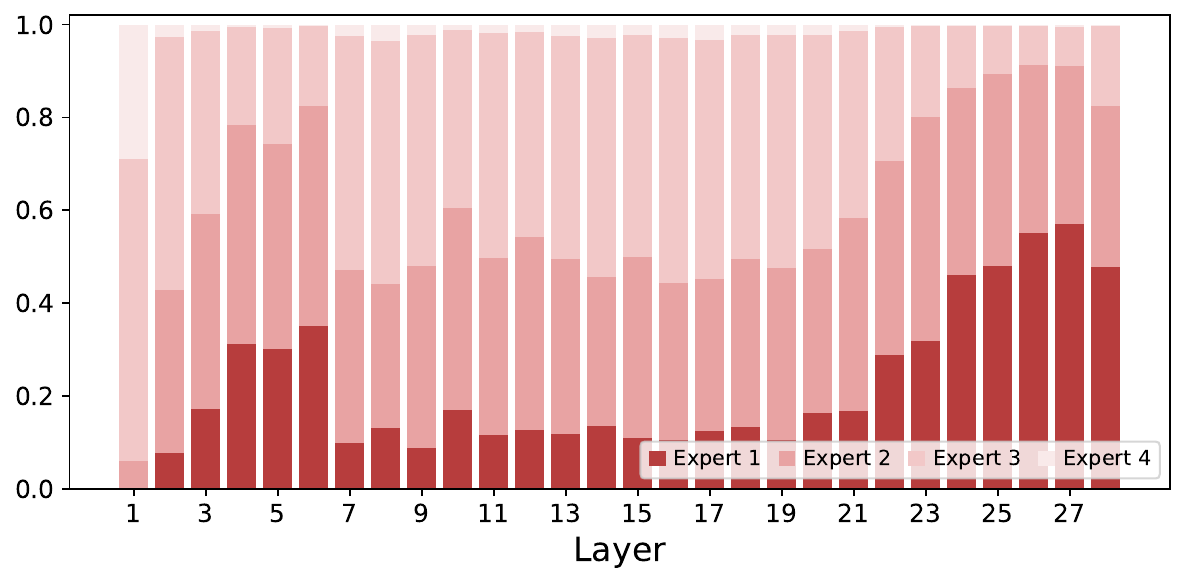}
    \caption{Image Understanding}
    \label{fig:sub_image_topk}
\end{subfigure}
\begin{subfigure}[b]{0.24\textwidth}
    \centering
    \includegraphics[width=\textwidth]{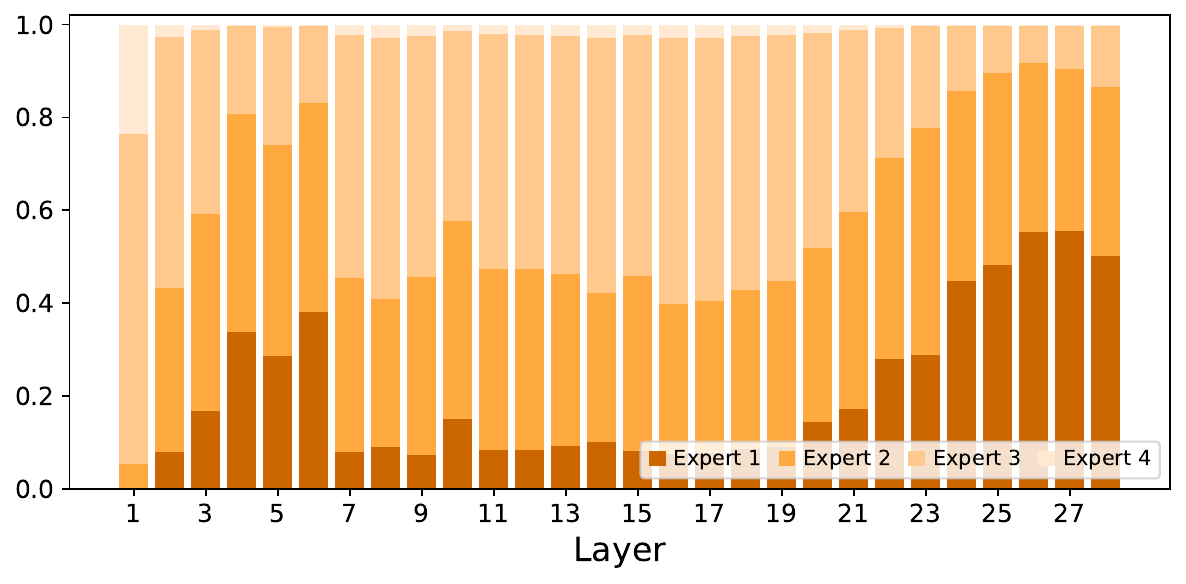}
    \caption{Video Understanding}
    \label{fig:sub_video_topk}
\end{subfigure}
\begin{subfigure}[b]{0.24\textwidth}
    \centering
    \includegraphics[width=\textwidth]{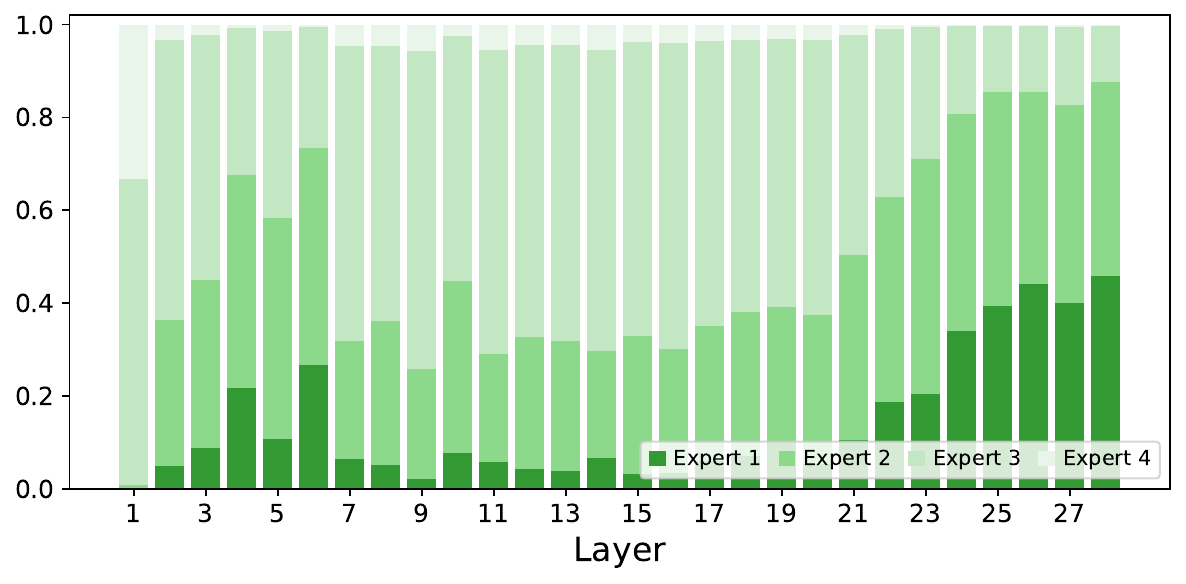}
    \caption{Audio Understanding}
    \label{fig:sub_audio_topk}
\end{subfigure}

\vspace{0.5cm} % 在两行之间添加一些垂直间距

% --- 第二行 ---
\begin{subfigure}[b]{0.24\textwidth}
    \centering
    \includegraphics[width=\textwidth]{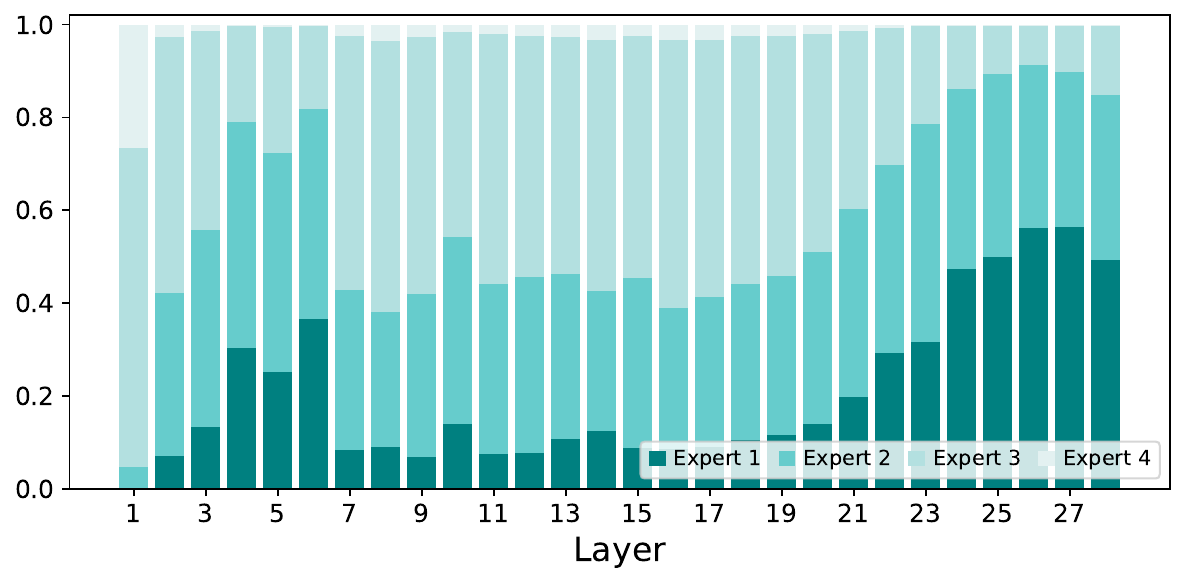}
    \caption{Omni Understanding}
    \label{fig:sub_omni_topk}
\end{subfigure}
\begin{subfigure}[b]{0.24\textwidth}
    \centering
    \includegraphics[width=\textwidth]{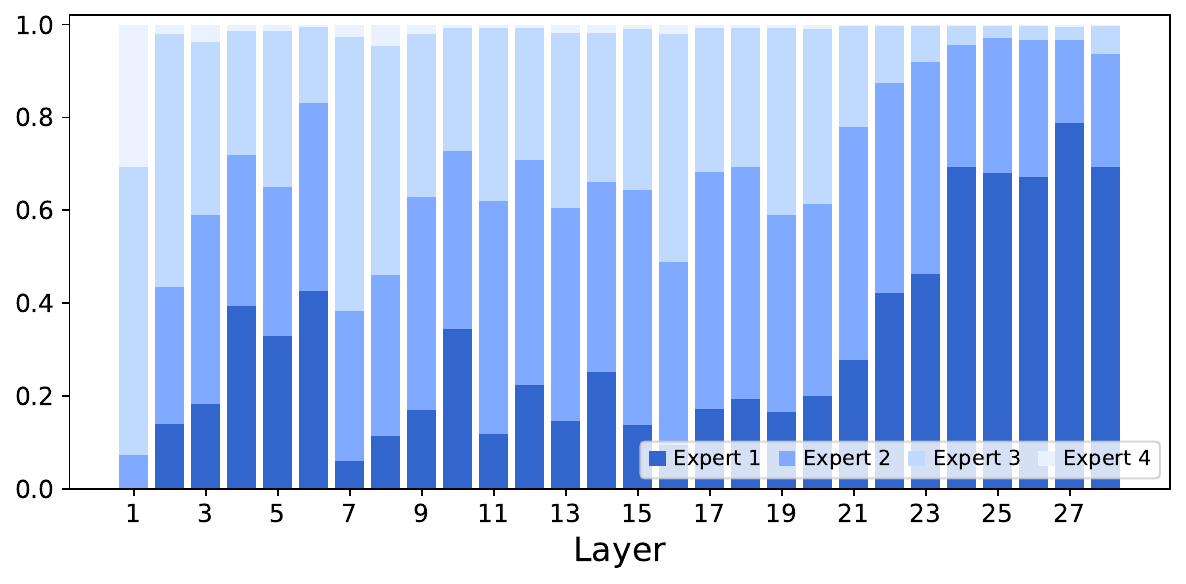}
    \caption{Image Generation}
    \label{fig:sub_image_gen_topk}
\end{subfigure}
\begin{subfigure}[b]{0.24\textwidth}
    \centering
    \includegraphics[width=\textwidth]{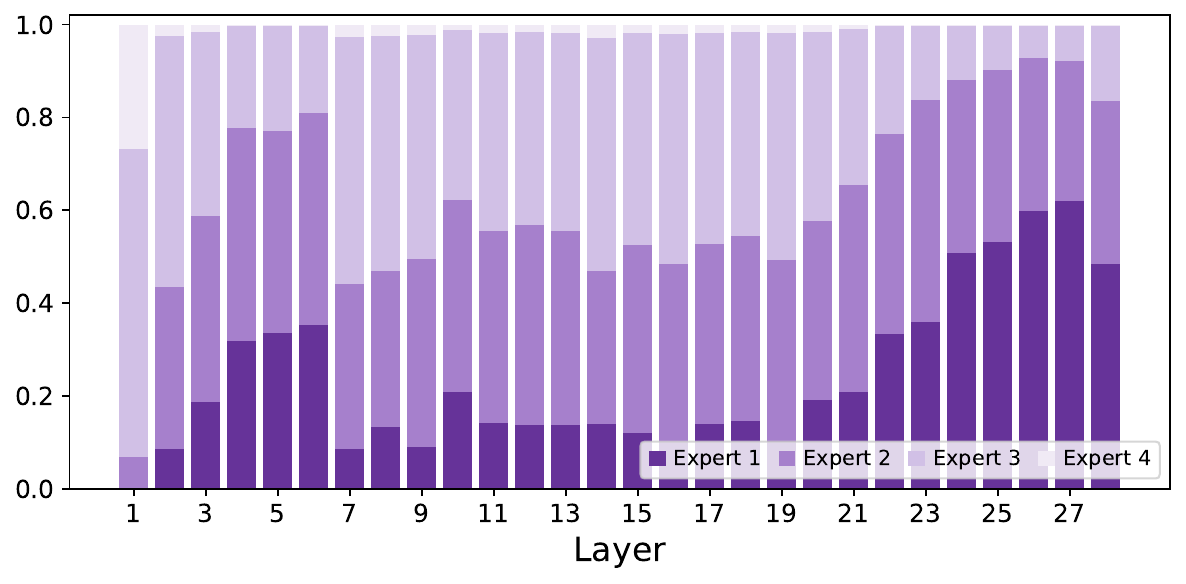}
    \caption{Image Edit}
    \label{fig:sub_image_edit_topk}
\end{subfigure}
\begin{subfigure}[b]{0.24\textwidth}
    \centering
    \includegraphics[width=\textwidth]{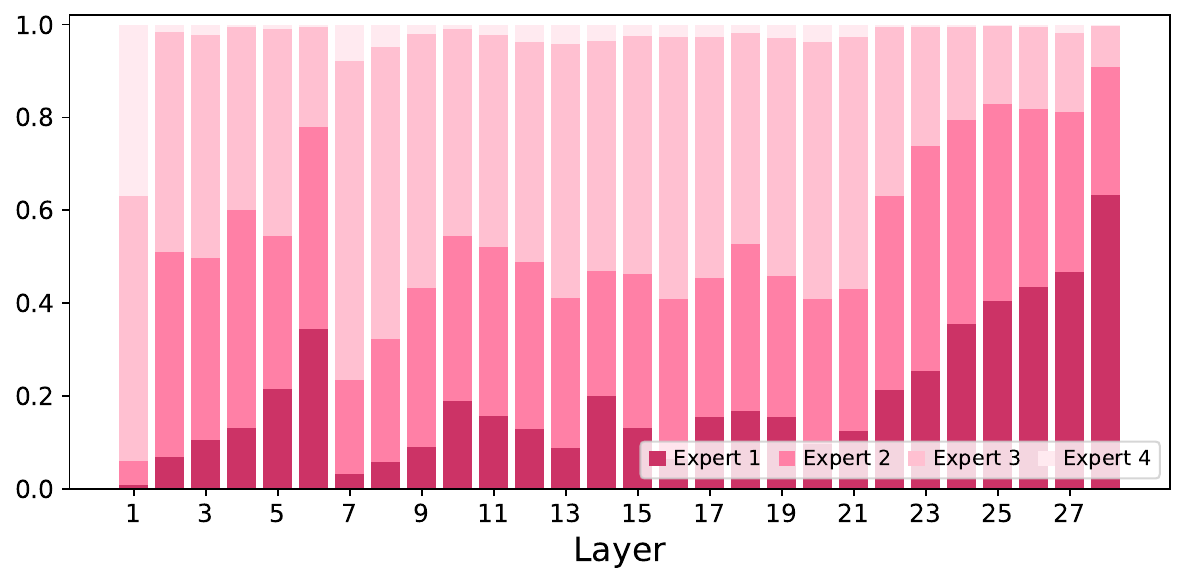}
    \caption{Audio Generation}
    \label{fig:sub_audio_gen_topk}
\end{subfigure}

% 整个图的总标题和标签
\caption{Visualization of the dynamic computational budget allocated by our Top-P routing mechanism. The figure illustrates the proportion of tokens activating a varying number of experts at each layer, revealing a ``peak-trough-peak-fall'' pattern where more computational resources are adaptively assigned to the middle layers. In this figure, "Expert 2" represents tokens activating either 2 experts.}
\label{fig:expert_topk}
\end{figure*}

\paragraph{Expert Activation}

To better understand the operational dynamics of our Dynamic-Capacity MoE, we visualise expert routing probabilities across layers for diverse tasks (Figure~\ref{fig:expert_activation}). The results reveal that initial layers exhibit nearly identical routing across tasks, indicating common set of experts for low-level feature extraction. In deeper layers, routing patterns diverge, reflecting task-specific specialization, while the probability of selecting the null expert (Expert 5) decreases sharply, suggesting that computation is increasingly concentrated in semantically critical stages. The activation profiles align with the intended expert roles: Expert 1 dominates vision-centric tasks (Image, Video and omnimodal Understanding), Experts 2 and 3 are prominent in audio-related tasks (Audio and omnimodal Understanding), and Expert 4 consistently contributes across tasks as a general-purpose semantic expert. Notably, understanding tasks show dynamic, layer-dependent routing, implying adaptive allocation of computational resources from concrete to abstract representations. In contrast, generation tasks exhibit stable, uniform routing across layers, consistent with a homogeneous refinement process. These observations confirm that our MoE design effectively promotes functional specialization and learns efficient, task-dependent computation pathways.

\paragraph{Computational Cost}

Analysis of the Top-P routing mechanism (Figure~\ref{fig:expert_topk}) shows a structured allocation of computational budget across layers, measured by the proportion of tokens activating different numbers of experts. The distribution follows a distinct peak–trough–peak–fall pattern: high load in early layers (1–3) for general-purpose feature extraction, a brief reduction in layers 4–6, a primary peak in middle-to-deep layers (7–21) corresponding to complex reasoning and feature integration, and a final decline in layers 21–27 as processing converges to output. While consistent across tasks, modality-specific differences emerge: temporal inputs such as Video and Audio exhibit a stronger initial peak than static Image, with more tokens activating multiple experts, indicating greater parallel resource needs for spatiotemporal processing. These results suggest that the routing mechanism learns both a global computational structure and fine-grained, modality-aware resource allocation.

\begin{figure*}[t]
\centering % 将整个子图网格居中

% --- 第一行 ---
\begin{subfigure}[b]{0.48\textwidth}
    \centering
    \includegraphics[width=\textwidth]{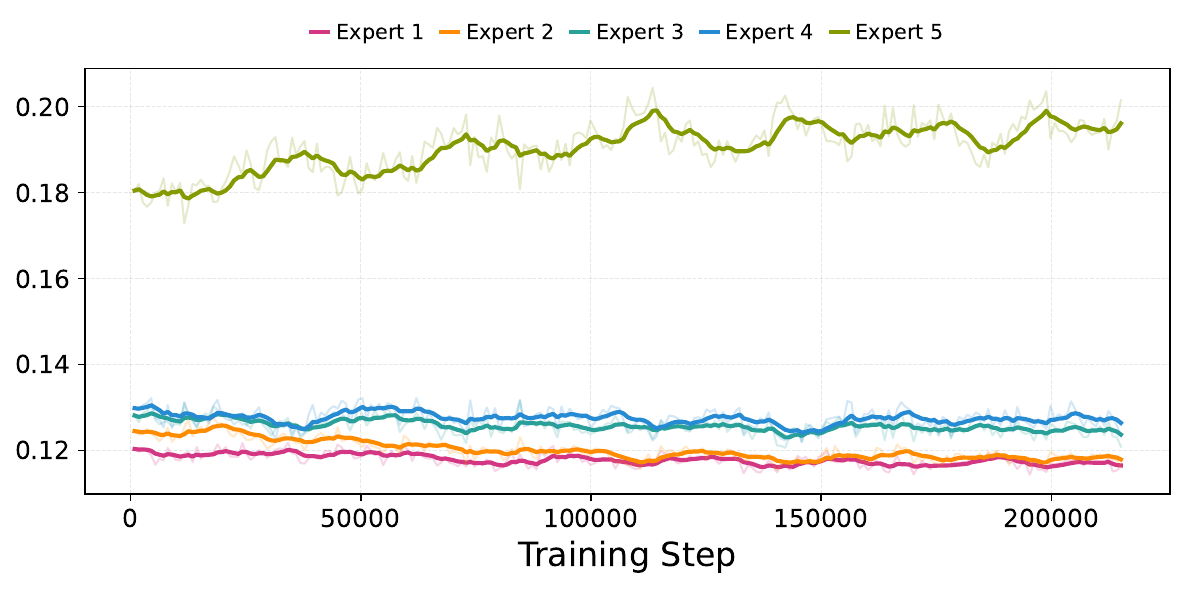}
    \caption{Layer0}
    \label{fig:sub_layer0}
\end{subfigure}
\begin{subfigure}[b]{0.48\textwidth}
    \centering
    \includegraphics[width=\textwidth]{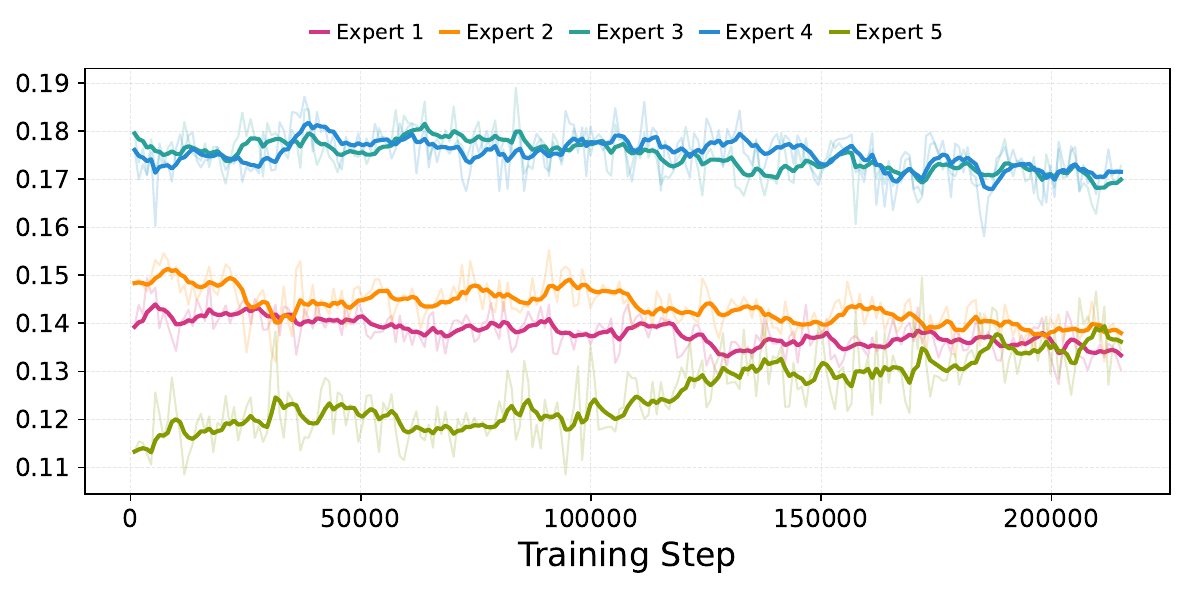}
    \caption{Layer9}
    \label{fig:sub_layer9}
\end{subfigure}

\vspace{0.3cm} % 在两行之间添加一些垂直间距

% --- 第二行 ---
\begin{subfigure}[b]{0.48\textwidth}
    \centering
    \includegraphics[width=\textwidth]{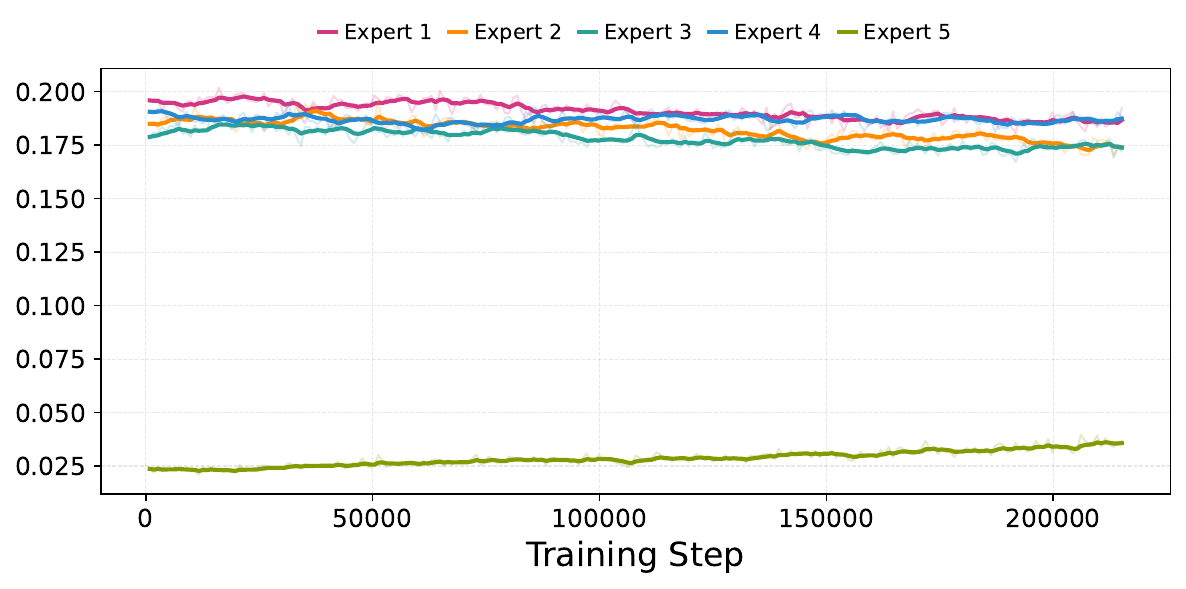}
    \caption{Layer18}
    \label{fig:sub_layer18}
\end{subfigure}
\begin{subfigure}[b]{0.48\textwidth}
    \centering
    \includegraphics[width=\textwidth]{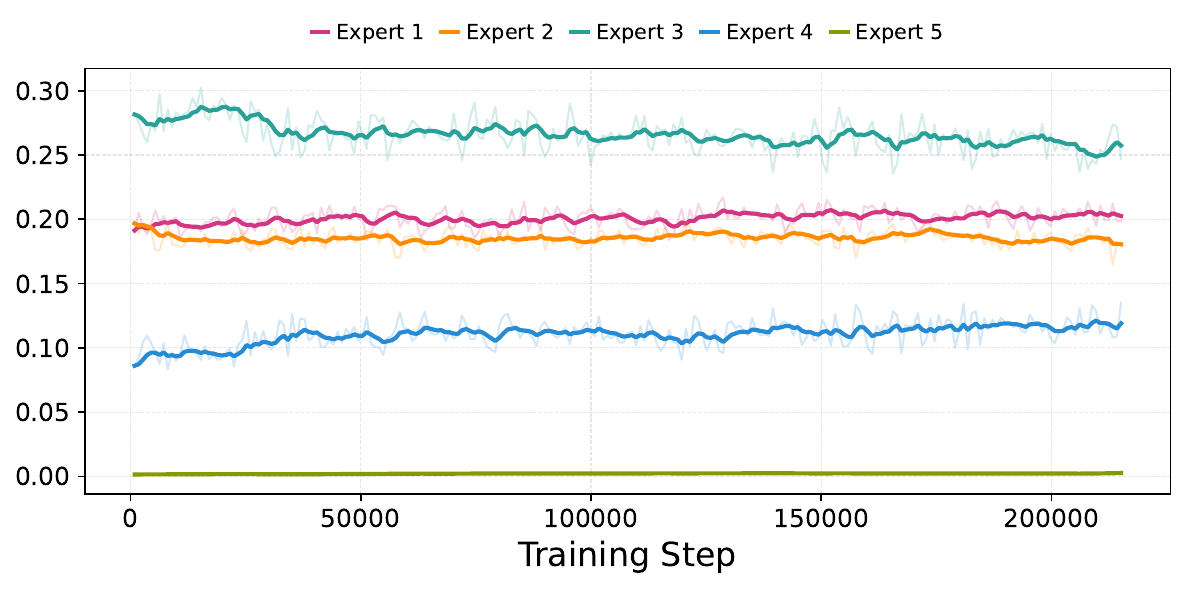}
    \caption{Layer27}
    \label{fig:sub_layer27}
\end{subfigure}

% 整个图的总标题和标签
\caption{Analysis of expert activation proportion of Uni-MoE 2.0 during annealing training steps. The five experts shown in this figure are four routed experts (E1-E4, colored) and the null expert (E5, green).}
\label{fig:expert_training_activation}
\end{figure*}

\paragraph{Routing Dynamics}

We tracked expert activation at four representative layers (Figure~\ref{fig:expert_training_activation}). Expert usage remains stable in the shallowest (Layer 0) and deepest (Layer 27) layers, while middle layers (Layers 9 and 18) show clear changes during training. This indicates that most refinement of expert specialization happens in the middle of the network, where both routing patterns and expert parameters are actively optimized. A key trend in these middle layers is the steady increase in activation of the null expert (Expert 5), meaning the model learns to skip tokens that no longer need processing at intermediate stages. This behavior confirms the value of the null expert for efficient computation and shows that our training strategy effectively improves inference efficiency.

% We tracked the evolution of expert activation proportions at four representative layers, as depicted in Figure~\ref{fig:expert_training_activation}. The analysis reveals that the refinement of expert specialization is primarily concentrated in the middle layers of the network. In the shallowest layer (Layer 0, Figure~\ref{fig:sub_layer0}) and the deepest layer (Layer 27, Figure~\ref{fig:sub_layer27}), the activation proportions for all experts remain remarkably stable throughout the training process. In stark contrast, the middle layers (Layer 9, Figure~\ref{fig:sub_layer9} and Layer 18,  Figure~\ref{fig:sub_layer18}) exhibit significant dynamic adjustments, where the relative utilization of experts continues to evolve. This contrast strongly suggests that the middle layers are the primary locus of optimization during annealing, where the model actively refines both its computational pathways and the specialized parameters of the experts.

% The most notable trend within these dynamic middle layers is the steady, monotonic increase in the activation proportion of the null expert (Expert 5). This indicates that as training progresses, the model learns to more effectively identify tokens that do not require further computation at these intermediate stages, routing them to be skipped. This learned behavior not only validates the utility of incorporating a null expert for creating more efficient computational pathways but also demonstrates that our training strategy successfully guides the model towards greater inference efficiency.

\subsection{Thinking vs. No-Thinking}

\begin{table*}[t]
    \centering
    \scriptsize
    \small
    \renewcommand{\arraystretch}{1.0}
    \setlength{\tabcolsep}{3pt} % 减小列间距
    \resizebox{0.9\textwidth}{!}{
    \begin{tabular}{lcccccccccc}
    \toprule
    \textbf{Method} & MathVista (testmini) & MathVerse (vision) & LogicVista (testmini) & MMMU (val) & {Avg.} \\
    \midrule
    \rowcolor{SkyBlue!15}
    \multicolumn{6}{l}{\textbf{Non-Thinking}} \\
    %MiniCPM-o 2.6 & 66.20 & - & 37.05 & 47.33 & - \\
    % Qwen2.5-Omni-7B & 56.20 & 25.63 & 33.93 & 44.44 & 40.05 \\
    % Ming-Lite-Omni & 69.50 & 27.41 & 34.38 & 51.78 & 45.77\\
    %Ming-Lite-Omni-1.5 & 69.00 & 25.25 & 37.05 & 53.44 & 46.19\\
    Uni-MoE-2.0 & 60.80 & 17.26 & 31.47 & 42.67 & 38.05\\
    Uni-MoE-2.0-Base & \underline{61.30} & 18.15 & 32.81 & \underline{46.67} & 39.73\\
    \midrule
    \rowcolor{SkyBlue!15}
    \multicolumn{6}{l}{\textbf{Thinking}} \\
    Uni-MoE-2.0-ColdStart & 55.50 & 19.54 & 28.35 & 39.67 & 35.77 \\
    Uni-MoE-2.0-GSPO & 58.90 & \underline{21.19} & \underline{33.71} & \textbf{47.11} & \underline{40.23} \\
    Uni-MoE-2.0-DPO & \textbf{63.90} & \textbf{22.97} & \textbf{34.82} & 45.78 & \textbf{41.87}\\
    \bottomrule
    \end{tabular}
    }
    \caption{
        \textbf{Comparison of Uni-MoE-2.0-Thinking and Base (non-thinking) version.}
         \textbf{Bold} indicates the highest score, and \underline{underline} indicates the second-highest score for each benchmark.
    }
    \label{tab:results_thinking}
\end{table*}

\subsubsection{Visual Reasoning}

In Table~\ref{tab:results_thinking}, we present the performance of Uni-MoE-2.0 models at various reinforcement learning stages, as well as their comparison with the original direct-answer models. It can be observed that a significant performance drop in the Uni-MoE-2.0-Cold Start model. We attribute this to the limited amount of Cold-Start data used, which contained complex and diverse tasks that differed substantially from the main evaluation focus on mathematical and scientific reasoning, resulting in a noticeable decline in generalization capability.

After continuing GSPO training based on the Cold-Start model, we find that Uni-MoE-2.0-GSPO shows performance recovery across all evaluation metrics, even surpassing the original Uni-MoE-2.0 model in MathVerse, LogicVista, and MMMU. In particular, MathVerse performance increased by 3.04\%.

Finally, after further applying DPO training on top of Uni-MoE-2.0-GSPO, the model exhibits a clear upward trend in performance, especially achieving a 5\% improvement on MathVista (testmini) and an average gain of 1.64\% across all evaluation metrics. This demonstrates the effectiveness of the DPO stage, showing that using a small amount of DPO data annotated by powerful commercial models can significantly enhance the model’s reasoning ability.

\begin{figure}[t]
    \centering
    \includegraphics[width=0.9\linewidth]{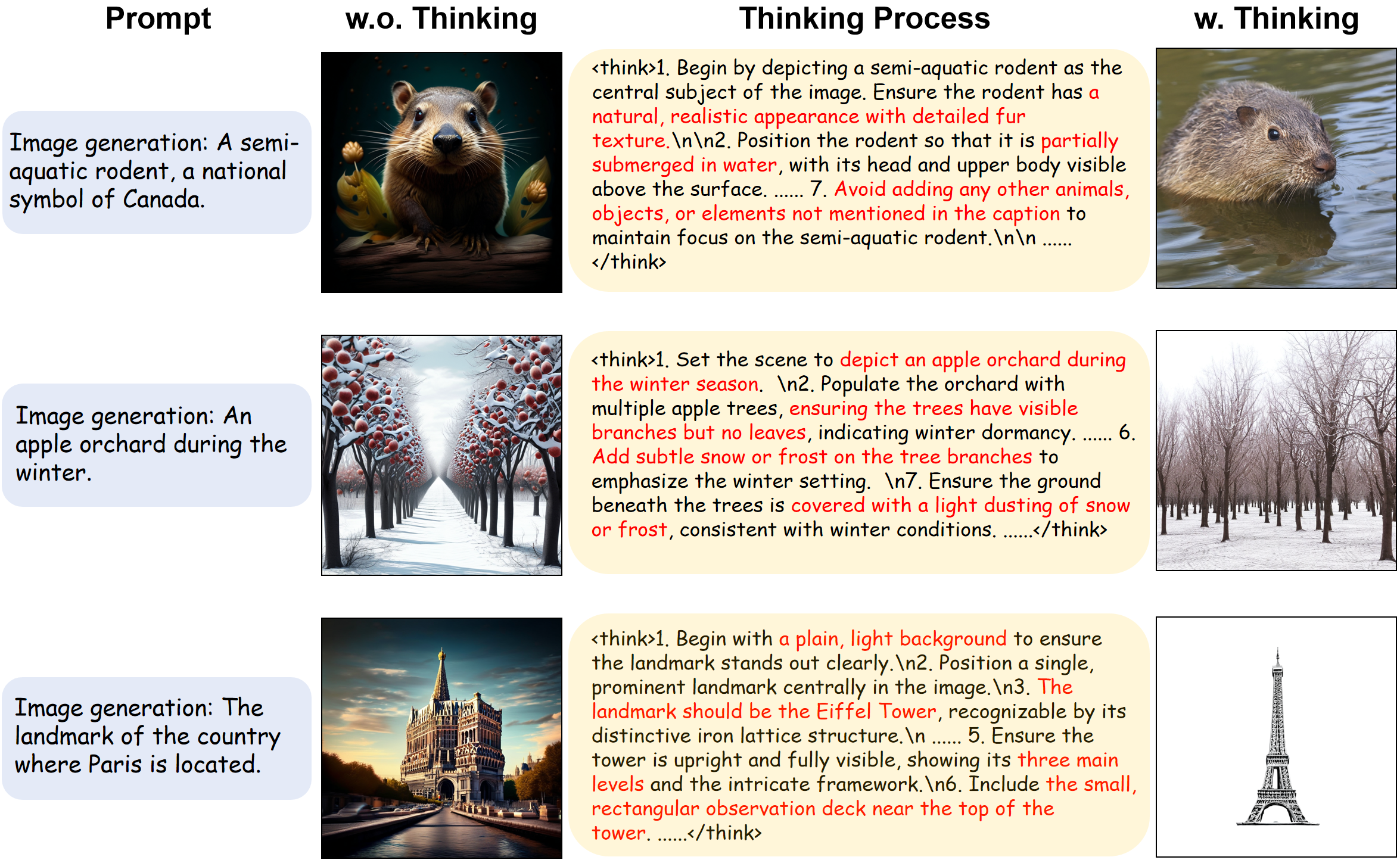}
    \caption{Comparison of image generation results with and without thinking guidance. The “w.o. Thinking” column shows images generated directly from prompts, while the “w. Thinking” column illustrates results produced after incorporating step-by-step reasoning. The middle column presents the structured thinking process guiding the model toward more accurate and contextually faithful image synthesis.}
    \label{fig:image_gen_thinking}
\end{figure}

\subsubsection{Visual Generation}

Figure~\ref{fig:image_gen_thinking} demonstrates that integrating a structured thinking process markedly improves the faithfulness and coherence of generated images. The baseline model, lacking explicit reasoning, frequently generates semantically inconsistent visuals (e.g., a rodent in an arid habitat or apple trees bearing fruit in winter). Conversely, when guided by a step-by-step reasoning chain, the model successfully parses prompts into constituent visual elements—such as texture, environment, and seasonal context—resulting in outputs that are precisely aligned with the prompt's semantic intent. Unlike Qwen2.5-Omni and Ming-Lite-Omni, our model incorporates a thinking chain for image generation, similar to the approach used in BAGEL.

\begin{TakeawayBox}{Takeaways: Thinking}
Our method of integrating a structured ``thinking'' process significantly enhances model performance. It boosts complex reasoning in multimodal understanding tasks with the GSPO-DPO training strategy and ensures faithful, coherent outputs in text-to-image generation tasks.
\end{TakeawayBox}

\section{Discussion and Future Work}
\label{sec:dicussion_future}

\paragraph{Audio Understanding and Generating} Our investigation reveals two key principles for token-based TTS: a dual-rate token strategy and strategic architectural scaling. We find that while a compact 20 tokens per second suffices for audio representation, generation requires 40 tokens per second to capture fine-grained acoustic details. Furthermore, smaller (0.5B) autoregressive models struggle with style conversion, a limitation we address by adopting a MoE architecture. For these models, training on token embeddings from a pre-trained base model proves more effective for capturing vocal style than end-to-end training. Together, these approaches form an effective framework for building efficient and expressive neural spoken systems. Building on these insights, our future work will follow Uni-MoE-Audio, using a single tokenizer for both understanding and generation tasks. This foundation will enable cross-modal training with text and video, facilitating context-aware and voice-preserved applications. Concurrently, we will optimize our MoE architecture with conditional routing for more efficient and controllable multi-speaker synthesis.

\paragraph{Image Generating and Editing} To facilitate collaborative development, we decoupled the image generation module from the base model. This architectural choice not only simplifies joint training but also enables flexible image generation control through natural language and specialized tokens. While the current text-to-image performance in automated metrics remains limited—due to the constrained version of our external diffusion model and the scarcity of text–image paired data—we observe notable improvements in image editing and low-level image processing tasks. These gains suggest the viability of a language-centric training strategy for integrating diverse image processing capabilities.
In future work, we will further refine the architecture for image editing and generation, allowing the base model to perceive and interpret such tasks during pre-training. Nevertheless, the iterative diffusion process for image synthesis and editing will remain externalized. This design reflects our view that multi-step temporal modelling (as in diffusion) operates at a fundamentally different temporal granularity compared to single-step next-token prediction.

\paragraph{Omnimodality Understanding} We identify that the model's enhanced omnimodality understanding capability stems primarily from its audio-text-vision joint coding training on large-scale video data. Future work will focus on scaling this video data and introducing new multimodal positional encoding methods to advance the model's comprehension abilities further.

\paragraph{Model Architecture} Our analysis indicates that a dense model can be directly extended to an MoE architecture. This approach enables the model to achieve more comprehensive capabilities even with limited additional data (75B tokens).
Building on this, our future work will focus on optimizing expert specialization and leveraging large-scale dense model distillation. This strategy will allow us to efficiently construct a unified, fine-grained MoE-based model for omnimodal understanding and generation.

\paragraph{Training and Data Recipe} Our training approach, which builds upon the progressive strategy from Uni-MoE 1.0, was augmented in version 2.0 with annealing experiments and reinforcement learning. Comparative results demonstrate that the omnimodal large model's training is highly sensitive to data recipe, yet our progressive training strategy ensured remarkable stability during the RL phase for MoE-based models. Moving forward, we will refine this iterative RL strategy by introducing distinct RL methods at various training stages to enhance model capabilities periodically.

\section{Conclusion}
\label{sec:conclusion}

In this paper, Uni-MoE-2.0-Omni represents a significant advancement in the development of open-source, omnimodal large models. By building upon the dense Qwen2.5-7B architecture and introducing key innovations—a dynamic-capacity MoE design, a progressive training strategy enhanced with iterative reinforcement learning, and a curated multimodal data matching technique—the model achieves robust capabilities in understanding, reasoning, and generating across text, image, and speech modalities. Notably, we explored a progressive model architecture evolution and training strategy optimization, which can extend dense large language models into efficient MoE-based omnimodal large models, achieving a leap from multimodal understanding to both understanding and generation.
Extensive evaluations across 85 benchmarks confirm that Uni-MoE-2.0-Omni sets a new state-of-the-art or is highly competitive with leading omnimodal large models, demonstrating particular strengths in video understanding, omnimodal comprehension, long speech understanding and generating, audio-visual, controllable image generation, and low-level image restoration tasks. The commitment to open-sourcing the model's code, checkpoints, and data ensures transparency and fosters further innovation in the field of multimodal artificial intelligence.

\section{Acknowledgment}

We would like to express our sincere gratitude to Qi Wang, Qixun Teng, Feifan Wen, and Zitao Li for their contributions to data collection and demo recording. We are thankful to Jinchao Li and Tongshu Bian for their assistance in paper proofreading and video production. We also thank Longyue Wang and Wenhan Luo for their valuable suggestions on paper writing.

\section{Contributors}
\label{sec:contributors}

\textbf{Project Leaders}

Min Zhang, Baotian Hu 

% Yunxin Li, Baotian Hu, Min Zhang

\textbf{Contributors} ($^*$ Core Contributions)

Yunxin Li$^*$, Xinyu Chen$^*$, Shenyuan Jiang$^*$, Haoyuan Shi$^*$, Zhenyu Liu$^*$, Xuanyu Zhang, Nanhao Deng, Zhenran Xu, Yicheng Ma, Meishan Zhang

% \textbf{Contributors}

% Yunxin Li, Xinyu Chen, Shenyuan Jiang, Haoyuan Shi, Zhenyu Liu, \\ Xuanyu Zhang, Nanhao Deng,  Zhenran Xu, Yicheng Ma, Meishan Zhang, Baotian Hu, Min Zhang

\textbf{Corresponding Authors}

Baotian Hu

Harbin Institute of Technology, Shenzhen

Email: hubaotian@hit.edu.cn

Min Zhang

Harbin Institute of Technology, Shenzhen

Email: zhangmin2021@hit.edu.cn

% \textbf{Contacts}

% Yunxin Li: liyunxin987@163.com

% Baotian Hu: hubaotian@hit.edu.cn

\newpage

\bibliographystyle{lychee}
\bibliography{custom}

@inproceedings{zhai2023sigmoid,
  title={Sigmoid loss for language image pre-training},
  author={Zhai, Xiaohua and Mustafa, Basil and Kolesnikov, Alexander and Beyer, Lucas},
  booktitle={Proceedings of the IEEE/CVF international conference on computer vision},
  pages={11975--11986},
  year={2023}
}

@article{Qwen2-VL,
  title={Qwen2-VL: Enhancing Vision-Language Model's Perception of the World at Any Resolution},
  author={Wang, Peng and Bai, Shuai and Tan, Sinan and Wang, Shijie and Fan, Zhihao and Bai, Jinze and Chen, Keqin and Liu, Xuejing and Wang, Jialin and Ge, Wenbin and Fan, Yang and Dang, Kai and Du, Mengfei and Ren, Xuancheng and Men, Rui and Liu, Dayiheng and Zhou, Chang and Zhou, Jingren and Lin, Junyang},
  journal={arXiv preprint arXiv:2409.12191},
  year={2024}
}

@article{qwen_omni,
  author       = {Jin Xu and
                  Zhifang Guo and
                  Jinzheng He and
                  Hangrui Hu and
                  Ting He and
                  Shuai Bai and
                  Keqin Chen and
                  Jialin Wang and
                  Yang Fan and
                  Kai Dang and
                  Bin Zhang and
                  Xiong Wang and
                  Yunfei Chu and
                  Junyang Lin},
  title        = {Qwen2.5-Omni Technical Report},
  journal      = {CoRR},
  volume       = {abs/2503.20215},
  year         = {2025},
  url          = {https://doi.org/10.48550/arXiv.2503.20215},
  doi          = {10.48550/ARXIV.2503.20215},
  eprinttype    = {arXiv},
  eprint       = {2503.20215},
  bibsource    = {dblp computer science bibliography, https://dblp.org}
}

@inproceedings{DBLP:conf/wacv/CuiMCYZLCLYLGLTCZLYMCWZ24,
  author       = {Can Cui and
                  Yunsheng Ma and
                  Xu Cao and
                  Wenqian Ye and
                  Yang Zhou and
                  Kaizhao Liang and
                  Jintai Chen and
                  Juanwu Lu and
                  Zichong Yang and
                  Kuei{-}Da Liao and
                  Tianren Gao and
                  Erlong Li and
                  Kun Tang and
                  Zhipeng Cao and
                  Tong Zhou and
                  Ao Liu and
                  Xinrui Yan and
                  Shuqi Mei and
                  Jianguo Cao and
                  Ziran Wang and
                  Chao Zheng},
  title        = {A Survey on Multimodal Large Language Models for Autonomous Driving},
  booktitle    = {{WACV} (Workshops)},
  pages        = {958--979},
  publisher    = {{IEEE}},
  year         = {2024}
}

@article{DBLP:journals/corr/abs-2505-09777,
  author       = {Alejo Lopez{-}Avila and
                  Jinhua Du},
  title        = {A Survey on Large Language Models in Multimodal Recommender Systems},
  journal      = {CoRR},
  volume       = {abs/2505.09777},
  year         = {2025}
}

@article{DBLP:journals/corr/abs-2306-13549,
  author       = {Shukang Yin and
                  Chaoyou Fu and
                  Sirui Zhao and
                  Ke Li and
                  Xing Sun and
                  Tong Xu and
                  Enhong Chen},
  title        = {A Survey on Multimodal Large Language Models},
  journal      = {CoRR},
  volume       = {abs/2306.13549},
  year         = {2023}
}

@ARTICLE{li_unimoe,
  author={Li, Yunxin and Jiang, Shenyuan and Hu, Baotian and Wang, Longyue and Zhong, Wanqi and Luo, Wenhan and Ma, Lin and Zhang, Min},
  journal={IEEE Transactions on Pattern Analysis and Machine Intelligence}, 
  title={Uni-MoE: Scaling Unified Multimodal LLMs With Mixture of Experts}, 
  year={2025},
  volume={47},
  number={5},
  pages={3424-3439},
  doi={10.1109/TPAMI.2025.3532688}}

@article{DBLP:journals/corr/abs-2307-10802,
  author       = {Yiyuan Zhang and
                  Kaixiong Gong and
                  Kaipeng Zhang and
                  Hongsheng Li and
                  Yu Qiao and
                  Wanli Ouyang and
                  Xiangyu Yue},
  title        = {Meta-Transformer: {A} Unified Framework for Multimodal Learning},
  journal      = {CoRR},
  volume       = {abs/2307.10802},
  year         = {2023}
}

@inproceedings{DBLP:conf/iclr/XieMBZWLGCYS25,
  author       = {Jinheng Xie and
                  Weijia Mao and
                  Zechen Bai and
                  David Junhao Zhang and
                  Weihao Wang and
                  Kevin Qinghong Lin and
                  Yuchao Gu and
                  Zhijie Chen and
                  Zhenheng Yang and
                  Mike Zheng Shou},
  title        = {Show-o: One Single Transformer to Unify Multimodal Understanding and
                  Generation},
  booktitle    = {{ICLR}},
  publisher    = {OpenReview.net},
  year         = {2025}
}

@article{ming_omni,
  author       = {Inclusion AI and
                  Biao Gong and
                  Cheng Zou and
                  Chuanyang Zheng and
                  Chunluan Zhou and
                  Canxiang Yan and
                  Chunxiang Jin and
                  Chunjie Shen and
                  Dandan Zheng and
                  Fudong Wang and
                  Furong Xu and
                  Guangming Yao and
                  Jun Zhou and
                  Jingdong Chen and
                  Jianxin Sun and
                  et al},
  title        = {Ming-Omni: {A} Unified Multimodal Model for Perception and Generation},
  journal      = {CoRR},
  volume       = {abs/2506.09344},
  year         = {2025},
  url          = {https://doi.org/10.48550/arXiv.2506.09344},
  doi          = {10.48550/ARXIV.2506.09344},
  eprinttype    = {arXiv},
  eprint       = {2506.09344},
  bibsource    = {dblp computer science bibliography, https://dblp.org}
}

@article{gemini25,
  author       = {Gheorghe Comanici and
                  Eric Bieber and
                  Mike Schaekermann and
                  Ice Pasupat and
                  Noveen Sachdeva and
                  Inderjit S. Dhillon and
                  Marcel Blistein and
                  Ori Ram and
                  Dan Zhang and
                  Evan Rosen and
                  Luke Marris and
                  Sam Petulla and
                  Colin Gaffney and
                  Asaf Aharoni and
                  Nathan Lintz and
                  Tiago Cardal Pais and
                  Henrik Jacobsson and
                  Idan Szpektor and
                  Nan{-}Jiang Jiang and
                  Krishna Haridasan and
                  et al},
  title        = {Gemini 2.5: Pushing the Frontier with Advanced Reasoning, Multimodality,
                  Long Context, and Next Generation Agentic Capabilities},
  journal      = {CoRR},
  volume       = {abs/2507.06261},
  year         = {2025},
  url          = {https://doi.org/10.48550/arXiv.2507.06261},
  doi          = {10.48550/ARXIV.2507.06261},
  eprinttype    = {arXiv},
  eprint       = {2507.06261},
  bibsource    = {dblp computer science bibliography, https://dblp.org}
}

@inproceedings{DBLP:conf/nips/SchickDDRLHZCS23,
  author       = {Timo Schick and
                  Jane Dwivedi{-}Yu and
                  Roberto Dess{\`{\i}} and
                  Roberta Raileanu and
                  Maria Lomeli and
                  Eric Hambro and
                  Luke Zettlemoyer and
                  Nicola Cancedda and
                  Thomas Scialom},
  title        = {Toolformer: Language Models Can Teach Themselves to Use Tools},
  booktitle    = {NeurIPS},
  year         = {2023}
}

@article{DBLP:journals/corr/abs-2501-12326,
  author       = {Yujia Qin and
                  Yining Ye and
                  Junjie Fang and
                  Haoming Wang and
                  Shihao Liang and
                  Shizuo Tian and
                  Junda Zhang and
                  Jiahao Li and
                  Yunxin Li and
                  Shijue Huang and
                  Wanjun Zhong and
                  Kuanye Li and
                  Jiale Yang and
                  Yu Miao and
                  Woyu Lin and
                  Longxiang Liu and
                  Xu Jiang and
                  Qianli Ma and
                  Jingyu Li and
                  Xiaojun Xiao and
                  Kai Cai and
                  Chuang Li and
                  Yaowei Zheng and
                  Chaolin Jin and
                  Chen Li and
                  Xiao Zhou and
                  Minchao Wang and
                  Haoli Chen and
                  Zhaojian Li and
                  Haihua Yang and
                  Haifeng Liu and
                  Feng Lin and
                  Tao Peng and
                  Xin Liu and
                  Guang Shi},
  title        = {{UI-TARS:} Pioneering Automated {GUI} Interaction with Native Agents},
  journal      = {CoRR},
  volume       = {abs/2501.12326},
  year         = {2025}
}

@inproceedings{DBLP:conf/nips/XieZCLZCHCSLLXZ24,
  author       = {Tianbao Xie and
                  Danyang Zhang and
                  Jixuan Chen and
                  Xiaochuan Li and
                  Siheng Zhao and
                  Ruisheng Cao and
                  Toh Jing Hua and
                  Zhoujun Cheng and
                  Dongchan Shin and
                  Fangyu Lei and
                  Yitao Liu and
                  Yiheng Xu and
                  Shuyan Zhou and
                  Silvio Savarese and
                  Caiming Xiong and
                  Victor Zhong and
                  Tao Yu},
  title        = {OSWorld: Benchmarking Multimodal Agents for Open-Ended Tasks in Real
                  Computer Environments},
  booktitle    = {NeurIPS},
  year         = {2024}
}

@article{gpt4o_system,
  author       = {Aaron Hurst and
                  Adam Lerer and
                  Adam P. Goucher and
                  Adam Perelman and
                  Aditya Ramesh and
                  Aidan Clark and
                  AJ Ostrow and
                  Akila Welihinda and
                  Alan Hayes and
                  Alec Radford and
                  Aleksander Madry and
                  Alex Baker{-}Whitcomb and
                  Alex Beutel and
                  Alex Borzunov and
                  Alex Carney and
                  Alex Chow and
                  Alex Kirillov and
                  Alex Nichol and
                  Alex Paino and
                  Alex Renzin and
                  et al},
  title        = {GPT-4o System Card},
  journal      = {CoRR},
  volume       = {abs/2410.21276},
  year         = {2024},
  url          = {https://doi.org/10.48550/arXiv.2410.21276},
  doi          = {10.48550/ARXIV.2410.21276},
  eprinttype    = {arXiv},
  eprint       = {2410.21276},
  bibsource    = {dblp computer science bibliography, https://dblp.org}
}

@article{DBLP:journals/corr/abs-2501-17811,
  author       = {Xiaokang Chen and
                  Zhiyu Wu and
                  Xingchao Liu and
                  Zizheng Pan and
                  Wen Liu and
                  Zhenda Xie and
                  Xingkai Yu and
                  Chong Ruan},
  title        = {Janus-Pro: Unified Multimodal Understanding and Generation with Data
                  and Model Scaling},
  journal      = {CoRR},
  volume       = {abs/2501.17811},
  year         = {2025}
}

@article{deng2025bagel,
  title   = {Emerging Properties in Unified Multimodal Pretraining},
  author  = {Deng, Chaorui and Zhu, Deyao and Li, Kunchang and Gou, Chenhui and Li, Feng and Wang, Zeyu and Zhong, Shu and Yu, Weihao and Nie, Xiaonan and Song, Ziang and Shi, Guang and Fan, Haoqi},
  journal = {arXiv preprint arXiv:2505.14683},
  year    = {2025}
}

@article{DBLP:journals/corr/abs-2506-18871,
  author       = {Chenyuan Wu and
                  Pengfei Zheng and
                  Ruiran Yan and
                  Shitao Xiao and
                  Xin Luo and
                  Yueze Wang and
                  Wanli Li and
                  Xiyan Jiang and
                  Yexin Liu and
                  Junjie Zhou and
                  Ze Liu and
                  Ziyi Xia and
                  Chaofan Li and
                  Haoge Deng and
                  Jiahao Wang and
                  Kun Luo and
                  Bo Zhang and
                  Defu Lian and
                  Xinlong Wang and
                  Zhongyuan Wang and
                  Tiejun Huang and
                  Zheng Liu},
  title        = {OmniGen2: Exploration to Advanced Multimodal Generation},
  journal      = {CoRR},
  volume       = {abs/2506.18871},
  year         = {2025}
}

@misc{liu2024gringradientinformedmoe,
      title={GRIN: GRadient-INformed MoE}, 
      author={Liyuan Liu and Young Jin Kim and Shuohang Wang and Chen Liang and Yelong Shen and Hao Cheng and Xiaodong Liu and Masahiro Tanaka and Xiaoxia Wu and Wenxiang Hu and Vishrav Chaudhary and Zeqi Lin and Chenruidong Zhang and Jilong Xue and Hany Awadalla and Jianfeng Gao and Weizhu Chen},
      year={2024},
      eprint={2409.12136},
      archivePrefix={arXiv},
      primaryClass={cs.CL},
      url={https://arxiv.org/abs/2409.12136}, 
}

@misc{zheng2025groupsequencepolicyoptimization,
      title={Group Sequence Policy Optimization}, 
      author={Chujie Zheng and Shixuan Liu and Mingze Li and Xiong-Hui Chen and Bowen Yu and Chang Gao and Kai Dang and Yuqiong Liu and Rui Men and An Yang and Jingren Zhou and Junyang Lin},
      year={2025},
      eprint={2507.18071},
      archivePrefix={arXiv},
      primaryClass={cs.LG},
      url={https://arxiv.org/abs/2507.18071}, 
}

@misc{rafailov2024directpreferenceoptimizationlanguage,
      title={Direct Preference Optimization: Your Language Model is Secretly a Reward Model}, 
      author={Rafael Rafailov and Archit Sharma and Eric Mitchell and Stefano Ermon and Christopher D. Manning and Chelsea Finn},
      year={2024},
      eprint={2305.18290},
      archivePrefix={arXiv},
      primaryClass={cs.LG},
      url={https://arxiv.org/abs/2305.18290}, 
}

@article{li2025veripo,
  title={VerIPO: Cultivating Long Reasoning in Video-LLMs via Verifier-Gudied Iterative Policy Optimization},
  author={Li, Yunxin and Chen, Xinyu and Li, Zitao and Liu, Zhenyu and Wang, Longyue and Luo, Wenhan and Hu, Baotian and Zhang, Min},
  journal={arXiv preprint arXiv:2505.19000},
  year={2025}
}

@article{DBLP:journals/corr/abs-2501-15368,
  author       = {Yadong Li and
                  Jun Liu and
                  Tao Zhang and
                  Song Chen and
                  Tianpeng Li and
                  Zehuan Li and
                  Lijun Liu and
                  Lingfeng Ming and
                  Guosheng Dong and
                  Da Pan and
                  Chong Li and
                  Yuanbo Fang and
                  Dongdong Kuang and
                  Mingrui Wang and
                  Chenglin Zhu and
                  Youwei Zhang and
                  Hongyu Guo and
                  Fengyu Zhang and
                  Yuran Wang and
                  Bowen Ding and
                  Wei Song and
                  Xu Li and
                  et al},
  title        = {Baichuan-Omni-1.5 Technical Report},
  journal      = {CoRR},
  volume       = {abs/2501.15368},
  year         = {2025}
}

@article{DBLP:journals/corr/abs-2504-21277,
  author       = {Guanghao Zhou and
                  Panjia Qiu and
                  Cen Chen and
                  Jie Wang and
                  Zheming Yang and
                  Jian Xu and
                  Minghui Qiu},
  title        = {Reinforced {MLLM:} {A} Survey on RL-Based Reasoning in Multimodal
                  Large Language Models},
  journal      = {CoRR},
  volume       = {abs/2504.21277},
  year         = {2025}
}

@article{DBLP:journals/corr/abs-2505-04921,
  author       = {Yunxin Li and
                  Zhenyu Liu and
                  Zitao Li and
                  Xuanyu Zhang and
                  Zhenran Xu and
                  Xinyu Chen and
                  Haoyuan Shi and
                  Shenyuan Jiang and
                  Xintong Wang and
                  Jifang Wang and
                  Shouzheng Huang and
                  Xinping Zhao and
                  Borui Jiang and
                  Lanqing Hong and
                  Longyue Wang and
                  Zhuotao Tian and
                  Baoxing Huai and
                  Wenhan Luo and
                  Weihua Luo and
                  Zheng Zhang and
                  Baotian Hu and
                  Min Zhang},
  title        = {Perception, Reason, Think, and Plan: {A} Survey on Large Multimodal
                  Reasoning Models},
  journal      = {CoRR},
  volume       = {abs/2505.04921},
  year         = {2025}
}

@article{DBLP:journals/corr/abs-2503-23278,
  author       = {Xinyi Hou and
                  Yanjie Zhao and
                  Shenao Wang and
                  Haoyu Wang},
  title        = {Model Context Protocol {(MCP):} Landscape, Security Threats, and Future
                  Research Directions},
  journal      = {CoRR},
  volume       = {abs/2503.23278},
  year         = {2025}
}

@misc{zhang2024lmmsevalrealitycheckevaluation,
      title={LMMs-Eval: Reality Check on the Evaluation of Large Multimodal Models}, 
      author={Kaichen Zhang and Bo Li and Peiyuan Zhang and Fanyi Pu and Joshua Adrian Cahyono and Kairui Hu and Shuai Liu and Yuanhan Zhang and Jingkang Yang and Chunyuan Li and Ziwei Liu},
      year={2024},
      eprint={2407.12772},
      archivePrefix={arXiv},
      primaryClass={cs.CL},
      url={https://arxiv.org/abs/2407.12772}, 
}

@inproceedings{liu2024mmbench,
  title={Mmbench: Is your multi-modal model an all-around player?},
  author={Liu, Yuan and Duan, Haodong and Zhang, Yuanhan and Li, Bo and Zhang, Songyang and Zhao, Wangbo and Yuan, Yike and Wang, Jiaqi and He, Conghui and Liu, Ziwei and others},
  booktitle={European conference on computer vision},
  pages={216--233},
  year={2024},
  organization={Springer}
}

@article{mmstar,
title={Are We on the Right Way for Evaluating Large Vision-Language Models?},
author={Chen, Lin and Li, Jinsong and Dong, Xiaoyi and Zhang, Pan and Zang, Yuhang and Chen, Zehui and Duan, Haodong and Wang, Jiaqi and Qiao, Yu and Lin, Dahua and others},
journal={arXiv preprint arXiv:2403.20330},
year={2024}
}

@misc{real_world_qa,
  title={Real World QA Benchmark
}, 
  author={X},
  year={2025},
  howpublished = {\url{https://huggingface.co/datasets/xai-org/RealworldQA}}
}

@inproceedings{gqa,
  title={Gqa: A new dataset for real-world visual reasoning and compositional question answering},
  author={Hudson, Drew A and Manning, Christopher D},
  booktitle={Proceedings of the IEEE/CVF conference on computer vision and pattern recognition},
  pages={6700--6709},
  year={2019}
}

@article{mme-realworld,
    title={MME-RealWorld: Could Your Multimodal LLM Challenge High-Resolution Real-World Scenarios that are Difficult for Humans?},
    author={Zhang, Yi-Fan and Zhang, Huanyu and Tian, Haochen and Fu, Chaoyou and Zhang, Shuangqing and Wu, Junfei and Li, Feng and Wang, Kun and Wen, Qingsong and Zhang, Zhang and others},
    journal={arXiv preprint arXiv:2408.13257},
    year={2024}
  }

@misc{cvbench,
      title={Cambrian-1: A Fully Open, Vision-Centric Exploration of Multimodal LLMs},
      author={Shengbang Tong and Ellis Brown and Penghao Wu and Sanghyun Woo and Manoj Middepogu and Sai Charitha Akula and Jihan Yang and Shusheng Yang and Adithya Iyer and Xichen Pan and Austin Wang and Rob Fergus and Yann LeCun and Saining Xie},
      year={2024},
      eprint={2406.16860},
}

@inproceedings{ai2d,
  title={A diagram is worth a dozen images},
  author={Kembhavi, Aniruddha and Salvato, Mike and Kolve, Eric and Seo, Minjoon and Hajishirzi, Hannaneh and Farhadi, Ali},
  booktitle={Computer Vision--ECCV 2016: 14th European Conference, Amsterdam, The Netherlands, October 11--14, 2016, Proceedings, Part IV 14},
  pages={235--251},
  year={2016},
  organization={Springer}
}

@inproceedings{mmmu,
  title={Mmmu: A massive multi-discipline multimodal understanding and reasoning benchmark for expert agi},
  author={Yue, Xiang and Ni, Yuansheng and Zhang, Kai and Zheng, Tianyu and Liu, Ruoqi and Zhang, Ge and Stevens, Samuel and Jiang, Dongfu and Ren, Weiming and Sun, Yuxuan and others},
  booktitle={Proceedings of the IEEE/CVF Conference on Computer Vision and Pattern Recognition},
  pages={9556--9567},
  year={2024}
}

@article{yue2024mmmupro,
  title={MMMU-Pro: A More Robust Multi-discipline Multimodal Understanding Benchmark},
  author={Xiang Yue and Tianyu Zheng and Yuansheng Ni and Yubo Wang and Kai Zhang and Shengbang Tong and Yuxuan Sun and Botao Yu and Ge Zhang and Huan Sun and Yu Su and Wenhu Chen and Graham Neubig},
  journal={arXiv preprint arXiv:2409.02813},
  year={2024}
}

@article{mathvista,
  title={Mathvista: Evaluating mathematical reasoning of foundation models in visual contexts},
  author={Lu, Pan and Bansal, Hritik and Xia, Tony and Liu, Jiacheng and Li, Chunyuan and Hajishirzi, Hannaneh and Cheng, Hao and Chang, Kai-Wei and Galley, Michel and Gao, Jianfeng},
  journal={arXiv preprint arXiv:2310.02255},
  year={2023}
}

@article{mathvision,
  title={Measuring multimodal mathematical reasoning with math-vision dataset},
  author={Wang, Ke and Pan, Junting and Shi, Weikang and Lu, Zimu and Ren, Houxing and Zhou, Aojun and Zhan, Mingjie and Li, Hongsheng},
  journal={Advances in Neural Information Processing Systems},
  volume={37},
  pages={95095--95169},
  year={2024}
}

@misc{logicvista,
      title={LogicVista: Multimodal LLM Logical Reasoning Benchmark in Visual Contexts}, 
      author={Yijia Xiao and Edward Sun and Tianyu Liu and Wei Wang},
      year={2024},
      eprint={2407.04973},
      archivePrefix={arXiv},
      primaryClass={cs.AI},
      url={https://arxiv.org/abs/2407.04973}, 
}

@inproceedings{docvqa,
  title={Docvqa: A dataset for vqa on document images},
  author={Mathew, Minesh and Karatzas, Dimosthenis and Jawahar, CV},
  booktitle={Proceedings of the IEEE/CVF winter conference on applications of computer vision},
  pages={2200--2209},
  year={2021}
}

@article{chartqa,
  title={Chartqa: A benchmark for question answering about charts with visual and logical reasoning},
  author={Masry, Ahmed and Long, Do Xuan and Tan, Jia Qing and Joty, Shafiq and Hoque, Enamul},
  journal={arXiv preprint arXiv:2203.10244},
  year={2022}
}

@article{seed2plus,
  title={SEED-Bench-2-Plus: Benchmarking Multimodal Large Language Models with Text-Rich Visual Comprehension},
  author={Li, Bohao and Ge, Yuying and Chen, Yi and Ge, Yixiao and Zhang, Ruimao and Shan, Ying},
  journal={arXiv preprint arXiv:2404.16790},
  year={2024}
}

@article{charxivdq,
  title={Charxiv: Charting gaps in realistic chart understanding in multimodal llms},
  author={Wang, Zirui and Xia, Mengzhou and He, Luxi and Chen, Howard and Liu, Yitao and Zhu, Richard and Liang, Kaiqu and Wu, Xindi and Liu, Haotian and Malladi, Sadhika and others},
  journal={Advances in Neural Information Processing Systems},
  volume={37},
  pages={113569--113697},
  year={2024}
}

@inproceedings{video_mme,
  title={Video-mme: The first-ever comprehensive evaluation benchmark of multi-modal llms in video analysis},
  author={Fu, Chaoyou and Dai, Yuhan and Luo, Yongdong and Li, Lei and Ren, Shuhuai and Zhang, Renrui and Wang, Zihan and Zhou, Chenyu and Shen, Yunhang and Zhang, Mengdan and others},
  booktitle={Proceedings of the Computer Vision and Pattern Recognition Conference},
  pages={24108--24118},
  year={2025}
}

@article{video_mmmu,
  title={Video-MMMU: Evaluating Knowledge Acquisition from Multi-Discipline Professional Videos},
  author={Hu, Kairui and Wu, Penghao and Pu, Fanyi and Xiao, Wang and Zhang, Yuanhan and Yue, Xiang and Li, Bo and Liu, Ziwei},
  journal={arXiv preprint arXiv:2501.13826},
  year={2025}
}

@article{longvideobench,
  title={Longvideobench: A benchmark for long-context interleaved video-language understanding},
  author={Wu, Haoning and Li, Dongxu and Chen, Bei and Li, Junnan},
  journal={Advances in Neural Information Processing Systems},
  volume={37},
  pages={28828--28857},
  year={2024}
}

@inproceedings{mvbench,
  title={Mvbench: A comprehensive multi-modal video understanding benchmark},
  author={Li, Kunchang and Wang, Yali and He, Yinan and Li, Yizhuo and Wang, Yi and Liu, Yi and Wang, Zun and Xu, Jilan and Chen, Guo and Luo, Ping and others},
  booktitle={CVPR},
  year={2024}
}

@inproceedings{egoschema,
  title={Egoschema: A diagnostic benchmark for very long-form video language understanding},
  author={Mangalam, Karttikeya and Akshulakov, Raiymbek and Malik, Jitendra},
  booktitle={NeurIPS},
  year={2023}
}

@article{vsi-bench,
    title={{Thinking in Space: How Multimodal Large Language Models See, Remember and Recall Spaces}},
    author={Yang, Jihan and Yang, Shusheng and Gupta, Anjali and Han, Rilyn and Fei-Fei, Li and Xie, Saining},
    year={2024},
    journal={arXiv preprint arXiv:2412.14171},
}

@misc{tomato,
      title={TOMATO: Assessing Visual Temporal Reasoning Capabilities in Multimodal Foundation Models}, 
      author={Ziyao Shangguan and Chuhan Li and Yuxuan Ding and Yanan Zheng and Yilun Zhao and Tesca Fitzgerald and Arman Cohan},
      year={2024},
      eprint={2410.23266},
      archivePrefix={arXiv},
      primaryClass={cs.CV},
      url={https://arxiv.org/abs/2410.23266}, 
}

@inproceedings{charades-sta,
  title={Tall: Temporal activity localization via language query},
  author={Gao, Jiyang and Sun, Chen and Yang, Zhenheng and Nevatia, Ram},
  booktitle={Proceedings of the IEEE international conference on computer vision},
  pages={5267--5275},
  year={2017}
}

@inproceedings{gpqa,
  title={Gpqa: A graduate-level google-proof q\&a benchmark},
  author={Rein, David and Hou, Betty Li and Stickland, Asa Cooper and Petty, Jackson and Pang, Richard Yuanzhe and Dirani, Julien and Michael, Julian and Bowman, Samuel R},
  booktitle={First Conference on Language Modeling},
  year={2024}
}

@article{mmlupro,
  title={Mmlu-pro: A more robust and challenging multi-task language understanding benchmark},
  author={Wang, Yubo and Ma, Xueguang and Zhang, Ge and Ni, Yuansheng and Chandra, Abhranil and Guo, Shiguang and Ren, Weiming and Arulraj, Aaran and He, Xuan and Jiang, Ziyan and others},
  journal={arXiv preprint arXiv:2406.01574},
  year={2024}
}

@article{hong2025worldsense,
  title={Worldsense: Evaluating real-world omnimodal understanding for multimodal llms},
  author={Hong, Jack and Yan, Shilin and Cai, Jiayin and Jiang, Xiaolong and Hu, Yao and Xie, Weidi},
  journal={arXiv preprint arXiv:2502.04326},
  year={2025}
}

@article{streamingbench,
  title={Streamingbench: Assessing the gap for mllms to achieve streaming video understanding},
  author={Lin, Junming and Fang, Zheng and Chen, Chi and Wan, Zihao and Luo, Fuwen and Li, Peng and Liu, Yang and Sun, Maosong},
  journal={arXiv preprint arXiv:2411.03628},
  year={2024}
}

@article{omnivideobench,
  title={OmniVideoBench: Towards Audio-Visual Understanding Evaluation for Omni MLLMs},
  author={Li, Caorui and Chen, Yu and Ji, Yiyan and Xu, Jin and Cui, Zhenyu and Li, Shihao and Zhang, Yuanxing and Tang, Jiafu and Song, Zhenghao and Zhang, Dingling and others},
  journal={arXiv preprint arXiv:2510.10689},
  year={2025}
}

@article{omnibench,
  title={Omnibench: Towards the future of universal omni-language models},
  author={Li, Yizhi and Zhang, Ge and Ma, Yinghao and Yuan, Ruibin and Zhu, Kang and Guo, Hangyu and Liang, Yiming and Liu, Jiaheng and Wang, Zekun and Yang, Jian and others},
  journal={arXiv preprint arXiv:2409.15272},
  year={2024}
}

@article{pixelprose,
  title={From pixels to prose: A large dataset of dense image captions},
  author={Singla, Vasu and Yue, Kaiyu and Paul, Sukriti and Shirkavand, Reza and Jayawardhana, Mayuka and Ganjdanesh, Alireza and Huang, Heng and Bhatele, Abhinav and Somepalli, Gowthami and Goldstein, Tom},
  journal={arXiv preprint arXiv:2406.10328},
  year={2024}
}

@article{grit,
  title={Kosmos-2: Grounding multimodal large language models to the world},
  author={Peng, Zhiliang and Wang, Wenhui and Dong, Li and Hao, Yaru and Huang, Shaohan and Ma, Shuming and Wei, Furu},
  journal={arXiv preprint arXiv:2306.14824},
  year={2023}
}

@inproceedings{cc3m,
  title={Conceptual 12m: Pushing web-scale image-text pre-training to recognize long-tail visual concepts},
  author={Changpinyo, Soravit and Sharma, Piyush and Ding, Nan and Soricut, Radu},
  booktitle={Proceedings of the IEEE/CVF conference on computer vision and pattern recognition},
  pages={3558--3568},
  year={2021}
}

@article{cambrian-10m,
  title={Cambrian-1: A fully open, vision-centric exploration of multimodal llms},
  author={Tong, Peter and Brown, Ellis and Wu, Penghao and Woo, Sanghyun and IYER, Adithya Jairam Vedagiri and Akula, Sai Charitha and Yang, Shusheng and Yang, Jihan and Middepogu, Manoj and Wang, Ziteng and others},
  journal={Advances in Neural Information Processing Systems},
  volume={37},
  pages={87310--87356},
  year={2024}
}

@article{Llava-onevision,
  title={Llava-onevision: Easy visual task transfer},
  author={Li, Bo and Zhang, Yuanhan and Guo, Dong and Zhang, Renrui and Li, Feng and Zhang, Hao and Zhang, Kaichen and Zhang, Peiyuan and Li, Yanwei and Liu, Ziwei and others},
  journal={arXiv preprint arXiv:2408.03326},
  year={2024}
}

@misc{docmatix,
      title={Building and better understanding vision-language models: insights and future directions.}, 
      author={Hugo Laurençon and Andrés Marafioti and Victor Sanh and Léo Tronchon},
      year={2024},
      eprint={2408.12637},
      archivePrefix={arXiv},
      primaryClass={cs.CV}
}

@article{mmk12,
  title={Mm-eureka: Exploring the frontiers of multimodal reasoning with rule-based reinforcement learning},
  author={Meng, Fanqing and Du, Lingxiao and Liu, Zongkai and Zhou, Zhixiang and Lu, Quanfeng and Fu, Daocheng and Han, Tiancheng and Shi, Botian and Wang, Wenhai and He, Junjun and others},
  journal={arXiv preprint arXiv:2503.07365},
  year={2025}
}

@inproceedings{v*,
  title={V?: Guided visual search as a core mechanism in multimodal llms},
  author={Wu, Penghao and Xie, Saining},
  booktitle={Proceedings of the IEEE/CVF Conference on Computer Vision and Pattern Recognition},
  pages={13084--13094},
  year={2024}
}

@article{vision-r1,
  title={Vision-r1: Incentivizing reasoning capability in multimodal large language models},
  author={Huang, Wenxuan and Jia, Bohan and Zhai, Zijie and Cao, Shaosheng and Ye, Zheyu and Zhao, Fei and Xu, Zhe and Hu, Yao and Lin, Shaohui},
  journal={arXiv preprint arXiv:2503.06749},
  year={2025}
}

@article{mmcoldstart,
  title={Advancing Multimodal Reasoning via Reinforcement Learning with Cold Start},
  author={Wei, Lai and Li, Yuting and Zheng, Kaipeng and Wang, Chen and Wang, Yue and Kong, Linghe and Sun, Lichao and Huang, Weiran},
  journal={arXiv preprint arXiv:2505.22334},
  year={2025}
}

@misc{valley,
      title={Valley: Video Assistant with Large Language model Enhanced abilitY},
      author={Ruipu Luo and Ziwang Zhao and Min Yang and Junwei Dong and Minghui Qiu and Pengcheng Lu and Tao Wang and Zhongyu Wei},
      year={2023},
      eprint={2306.07207},
      archivePrefix={arXiv},
      primaryClass={cs.CV}
}

@article{chen2024sharegpt4video,
        title={ShareGPT4Video: Improving Video Understanding and Generation with Better Captions},
        author={Chen, Lin and Wei, Xilin and Li, Jinsong and Dong, Xiaoyi and Zhang, Pan and Zang, Yuhang and Chen, Zehui and Duan, Haodong and Lin, Bin and Tang, Zhenyu and Yuan, Li and Qiao, Yu and Lin, Dahua and Zhao, Feng and Wang, Jiaqi},
        journal={arXiv preprint arXiv:2406.04325},
        year={2024}
      }

@misc{llava-video-178k,
    title={Video Instruction Tuning With Synthetic Data}, 
    author={Yuanhan Zhang and Jinming Wu and Wei Li and Bo Li and Zejun Ma and Ziwei Liu and Chunyuan Li},
    year={2024},
    eprint={2410.02713},
    archivePrefix={arXiv},
    primaryClass={cs.CV},
    url={https://arxiv.org/abs/2410.02713}, 
}

@misc{li2024videovista,
        title={VideoVista: A Versatile Benchmark for Video Understanding and Reasoning}, 
        author={Yunxin Li and Xinyu Chen and Baotian Hu and Longyue Wang and Haoyuan Shi and Min Zhang},
        year={2024},
        eprint={2406.11303},
        archivePrefix={arXiv}
}

@article{videoGPTplus,
    title={VideoGPT+: Integrating Image and Video Encoders for Enhanced Video Understanding},
    author={Maaz, Muhammad and Rasheed, Hanoona and Khan, Salman and Khan, Fahad Shahbaz},
    journal={arxiv},
    year={2024},
    url={https://arxiv.org/abs/2406.09418}
}

@misc{FineVideo,
  title={FineVideo},
  author={Farré, Miquel and Marafioti, Andi and Tunstall, Lewis and Von Werra, Leandro and Wolf, Thomas},
  year={2024},
  howpublished={\url{https://huggingface.co/datasets/HuggingFaceFV/finevideo}},
}

@misc{timechatonline,
    title={TimeChat-Online: 80% Visual Tokens are Naturally Redundant in Streaming Videos}, 
    author={Linli Yao and Yicheng Li and Yuancheng Wei and Lei Li and Shuhuai Ren and Yuanxin Liu and Kun Ouyang and Lean Wang and Shicheng Li and Sida Li and Lingpeng Kong and Qi Liu and Yuanxing Zhang and Xu Sun},
    year={2025},
    eprint={2504.17343},
    archivePrefix={arXiv},
    primaryClass={cs.CV},
    url={https://arxiv.org/abs/2504.17343},
}

@article{cinepile,
  title={CinePile: A Long Video Question Answering Dataset and Benchmark},
  author={Rawal, Ruchit and Saifullah, Khalid and Basri, Ronen and Jacobs, David and Somepalli, Gowthami and Goldstein, Tom},
  journal={arXiv preprint arXiv:2405.08813},
  year={2024}
}

@article{sf20k,
  title={Long Story Short: Story-level Video Understanding from 20K Short Films}, 
  author={Ridouane Ghermi and Xi Wang and Vicky Kalogeiton and Ivan Laptev},
  journal={arXiv preprint arXiv:2406.10221},
  year={2025},
}

@article{neptune24,
      title={Neptune: The Long Orbit to Benchmarking Long Video Understanding},
      author={Nagrani, Arsha and Zhang, Mingda and Mehran, Ramin and Hornung, Rachel and Gundavarapu, Nitesh Bharadwaj and Jha, Nilpa and Myers, Austin and Zhou, Xingyi and Gong, Boqing and Schmid, Cordelia and Sirotenko, Mikhail and Zhu, Yukun and Weyand, Tobias},
      journal={arXiv preprint arXiv:2412.09582},
      year={2024},
}

@inproceedings{jia2022egotaskqa,
    title = {EgoTaskQA: Understanding Human Tasks in Egocentric Videos},
    author = {Jia, Baoxiong and Lei, Ting and Zhu, Song-Chun and Huang, Siyuan},
    booktitle = {The 36th Conference on Neural Information Processing Systems (NeurIPS 2022) Track on Datasets and Benchmarks},
    year = {2022}
}

@inproceedings{xie2024funqa,
    author={Binzhu Xie and Sicheng Zhang and Zitang Zhou and Bo Li and Yuanhan Zhang and Jack Hessel and Jingkang Yang and Ziwei Liu},
    year={2024},
    title={FunQA: Towards Surprising Video Comprehension},
    booktitle = {European Conference on Computer Vision (ECCV)},
    url={https://www.ecva.net/papers/eccv_2024/papers_ECCV/papers/00010.pdf}, 
}

@misc{yang2024vript,
      title={Vript: A Video Is Worth Thousands of Words}, 
      author={Dongjie Yang and Suyuan Huang and Chengqiang Lu and Xiaodong Han and Haoxin Zhang and Yan Gao and Yao Hu and Hai Zhao},
      year={2024},
      eprint={2406.06040},
      archivePrefix={arXiv},
      primaryClass={cs.CV}
}

@misc{tarsier2,
      title={Tarsier2: Advancing Large Vision-Language Models from Detailed Video Description to Comprehensive Video Understanding}, 
      author={Liping Yuan and Jiawei Wang and Haomiao Sun and Yuchen Zhang and Yuan Lin},
      year={2025},
      eprint={2501.07888},
      archivePrefix={arXiv},
      primaryClass={cs.CV},
      url={https://arxiv.org/abs/2501.07888}, 
}

@article{wang2024internvideo2,
  title={Internvideo2: Scaling video foundation models for multimodal video understanding},
  author={Wang, Yi and Li, Kunchang and Li, Xinhao and Yu, Jiashuo and He, Yinan and Chen, Guo and Pei, Baoqi and Zheng, Rongkun and Xu, Jilan and Wang, Zun and others},
  journal={arXiv preprint arXiv:2403.15377},
  year={2024}
}

@article{ouyang2025sr_91k,
  title={SpaceR: Reinforcing MLLMs in Video Spatial Reasoning},
  author={Ouyang, Kun and Liu, Yuanxin and Wu, Haoning and Liu, Yi and Zhou, Hao and Zhou, Jie and Meng, Fandong and Sun, Xu},
  journal={arXiv preprint arXiv:2504.01805},
  year={2025}
}

@misc{OpenOrca,
  title = {OpenOrca: An Open Dataset of GPT Augmented FLAN Reasoning Traces},
  author = {Wing Lian and Bleys Goodson and Eugene Pentland and Austin Cook and Chanvichet Vong and "Teknium"},
  year = {2023},
  publisher = {HuggingFace},
  journal = {HuggingFace repository},
  howpublished = {\url{https://https://huggingface.co/datasets/Open-Orca/OpenOrca}},
}

@article{yu2025dapo,
  title={Dapo: An open-source llm reinforcement learning system at scale},
  author={Yu, Qiying and Zhang, Zheng and Zhu, Ruofei and Yuan, Yufeng and Zuo, Xiaochen and Yue, Yu and Dai, Weinan and Fan, Tiantian and Liu, Gaohong and Liu, Lingjun and others},
  journal={arXiv preprint arXiv:2503.14476},
  year={2025}
}

@misc{Nemotron,
      title={Llama-Nemotron: Efficient Reasoning Models}, 
      author={Akhiad Bercovich and Itay Levy and Izik Golan and Mohammad Dabbah and Ran El-Yaniv and Omri Puny and Ido Galil and Zach Moshe and Tomer Ronen and Najeeb Nabwani and Ido Shahaf and Oren Tropp and et al},
      year={2025},
      eprint={2505.00949},
      archivePrefix={arXiv},
      primaryClass={cs.CL},
      url={https://arxiv.org/abs/2505.00949}, 
}

@misc{Mixture-of-Thoughts,
    title = {Open R1: A fully open reproduction of DeepSeek-R1},
    url = {https://github.com/huggingface/open-r1},
    author = {Hugging Face},
    month = {January},
    year = {2025}
}

@inproceedings{chen2025videovista_cultural,
  title={VideoVista-CulturalLingo: 360° Horizons-Bridging Cultures, Languages, and Domains in Video Comprehension},
  author={Chen, Xinyu and Li, Yunxin and Shi, Haoyuan and Hu, Baotian and Luo, Wenhan and Wang, Yaowei and Zhang, Min},
  booktitle={Proceedings of the 63rd Annual Meeting of the Association for Computational Linguistics (Volume 1: Long Papers)},
  pages={27102--27128},
  year={2025}
}

@article{wang2025internvl3,
  title={Internvl3. 5: Advancing open-source multimodal models in versatility, reasoning, and efficiency},
  author={Wang, Weiyun and Gao, Zhangwei and Gu, Lixin and Pu, Hengjun and Cui, Long and Wei, Xingguang and Liu, Zhaoyang and Jing, Linglin and Ye, Shenglong and Shao, Jie and others},
  journal={arXiv preprint arXiv:2508.18265},
  year={2025}
}

@inproceedings{whisper,
  author       = {Alec Radford and
                  Jong Wook Kim and
                  Tao Xu and
                  Greg Brockman and
                  Christine McLeavey and
                  Ilya Sutskever},
  editor       = {Andreas Krause and
                  Emma Brunskill and
                  Kyunghyun Cho and
                  Barbara Engelhardt and
                  Sivan Sabato and
                  Jonathan Scarlett},
  title        = {Robust Speech Recognition via Large-Scale Weak Supervision},
  booktitle    = {International Conference on Machine Learning, {ICML} 2023, 23-29 July
                  2023, Honolulu, Hawaii, {USA}},
  series       = {Proceedings of Machine Learning Research},
  volume       = {202},
  pages        = {28492--28518},
  publisher    = {{PMLR}},
  year         = {2023},
  url          = {https://proceedings.mlr.press/v202/radford23a.html},
  bibsource    = {dblp computer science bibliography, https://dblp.org}
}

@article{lin2024pixwizard,
  title={Pixwizard: Versatile image-to-image visual assistant with open-language instructions},
  author={Lin, Weifeng and Wei, Xinyu and Zhang, Renrui and Zhuo, Le and Zhao, Shitian and Huang, Siyuan and Teng, Huan and Xie, Junlin and Qiao, Yu and Gao, Peng and others},
  journal={arXiv preprint arXiv:2409.15278},
  year={2024}
}

@article{mls,
  title={Mls: A large-scale multilingual dataset for speech research},
  author={Pratap, Vineel and Xu, Qiantong and Sriram, Anuroop and Synnaeve, Gabriel and Collobert, Ronan},
  journal={arXiv preprint arXiv:2012.03411},
  year={2020}
}

@inproceedings{giga,
  title={GigaSpeech: An Evolving, Multi-domain ASR Corpus with 10,000 Hours of Transcribed Audio},
  booktitle={Proc. Interspeech 2021},
  year={2021},
  author={Guoguo Chen and Shuzhou Chai and Guanbo Wang and Jiayu Du and Wei-Qiang Zhang and Chao Weng and Dan Su and Daniel Povey and Jan Trmal and Junbo Zhang and Mingjie Jin and Sanjeev Khudanpur and Shinji Watanabe and Shuaijiang Zhao and Wei Zou and Xiangang Li and Xuchen Yao and Yongqing Wang and Yujun Wang and Zhao You and Zhiyong Yan}
}

@inproceedings{commonvoice,
  author = {Ardila, R. and Branson, M. and Davis, K. and Henretty, M. and Kohler, M. and Meyer, J. and Morais, R. and Saunders, L. and Tyers, F. M. and Weber, G.},
  title = {Common Voice: A Massively-Multilingual Speech Corpus},
  booktitle = {Proceedings of the 12th Conference on Language Resources and Evaluation (LREC 2020)},
  pages = {4211--4215},
  year = 2020
}

@article{wavcaps,
  author       = {Xinhao Mei and
                  Chutong Meng and
                  Haohe Liu and
                  Qiuqiang Kong and
                  Tom Ko and
                  Chengqi Zhao and
                  Mark D. Plumbley and
                  Yuexian Zou and
                  Wenwu Wang},
  title        = {WavCaps: {A} ChatGPT-Assisted Weakly-Labelled Audio Captioning Dataset
                  for Audio-Language Multimodal Research},
  journal      = {{IEEE} {ACM} Trans. Audio Speech Lang. Process.},
  volume       = {32},
  pages        = {3339--3354},
  year         = {2024},
  url          = {https://doi.org/10.1109/TASLP.2024.3419446},
  doi          = {10.1109/TASLP.2024.3419446},
  bibsource    = {dblp computer science bibliography, https://dblp.org}
}

@inproceedings{clotho,
  author       = {Konstantinos Drossos and
                  Samuel Lipping and
                  Tuomas Virtanen},
  title        = {Clotho: an Audio Captioning Dataset},
  booktitle    = {2020 {IEEE} International Conference on Acoustics, Speech and Signal
                  Processing, {ICASSP} 2020, Barcelona, Spain, May 4-8, 2020},
  pages        = {736--740},
  publisher    = {{IEEE}},
  year         = {2020},
  url          = {https://doi.org/10.1109/ICASSP40776.2020.9052990},
  doi          = {10.1109/ICASSP40776.2020.9052990},
  bibsource    = {dblp computer science bibliography, https://dblp.org}
}

@inproceedings{meld,
  author       = {Soujanya Poria and
                  Devamanyu Hazarika and
                  Navonil Majumder and
                  Gautam Naik and
                  Erik Cambria and
                  Rada Mihalcea},
  editor       = {Anna Korhonen and
                  David R. Traum and
                  Llu{\'{\i}}s M{\`{a}}rquez},
  title        = {{MELD:} {A} Multimodal Multi-Party Dataset for Emotion Recognition
                  in Conversations},
  booktitle    = {Proceedings of the 57th Conference of the Association for Computational
                  Linguistics, {ACL} 2019, Florence, Italy, July 28- August 2, 2019,
                  Volume 1: Long Papers},
  pages        = {527--536},
  publisher    = {Association for Computational Linguistics},
  year         = {2019},
  url          = {https://doi.org/10.18653/v1/p19-1050},
  doi          = {10.18653/V1/P19-1050},
  bibsource    = {dblp computer science bibliography, https://dblp.org}
}

@inproceedings{musicbench,
  author       = {Jan Melechovsk{\'{y}} and
                  Zixun Guo and
                  Deepanway Ghosal and
                  Navonil Majumder and
                  Dorien Herremans and
                  Soujanya Poria},
  editor       = {Kevin Duh and
                  Helena G{\'{o}}mez{-}Adorno and
                  Steven Bethard},
  title        = {Mustango: Toward Controllable Text-to-Music Generation},
  booktitle    = {Proceedings of the 2024 Conference of the North American Chapter of
                  the Association for Computational Linguistics: Human Language Technologies
                  (Volume 1: Long Papers), {NAACL} 2024, Mexico City, Mexico, June 16-21,
                  2024},
  pages        = {8293--8316},
  publisher    = {Association for Computational Linguistics},
  year         = {2024},
  url          = {https://doi.org/10.18653/v1/2024.naacl-long.459},
  doi          = {10.18653/V1/2024.NAACL-LONG.459},
  bibsource    = {dblp computer science bibliography, https://dblp.org}
}

@inproceedings{lpmusic,
  author       = {Seungheon Doh and
                  Keunwoo Choi and
                  Jongpil Lee and
                  Juhan Nam},
  editor       = {Augusto Sarti and
                  Fabio Antonacci and
                  Mark Sandler and
                  Paolo Bestagini and
                  Simon Dixon and
                  Beici Liang and
                  Ga{\"{e}}l Richard and
                  Johan Pauwels},
  title        = {LP-MusicCaps: LLM-Based Pseudo Music Captioning},
  booktitle    = {Proceedings of the 24th International Society for Music Information
                  Retrieval Conference, {ISMIR} 2023, Milan, Italy, November 5-9, 2023},
  pages        = {409--416},
  year         = {2023},
  url          = {https://doi.org/10.5281/zenodo.10265311},
  doi          = {10.5281/ZENODO.10265311},
  bibsource    = {dblp computer science bibliography, https://dblp.org}
}

@inproceedings{clothoaqa,
  author       = {Samuel Lipping and
                  Parthasaarathy Sudarsanam and
                  Konstantinos Drossos and
                  Tuomas Virtanen},
  title        = {Clotho-AQA: {A} Crowdsourced Dataset for Audio Question Answering},
  booktitle    = {30th European Signal Processing Conference, {EUSIPCO} 2022, Belgrade,
                  Serbia, August 29 - Sept. 2, 2022},
  pages        = {1140--1144},
  publisher    = {{IEEE}},
  year         = {2022},
  url          = {https://ieeexplore.ieee.org/document/9909680},
  bibsource    = {dblp computer science bibliography, https://dblp.org}
}

@inproceedings{audiocaps,
  author       = {Chris Dongjoo Kim and
                  Byeongchang Kim and
                  Hyunmin Lee and
                  Gunhee Kim},
  editor       = {Jill Burstein and
                  Christy Doran and
                  Thamar Solorio},
  title        = {AudioCaps: Generating Captions for Audios in The Wild},
  booktitle    = {Proceedings of the 2019 Conference of the North American Chapter of
                  the Association for Computational Linguistics: Human Language Technologies,
                  {NAACL-HLT} 2019, Minneapolis, MN, USA, June 2-7, 2019, Volume 1 (Long
                  and Short Papers)},
  pages        = {119--132},
  publisher    = {Association for Computational Linguistics},
  year         = {2019},
  url          = {https://doi.org/10.18653/v1/n19-1011},
  doi          = {10.18653/V1/N19-1011},
  bibsource    = {dblp computer science bibliography, https://dblp.org}
}

@article{asvp_esd,
  title={ASVP-ESD: A dataset and its benchmark for emotion recognition using both speech and non-speech utterances},
  author={Landry, Dejoli and He, Qianhua and Yan, Haikang and Li, Yanxiong},
  journal={Global Scientific Journals},
  volume={8},
  pages={1793--1798},
  year={2020}
}

@article{crema_d,
  author       = {Houwei Cao and
                  David G. Cooper and
                  Michael K. Keutmann and
                  Ruben C. Gur and
                  Ani Nenkova and
                  Ragini Verma},
  title        = {{CREMA-D:} Crowd-Sourced Emotional Multimodal Actors Dataset},
  journal      = {{IEEE} Trans. Affect. Comput.},
  volume       = {5},
  number       = {4},
  pages        = {377--390},
  year         = {2014},
  url          = {https://doi.org/10.1109/TAFFC.2014.2336244},
  doi          = {10.1109/TAFFC.2014.2336244},
  bibsource    = {dblp computer science bibliography, https://dblp.org}
}

@article{emov,
  author       = {Adaeze Adigwe and
                  No{\'{e}} Tits and
                  Kevin El Haddad and
                  Sarah Ostadabbas and
                  Thierry Dutoit},
  title        = {The Emotional Voices Database: Towards Controlling the Emotion Dimension
                  in Voice Generation Systems},
  journal      = {CoRR},
  volume       = {abs/1806.09514},
  year         = {2018},
  url          = {http://arxiv.org/abs/1806.09514},
  eprinttype    = {arXiv},
  eprint       = {1806.09514},
  bibsource    = {dblp computer science bibliography, https://dblp.org}
}

@article{esd,
  author       = {Kun Zhou and
                  Berrak Sisman and
                  Rui Liu and
                  Haizhou Li},
  title        = {Emotional voice conversion: Theory, databases and {ESD}},
  journal      = {Speech Commun.},
  volume       = {137},
  pages        = {1--18},
  year         = {2022},
  url          = {https://doi.org/10.1016/j.specom.2021.11.006},
  doi          = {10.1016/J.SPECOM.2021.11.006},
  bibsource    = {dblp computer science bibliography, https://dblp.org}
}

@inproceedings{jlcorpus,
  author       = {Jesin James and
                  Li Tian and
                  Catherine Inez Watson},
  editor       = {B. Yegnanarayana},
  title        = {An Open Source Emotional Speech Corpus for Human Robot Interaction
                  Applications},
  booktitle    = {19th Annual Conference of the International Speech Communication Association,
                  Interspeech 2018, Hyderabad, India, September 2-6, 2018},
  pages        = {2768--2772},
  publisher    = {{ISCA}},
  year         = {2018},
  url          = {https://doi.org/10.21437/Interspeech.2018-1349},
  doi          = {10.21437/INTERSPEECH.2018-1349},
  bibsource    = {dblp computer science bibliography, https://dblp.org}
}

@article{ravdess,
  title={The Ryerson Audio-Visual Database of Emotional Speech and Song (RAVDESS): A dynamic, multimodal set of facial and vocal expressions in North American English},
  author={Livingstone, Steven R and Russo, Frank A},
  journal={PloS one},
  volume={13},
  number={5},
  pages={e0196391},
  year={2018},
  publisher={Public Library of Science San Francisco, CA USA}
}

@inproceedings{MusiLingo,
  author       = {Zihao Deng and
                  Yinghao Ma and
                  Yudong Liu and
                  Rongchen Guo and
                  Ge Zhang and
                  Wenhu Chen and
                  Wenhao Huang and
                  Emmanouil Benetos},
  editor       = {Kevin Duh and
                  Helena G{\'{o}}mez{-}Adorno and
                  Steven Bethard},
  title        = {MusiLingo: Bridging Music and Text with Pre-trained Language Models
                  for Music Captioning and Query Response},
  booktitle    = {Findings of the Association for Computational Linguistics: {NAACL}
                  2024, Mexico City, Mexico, June 16-21, 2024},
  pages        = {3643--3655},
  publisher    = {Association for Computational Linguistics},
  year         = {2024},
  url          = {https://doi.org/10.18653/v1/2024.findings-naacl.231},
  doi          = {10.18653/V1/2024.FINDINGS-NAACL.231},
  bibsource    = {dblp computer science bibliography, https://dblp.org}
}

@inproceedings{librispeech,
  author       = {Vassil Panayotov and
                  Guoguo Chen and
                  Daniel Povey and
                  Sanjeev Khudanpur},
  title        = {Librispeech: An {ASR} corpus based on public domain audio books},
  booktitle    = {2015 {IEEE} International Conference on Acoustics, Speech and Signal
                  Processing, {ICASSP} 2015, South Brisbane, Queensland, Australia,
                  April 19-24, 2015},
  pages        = {5206--5210},
  publisher    = {{IEEE}},
  year         = {2015},
  url          = {https://doi.org/10.1109/ICASSP.2015.7178964},
  doi          = {10.1109/ICASSP.2015.7178964},
  bibsource    = {dblp computer science bibliography, https://dblp.org}
}

@inproceedings{race,
  author       = {Guokun Lai and
                  Qizhe Xie and
                  Hanxiao Liu and
                  Yiming Yang and
                  Eduard H. Hovy},
  editor       = {Martha Palmer and
                  Rebecca Hwa and
                  Sebastian Riedel},
  title        = {{RACE:} Large-scale ReAding Comprehension Dataset From Examinations},
  booktitle    = {Proceedings of the 2017 Conference on Empirical Methods in Natural
                  Language Processing, {EMNLP} 2017, Copenhagen, Denmark, September
                  9-11, 2017},
  pages        = {785--794},
  publisher    = {Association for Computational Linguistics},
  year         = {2017},
  url          = {https://doi.org/10.18653/v1/d17-1082},
  doi          = {10.18653/V1/D17-1082},
  bibsource    = {dblp computer science bibliography, https://dblp.org}
}

@inproceedings{aishell1,
  author       = {Hui Bu and
                  Jiayu Du and
                  Xingyu Na and
                  Bengu Wu and
                  Hao Zheng},
  title        = {{AISHELL-1:} An open-source Mandarin speech corpus and a speech recognition
                  baseline},
  booktitle    = {20th Conference of the Oriental Chapter of the International Coordinating
                  Committee on Speech Databases and Speech {I/O} Systems and Assessment,
                  {O-COCOSDA} 2017, Seoul, South Korea, November 1-3, 2017},
  pages        = {1--5},
  publisher    = {{IEEE}},
  year         = {2017},
  url          = {https://doi.org/10.1109/ICSDA.2017.8384449},
  doi          = {10.1109/ICSDA.2017.8384449},
  bibsource    = {dblp computer science bibliography, https://dblp.org}
}

@article{aishell3,
  author       = {Yao Shi and
                  Hui Bu and
                  Xin Xu and
                  Shaoji Zhang and
                  Ming Li},
  title        = {{AISHELL-3:} {A} Multi-speaker Mandarin {TTS} Corpus and the Baselines},
  journal      = {CoRR},
  volume       = {abs/2010.11567},
  year         = {2020},
  url          = {https://arxiv.org/abs/2010.11567},
  eprinttype    = {arXiv},
  eprint       = {2010.11567},
  bibsource    = {dblp computer science bibliography, https://dblp.org}
}

@inproceedings{people,
  author       = {Daniel Galvez and
                  Greg Diamos and
                  Juan Torres and
                  Keith Achorn and
                  Juan Felipe Cer{\'{o}}n and
                  Anjali Gopi and
                  David Kanter and
                  Max Lam and
                  Mark Mazumder and
                  Vijay Janapa Reddi},
  editor       = {Joaquin Vanschoren and
                  Sai{-}Kit Yeung},
  title        = {The People's Speech: {A} Large-Scale Diverse English Speech Recognition
                  Dataset for Commercial Usage},
  booktitle    = {Proceedings of the Neural Information Processing Systems Track on
                  Datasets and Benchmarks 1, NeurIPS Datasets and Benchmarks 2021, December
                  2021, virtual},
  year         = {2021},
  url          = {https://datasets-benchmarks-proceedings.neurips.cc/paper/2021/hash/202cb962ac59075b964b07152d234b70-Abstract-round1.html},
  bibsource    = {dblp computer science bibliography, https://dblp.org}
}

@inproceedings{avqa,
  author       = {Pinci Yang and
                  Xin Wang and
                  Xuguang Duan and
                  Hong Chen and
                  Runze Hou and
                  Cong Jin and
                  Wenwu Zhu},
  editor       = {Jo{\~{a}}o Magalh{\~{a}}es and
                  Alberto Del Bimbo and
                  Shin'ichi Satoh and
                  Nicu Sebe and
                  Xavier Alameda{-}Pineda and
                  Qin Jin and
                  Vincent Oria and
                  Laura Toni},
  title        = {{AVQA:} {A} Dataset for Audio-Visual Question Answering on Videos},
  booktitle    = {{MM} '22: The 30th {ACM} International Conference on Multimedia, Lisboa,
                  Portugal, October 10 - 14, 2022},
  pages        = {3480--3491},
  publisher    = {{ACM}},
  year         = {2022},
  url          = {https://doi.org/10.1145/3503161.3548291},
  doi          = {10.1145/3503161.3548291},
  bibsource    = {dblp computer science bibliography, https://dblp.org}
}

@inproceedings{musicavqa,
  author       = {Guangyao Li and
                  Yake Wei and
                  Yapeng Tian and
                  Chenliang Xu and
                  Ji{-}Rong Wen and
                  Di Hu},
  title        = {Learning to Answer Questions in Dynamic Audio-Visual Scenarios},
  booktitle    = {{IEEE/CVF} Conference on Computer Vision and Pattern Recognition,
                  {CVPR} 2022, New Orleans, LA, USA, June 18-24, 2022},
  pages        = {19086--19096},
  publisher    = {{IEEE}},
  year         = {2022},
  url          = {https://doi.org/10.1109/CVPR52688.2022.01852},
  doi          = {10.1109/CVPR52688.2022.01852},
  bibsource    = {dblp computer science bibliography, https://dblp.org}
}

@inproceedings{ultra,
  author       = {Ning Ding and
                  Yulin Chen and
                  Bokai Xu and
                  Yujia Qin and
                  Shengding Hu and
                  Zhiyuan Liu and
                  Maosong Sun and
                  Bowen Zhou},
  editor       = {Houda Bouamor and
                  Juan Pino and
                  Kalika Bali},
  title        = {Enhancing Chat Language Models by Scaling High-quality Instructional
                  Conversations},
  booktitle    = {Proceedings of the 2023 Conference on Empirical Methods in Natural
                  Language Processing, {EMNLP} 2023, Singapore, December 6-10, 2023},
  pages        = {3029--3051},
  publisher    = {Association for Computational Linguistics},
  year         = {2023},
  url          = {https://doi.org/10.18653/v1/2023.emnlp-main.183},
  doi          = {10.18653/V1/2023.EMNLP-MAIN.183},
  bibsource    = {dblp computer science bibliography, https://dblp.org}
}

@misc{belle,
  title={Belle: Be everyone’s large language model engine},
  author={Ji, Yunjie and Deng, Yong and Gong, Yan and Peng, Yiping and Niu, Qiang and Ma, Baochang and Li, Xiangang},
  year={2023}
}

@misc{OpenHermes,
  title = {OpenHermes 2.5: An Open Dataset of Synthetic Data for Generalist LLM Assistants},
  author = {Teknium},
  year = {2023},
  publisher = {HuggingFace},
  url = {https://huggingface.co/datasets/teknium/OpenHermes-2.5}
}

@article{moss,
  author       = {Tianxiang Sun and
                  Xiaotian Zhang and
                  Zhengfu He and
                  Peng Li and
                  Qinyuan Cheng and
                  Xiangyang Liu and
                  Hang Yan and
                  Yunfan Shao and
                  Qiong Tang and
                  Shiduo Zhang and
                  Xingjian Zhao and
                  Ke Chen and
                  Yining Zheng and
                  Zhejian Zhou and
                  Ruixiao Li and
                  Jun Zhan and
                  Yunhua Zhou and
                  Linyang Li and
                  Xiaogui Yang and
                  Lingling Wu and
                  Zhangyue Yin and
                  Xuanjing Huang and
                  Yu{-}Gang Jiang and
                  Xipeng Qiu},
  title        = {{MOSS:} An Open Conversational Large Language Model},
  journal      = {Mach. Intell. Res.},
  volume       = {21},
  number       = {5},
  pages        = {888--905},
  year         = {2024},
  url          = {https://doi.org/10.1007/s11633-024-1502-8},
  doi          = {10.1007/S11633-024-1502-8},
  bibsource    = {dblp computer science bibliography, https://dblp.org}
}

@misc{alpaca,
  author = {Rohan Taori and Ishaan Gulrajani and Tianyi Zhang and Yann Dubois and Xuechen Li and Carlos Guestrin and Percy Liang and Tatsunori B. Hashimoto },
  title = {Stanford Alpaca: An Instruction-following LLaMA model},
  year = {2023},
  publisher = {GitHub},
  journal = {GitHub repository},
  howpublished = {\url{https://github.com/tatsu-lab/stanford_alpaca}},
}

@inproceedings{pixmo,
  author       = {Matt Deitke and
                  Christopher Clark and
                  Sangho Lee and
                  Rohun Tripathi and
                  Yue Yang and
                  Jae Sung Park and
                  Mohammadreza Salehi and
                  Niklas Muennighoff and
                  Kyle Lo and
                  Luca Soldaini and
                  Jiasen Lu and
                  Taira Anderson and
                  Erin Bransom and
                  Kiana Ehsani and
                  Huong Ngo and
                  Yen{-}Sung Chen and
                  et al},
  title        = {Molmo and PixMo: Open Weights and Open Data for State-of-the-Art Vision-Language
                  Models},
  booktitle    = {{IEEE/CVF} Conference on Computer Vision and Pattern Recognition,
                  {CVPR} 2025, Nashville, TN, USA, June 11-15, 2025},
  pages        = {91--104},
  publisher    = {Computer Vision Foundation / {IEEE}},
  year         = {2025},
  url          = {https://openaccess.thecvf.com/content/CVPR2025/html/Deitke\_Molmo\_and\_PixMo\_Open\_Weights\_and\_Open\_Data\_for\_State-of-the-Art\_CVPR\_2025\_paper.html},
  doi          = {10.1109/CVPR52734.2025.00018},
  bibsource    = {dblp computer science bibliography, https://dblp.org}
}

@article{ji2024wavtokenizer,
  title={Wavtokenizer: an efficient acoustic discrete codec tokenizer for audio language modeling},
  author={Ji, Shengpeng and Jiang, Ziyue and Wang, Wen and Chen, Yifu and Fang, Minghui and Zuo, Jialong and Yang, Qian and Cheng, Xize and Wang, Zehan and Li, Ruiqi and others},
  journal={arXiv preprint arXiv:2408.16532},
  year={2024}
}

@article{tinystories,
  author       = {Ronen Eldan and
                  Yuanzhi Li},
  title        = {TinyStories: How Small Can Language Models Be and Still Speak Coherent
                  English?},
  journal      = {CoRR},
  volume       = {abs/2305.07759},
  year         = {2023},
  url          = {https://doi.org/10.48550/arXiv.2305.07759},
  doi          = {10.48550/ARXIV.2305.07759},
  eprinttype    = {arXiv},
  eprint       = {2305.07759},
  bibsource    = {dblp computer science bibliography, https://dblp.org}
}

@inproceedings{speechcraft,
  author       = {Zeyu Jin and
                  Jia Jia and
                  Qixin Wang and
                  Kehan Li and
                  Shuoyi Zhou and
                  Songtao Zhou and
                  Xiaoyu Qin and
                  Zhiyong Wu},
  editor       = {Jianfei Cai and
                  Mohan S. Kankanhalli and
                  Balakrishnan Prabhakaran and
                  Susanne Boll and
                  Ramanathan Subramanian and
                  Liang Zheng and
                  Vivek K. Singh and
                  Pablo C{\'{e}}sar and
                  Lexing Xie and
                  Dong Xu},
  title        = {SpeechCraft: {A} Fine-Grained Expressive Speech Dataset with Natural
                  Language Description},
  booktitle    = {Proceedings of the 32nd {ACM} International Conference on Multimedia,
                  {MM} 2024, Melbourne, VIC, Australia, 28 October 2024 - 1 November
                  2024},
  pages        = {1255--1264},
  publisher    = {{ACM}},
  year         = {2024},
  url          = {https://doi.org/10.1145/3664647.3681674},
  doi          = {10.1145/3664647.3681674},
  bibsource    = {dblp computer science bibliography, https://dblp.org}
}

@article{vccm,
  author       = {Shengpeng Ji and
                  Jialong Zuo and
                  Minghui Fang and
                  Siqi Zheng and
                  Qian Chen and
                  Wen Wang and
                  Ziyue Jiang and
                  Hai Huang and
                  Xize Cheng and
                  Rongjie Huang and
                  Zhou Zhao},
  title        = {ControlSpeech: Towards Simultaneous Zero-shot Speaker Cloning and
                  Zero-shot Language Style Control With Decoupled Codec},
  journal      = {CoRR},
  volume       = {abs/2406.01205},
  year         = {2024},
  url          = {https://doi.org/10.48550/arXiv.2406.01205},
  doi          = {10.48550/ARXIV.2406.01205},
  eprinttype    = {arXiv},
  eprint       = {2406.01205},
  bibsource    = {dblp computer science bibliography, https://dblp.org}
}

@article{stream,
  author       = {Shaolei Zhang and
                  Shoutao Guo and
                  Qingkai Fang and
                  Yan Zhou and
                  Yang Feng},
  title        = {Stream-Omni: Simultaneous Multimodal Interactions with Large Language-Vision-Speech
                  Model},
  journal      = {CoRR},
  volume       = {abs/2506.13642},
  year         = {2025},
  url          = {https://doi.org/10.48550/arXiv.2506.13642},
  doi          = {10.48550/ARXIV.2506.13642},
  eprinttype    = {arXiv},
  eprint       = {2506.13642},
  bibsource    = {dblp computer science bibliography, https://dblp.org}
}

@article{miniomni,
  author       = {Zhifei Xie and
                  Changqiao Wu},
  title        = {Mini-Omni: Language Models Can Hear, Talk While Thinking in Streaming},
  journal      = {CoRR},
  volume       = {abs/2408.16725},
  year         = {2024},
  url          = {https://doi.org/10.48550/arXiv.2408.16725},
  doi          = {10.48550/ARXIV.2408.16725},
  eprinttype    = {arXiv},
  eprint       = {2408.16725},
  bibsource    = {dblp computer science bibliography, https://dblp.org}
}

@article{llavavid,
  author       = {Yuanhan Zhang and
                  Jinming Wu and
                  Wei Li and
                  Bo Li and
                  Zejun Ma and
                  Ziwei Liu and
                  Chunyuan Li},
  title        = {LLaVA-Video: Video Instruction Tuning With Synthetic Data},
  journal      = {Trans. Mach. Learn. Res.},
  volume       = {2025},
  year         = {2025},
  url          = {https://openreview.net/forum?id=EElFGvt39K},
  bibsource    = {dblp computer science bibliography, https://dblp.org}
}

@inproceedings{mmau,
  author       = {S. Sakshi and
                  Utkarsh Tyagi and
                  Sonal Kumar and
                  Ashish Seth and
                  Ramaneswaran Selvakumar and
                  Oriol Nieto and
                  Ramani Duraiswami and
                  Sreyan Ghosh and
                  Dinesh Manocha},
  title        = {{MMAU:} {A} Massive Multi-Task Audio Understanding and Reasoning Benchmark},
  booktitle    = {The Thirteenth International Conference on Learning Representations,
                  {ICLR} 2025, Singapore, April 24-28, 2025},
  publisher    = {OpenReview.net},
  year         = {2025},
  url          = {https://openreview.net/forum?id=TeVAZXr3yv},
  bibsource    = {dblp computer science bibliography, https://dblp.org}
}

@inproceedings{libritts,
  author       = {Heiga Zen and
                  Viet Dang and
                  Rob Clark and
                  Yu Zhang and
                  Ron J. Weiss and
                  Ye Jia and
                  Zhifeng Chen and
                  Yonghui Wu},
  editor       = {Gernot Kubin and
                  Zdravko Kacic},
  title        = {LibriTTS: {A} Corpus Derived from LibriSpeech for Text-to-Speech},
  booktitle    = {20th Annual Conference of the International Speech Communication Association,
                  Interspeech 2019, Graz, Austria, September 15-19, 2019},
  pages        = {1526--1530},
  publisher    = {{ISCA}},
  year         = {2019},
  url          = {https://doi.org/10.21437/Interspeech.2019-2441},
  doi          = {10.21437/INTERSPEECH.2019-2441},
  bibsource    = {dblp computer science bibliography, https://dblp.org}
}

@article{seedtts,
  author       = {Philip Anastassiou and
                  Jiawei Chen and
                  Jitong Chen and
                  Yuanzhe Chen and
                  Zhuo Chen and
                  Ziyi Chen and
                  Jian Cong and
                  Lelai Deng and
                  Chuang Ding and
                  Lu Gao and
                  Mingqing Gong and
                  Peisong Huang and
                  Qingqing Huang and
                  Zhiying Huang and
                  Yuanyuan Huo and
                  Dongya Jia and
                  et al
                  },
  title        = {Seed-TTS: {A} Family of High-Quality Versatile Speech Generation Models},
  journal      = {CoRR},
  volume       = {abs/2406.02430},
  year         = {2024},
  url          = {https://doi.org/10.48550/arXiv.2406.02430},
  doi          = {10.48550/ARXIV.2406.02430},
  eprinttype    = {arXiv},
  eprint       = {2406.02430},
  bibsource    = {dblp computer science bibliography, https://dblp.org}
}

@inproceedings{llamaq,
  author       = {Eliya Nachmani and
                  Alon Levkovitch and
                  Roy Hirsch and
                  Julian Salazar and
                  Chulayuth Asawaroengchai and
                  Soroosh Mariooryad and
                  Ehud Rivlin and
                  R. J. Skerry{-}Ryan and
                  Michelle Tadmor Ramanovich},
  title        = {Spoken Question Answering and Speech Continuation Using Spectrogram-Powered
                  {LLM}},
  booktitle    = {The Twelfth International Conference on Learning Representations,
                  {ICLR} 2024, Vienna, Austria, May 7-11, 2024},
  publisher    = {OpenReview.net},
  year         = {2024},
  url          = {https://openreview.net/forum?id=izrOLJov5y},
  bibsource    = {dblp computer science bibliography, https://dblp.org}
}

@inproceedings{webq,
  author       = {Jonathan Berant and
                  Andrew Chou and
                  Roy Frostig and
                  Percy Liang},
  title        = {Semantic Parsing on Freebase from Question-Answer Pairs},
  booktitle    = {Proceedings of the 2013 Conference on Empirical Methods in Natural
                  Language Processing, {EMNLP} 2013, 18-21 October 2013, Grand Hyatt
                  Seattle, Seattle, Washington, USA, {A} meeting of SIGDAT, a Special
                  Interest Group of the {ACL}},
  pages        = {1533--1544},
  publisher    = {{ACL}},
  year         = {2013},
  url          = {https://doi.org/10.18653/v1/d13-1160},
  doi          = {10.18653/V1/D13-1160},
  bibsource    = {dblp computer science bibliography, https://dblp.org}
}

@article{bigbench,
  author       = {Aarohi Srivastava and
                  Abhinav Rastogi and
                  Abhishek Rao and
                  Abu Awal Md Shoeb and
                  Abubakar Abid and
                  Adam Fisch and
                  Adam R. Brown and
                  Adam Santoro and
                  Aditya Gupta and
                  Adri{\`{a}} Garriga{-}Alonso and
                  Agnieszka Kluska and
                  Aitor Lewkowycz and
                  Akshat Agarwal and
                  Alethea Power and
                  Alex Ray and
                  Alex Warstadt and
                  Alexander W. Kocurek and et al},
  title        = {Beyond the Imitation Game: Quantifying and extrapolating the capabilities
                  of language models},
  journal      = {CoRR},
  volume       = {abs/2206.04615},
  year         = {2022},
  url          = {https://doi.org/10.48550/arXiv.2206.04615},
  doi          = {10.48550/ARXIV.2206.04615},
  eprinttype    = {arXiv},
  eprint       = {2206.04615},
  bibsource    = {dblp computer science bibliography, https://dblp.org}
}

@inproceedings{multichallenge,
  author       = {Kaustubh Deshpande and
                  Ved Sirdeshmukh and
                  Johannes Baptist Mols and
                  Lifeng Jin and
                  Ed{-}Yeremai Hernandez{-}Cardona and
                  Dean Lee and
                  Jeremy Kritz and
                  Willow E. Primack and
                  Summer Yue and
                  Chen Xing},
  editor       = {Wanxiang Che and
                  Joyce Nabende and
                  Ekaterina Shutova and
                  Mohammad Taher Pilehvar},
  title        = {MultiChallenge: {A} Realistic Multi-Turn Conversation Evaluation Benchmark
                  Challenging to Frontier LLMs},
  booktitle    = {Findings of the Association for Computational Linguistics, {ACL} 2025,
                  Vienna, Austria, July 27 - August 1, 2025},
  pages        = {18632--18702},
  publisher    = {Association for Computational Linguistics},
  year         = {2025},
  url          = {https://aclanthology.org/2025.findings-acl.958/},
  bibsource    = {dblp computer science bibliography, https://dblp.org}
}

@inproceedings{aokvqa,
  author       = {Dustin Schwenk and
                  Apoorv Khandelwal and
                  Christopher Clark and
                  Kenneth Marino and
                  Roozbeh Mottaghi},
  editor       = {Shai Avidan and
                  Gabriel J. Brostow and
                  Moustapha Ciss{\'{e}} and
                  Giovanni Maria Farinella and
                  Tal Hassner},
  title        = {{A-OKVQA:} {A} Benchmark for Visual Question Answering Using World
                  Knowledge},
  booktitle    = {Computer Vision - {ECCV} 2022 - 17th European Conference, Tel Aviv,
                  Israel, October 23-27, 2022, Proceedings, Part {VIII}},
  series       = {Lecture Notes in Computer Science},
  volume       = {13668},
  pages        = {146--162},
  publisher    = {Springer},
  year         = {2022},
  url          = {https://doi.org/10.1007/978-3-031-20074-8\_9},
  doi          = {10.1007/978-3-031-20074-8\_9},
  bibsource    = {dblp computer science bibliography, https://dblp.org}
}

@inproceedings{vqa,
  author       = {Stanislaw Antol and
                  Aishwarya Agrawal and
                  Jiasen Lu and
                  Margaret Mitchell and
                  Dhruv Batra and
                  C. Lawrence Zitnick and
                  Devi Parikh},
  title        = {{VQA:} Visual Question Answering},
  booktitle    = {2015 {IEEE} International Conference on Computer Vision, {ICCV} 2015,
                  Santiago, Chile, December 7-13, 2015},
  pages        = {2425--2433},
  publisher    = {{IEEE} Computer Society},
  year         = {2015},
  url          = {https://doi.org/10.1109/ICCV.2015.279},
  doi          = {10.1109/ICCV.2015.279},
  bibsource    = {dblp computer science bibliography, https://dblp.org}
}

@inproceedings{anet,
  title={ActivityNet: A Large-Scale Video Benchmark for Human Activity Understanding},
  author={Fabian Caba Heilbron, Victor Escorcia, Bernard Ghanem and Juan Carlos Niebles},
  booktitle={Proceedings of the IEEE Conference on Computer Vision and Pattern Recognition},
  pages={961--970},
  year={2015}
}

@inproceedings{plummer2015flickr30k,
  title={Flickr30k entities: Collecting region-to-phrase correspondences for richer image-to-sentence models},
  author={Plummer, Bryan A and Wang, Liwei and Cervantes, Chris M and Caicedo, Juan C and Hockenmaier, Julia and Lazebnik, Svetlana},
  booktitle={Proceedings of the IEEE international conference on computer vision},
  pages={2641--2649},
  year={2015}
}

@inproceedings{lin2014microsoft,
  title={Microsoft coco: Common objects in context},
  author={Lin, Tsung-Yi and Maire, Michael and Belongie, Serge and Hays, James and Perona, Pietro and Ramanan, Deva and Doll{\'a}r, Piotr and Zitnick, C Lawrence},
  booktitle={European conference on computer vision},
  pages={740--755},
  year={2014},
  organization={Springer}
}

@article{llava-pretrain-gen,
  title={Visual instruction tuning},
  author={Liu, Haotian and Li, Chunyuan and Wu, Qingyang and Lee, Yong Jae},
  journal={Advances in neural information processing systems},
  volume={36},
  pages={34892--34916},
  year={2023}
}

@article{sun2023journeydb,
  title={Journeydb: A benchmark for generative image understanding},
  author={Sun, Keqiang and Pan, Junting and Ge, Yuying and Li, Hao and Duan, Haodong and Wu, Xiaoshi and Zhang, Renrui and Zhou, Aojun and Qin, Zipeng and Wang, Yi and others},
  journal={Advances in neural information processing systems},
  volume={36},
  pages={49659--49678},
  year={2023}
}

@article{zhang2025reasongen,
  title={ReasonGen-R1: CoT for Autoregressive Image generation models through SFT and RL},
  author={Zhang, Yu and Li, Yunqi and Yang, Yifan and Wang, Rui and Yang, Yuqing and Qi, Dai and Bao, Jianmin and Chen, Dongdong and Luo, Chong and Qiu, Lili},
  journal={arXiv preprint arXiv:2505.24875},
  year={2025}
}

@article{chen2025opengpt,
  title={OpenGPT-4o-Image: A Comprehensive Dataset for Advanced Image Generation and Editing},
  author={Chen, Zhihong and Bai, Xuehai and Shi, Yang and Fu, Chaoyou and Zhang, Huanyu and Wang, Haotian and Sun, Xiaoyan and Zhang, Zhang and Wang, Liang and Zhang, Yuanxing and others},
  journal={arXiv preprint arXiv:2509.24900},
  year={2025}
}

@article{xing2025scalecap,
  title={ScaleCap: Inference-Time Scalable Image Captioning via Dual-Modality Debiasing},
  author={Xing, Long and Huang, Qidong and Dong, Xiaoyi and Zhang, Pan and Zang, Yuhang and Cao, Yuhang and Li, Jinsong and Ding, Shuangrui and Zhang, Weiming and Yu, Nenghai and others},
  journal={arXiv preprint arXiv:2506.19848},
  year={2025}
}

@article{wang2025textatlas5m,
  title={Textatlas5m: A large-scale dataset for dense text image generation},
  author={Wang, Alex Jinpeng and Mao, Dongxing and Zhang, Jiawei and Han, Weiming and Dong, Zhuobai and Li, Linjie and Lin, Yiqi and Yang, Zhengyuan and Qin, Libo and Zhang, Fuwei and others},
  journal={arXiv preprint arXiv:2502.07870},
  year={2025}
}

@inproceedings{brooks2023instructpix2pix,
  title={Instructpix2pix: Learning to follow image editing instructions},
  author={Brooks, Tim and Holynski, Aleksander and Efros, Alexei A},
  booktitle={Proceedings of the IEEE/CVF conference on computer vision and pattern recognition},
  pages={18392--18402},
  year={2023}
}

@article{zhang2023magicbrush,
  title={Magicbrush: A manually annotated dataset for instruction-guided image editing},
  author={Zhang, Kai and Mo, Lingbo and Chen, Wenhu and Sun, Huan and Su, Yu},
  journal={Advances in Neural Information Processing Systems},
  volume={36},
  pages={31428--31449},
  year={2023}
}

@article{hui2024hq,
  title={Hq-edit: A high-quality dataset for instruction-based image editing},
  author={Hui, Mude and Yang, Siwei and Zhao, Bingchen and Shi, Yichun and Wang, Heng and Wang, Peng and Zhou, Yuyin and Xie, Cihang},
  journal={arXiv preprint arXiv:2404.09990},
  year={2024}
}

@article{zhao2024ultraedit,
  title={Ultraedit: Instruction-based fine-grained image editing at scale},
  author={Zhao, Haozhe and Ma, Xiaojian Shawn and Chen, Liang and Si, Shuzheng and Wu, Rujie and An, Kaikai and Yu, Peiyu and Zhang, Minjia and Li, Qing and Chang, Baobao},
  journal={Advances in Neural Information Processing Systems},
  volume={37},
  pages={3058--3093},
  year={2024}
}

@article{kuprashevich2025nohumansrequired,
  title={Nohumansrequired: Autonomous high-quality image editing triplet mining},
  author={Kuprashevich, Maksim and Alekseenko, Grigorii and Tolstykh, Irina and Fedorov, Georgii and Suleimanov, Bulat and Dokholyan, Vladimir and Gordeev, Aleksandr},
  journal={arXiv preprint arXiv:2507.14119},
  year={2025}
}

@inproceedings{li2024controlnet++,
  title={Controlnet++: Improving conditional controls with efficient consistency feedback: Project page: liming-ai. github. io/controlnet\_plus\_plus},
  author={Li, Ming and Yang, Taojiannan and Kuang, Huafeng and Wu, Jie and Wang, Zhaoning and Xiao, Xuefeng and Chen, Chen},
  booktitle={European Conference on Computer Vision},
  pages={129--147},
  year={2024},
  organization={Springer}
}

@inproceedings{ancuti2019dense,
  title={Dense-haze: A benchmark for image dehazing with dense-haze and haze-free images},
  author={Ancuti, Codruta O and Ancuti, Cosmin and Sbert, Mateu and Timofte, Radu},
  booktitle={2019 IEEE international conference on image processing (ICIP)},
  pages={1014--1018},
  year={2019},
  organization={IEEE}
}

@inproceedings{ancuti2020nh,
  title={NH-HAZE: An image dehazing benchmark with non-homogeneous hazy and haze-free images},
  author={Ancuti, Codruta O and Ancuti, Cosmin and Timofte, Radu},
  booktitle={Proceedings of the IEEE/CVF conference on computer vision and pattern recognition workshops},
  pages={444--445},
  year={2020}
}

@article{li2018benchmarking,
  title={Benchmarking single-image dehazing and beyond},
  author={Li, Boyi and Ren, Wenqi and Fu, Dengpan and Tao, Dacheng and Feng, Dan and Zeng, Wenjun and Wang, Zhangyang},
  journal={IEEE transactions on image processing},
  volume={28},
  number={1},
  pages={492--505},
  year={2018},
  publisher={IEEE}
}

@inproceedings{zhang2023weatherstream,
  title={Weatherstream: Light transport automation of single image deweathering},
  author={Zhang, Howard and Ba, Yunhao and Yang, Ethan and Mehra, Varan and Gella, Blake and Suzuki, Akira and Pfahnl, Arnold and Chandrappa, Chethan Chinder and Wong, Alex and Kadambi, Achuta},
  booktitle={Proceedings of the IEEE/CVF conference on computer vision and pattern recognition},
  pages={13499--13509},
  year={2023}
}

@inproceedings{li2019heavy,
  title={Heavy rain image restoration: Integrating physics model and conditional adversarial learning},
  author={Li, Ruoteng and Cheong, Loong-Fah and Tan, Robby T},
  booktitle={Proceedings of the IEEE/CVF conference on computer vision and pattern recognition},
  pages={1633--1642},
  year={2019}
}

@inproceedings{fu2017removing,
  title={Removing rain from single images via a deep detail network},
  author={Fu, Xueyang and Huang, Jiabin and Zeng, Delu and Huang, Yue and Ding, Xinghao and Paisley, John},
  booktitle={Proceedings of the IEEE conference on computer vision and pattern recognition},
  pages={3855--3863},
  year={2017}
}

@inproceedings{quan2021removing,
  title={Removing raindrops and rain streaks in one go},
  author={Quan, Ruijie and Yu, Xin and Liang, Yuanzhi and Yang, Yi},
  booktitle={Proceedings of the IEEE/CVF conference on computer vision and pattern recognition},
  pages={9147--9156},
  year={2021}
}

@inproceedings{qian2018attentive,
  title={Attentive generative adversarial network for raindrop removal from a single image},
  author={Qian, Rui and Tan, Robby T and Yang, Wenhan and Su, Jiajun and Liu, Jiaying},
  booktitle={Proceedings of the IEEE conference on computer vision and pattern recognition},
  pages={2482--2491},
  year={2018}
}

@inproceedings{zhu2023learning,
  title={Learning weather-general and weather-specific features for image restoration under multiple adverse weather conditions},
  author={Zhu, Yurui and Wang, Tianyu and Fu, Xueyang and Yang, Xuanyu and Guo, Xin and Dai, Jifeng and Qiao, Yu and Hu, Xiaowei},
  booktitle={Proceedings of the IEEE/CVF conference on computer vision and pattern recognition},
  pages={21747--21758},
  year={2023}
}

@article{liu2018desnownet,
  title={Desnownet: Context-aware deep network for snow removal},
  author={Liu, Yun-Fu and Jaw, Da-Wei and Huang, Shih-Chia and Hwang, Jenq-Neng},
  journal={IEEE Transactions on Image Processing},
  volume={27},
  number={6},
  pages={3064--3073},
  year={2018},
  publisher={IEEE}
}

@article{yang2021sparse,
  title={Sparse gradient regularized deep retinex network for robust low-light image enhancement},
  author={Yang, Wenhan and Wang, Wenjing and Huang, Haofeng and Wang, Shiqi and Liu, Jiaying},
  journal={IEEE Transactions on Image Processing},
  volume={30},
  pages={2072--2086},
  year={2021},
  publisher={IEEE}
}

@inproceedings{liu2021wdnet,
  title={Wdnet: Watermark-decomposition network for visible watermark removal},
  author={Liu, Yang and Zhu, Zhen and Bai, Xiang},
  booktitle={Proceedings of the IEEE/CVF winter conference on applications of computer vision},
  pages={3685--3693},
  year={2021}
}

@inproceedings{agustsson2017ntire,
  title={Ntire 2017 challenge on single image super-resolution: Dataset and study},
  author={Agustsson, Eirikur and Timofte, Radu},
  booktitle={Proceedings of the IEEE conference on computer vision and pattern recognition workshops},
  pages={126--135},
  year={2017}
}

@inproceedings{wang2023ntire,
  title={NTIRE 2023 challenge on stereo image super-resolution: Methods and results},
  author={Wang, Longguang and Guo, Yulan and Wang, Yingqian and Li, Juncheng and Gu, Shuhang and Timofte, Radu and Cheng, Ming and Ma, Haoyu and Ma, Qiufang and Sun, Xiaopeng and others},
  booktitle={Proceedings of the IEEE/CVF conference on computer vision and pattern recognition},
  pages={1346--1372},
  year={2023}
}

@inproceedings{nah2017deep,
  title={Deep multi-scale convolutional neural network for dynamic scene deblurring},
  author={Nah, Seungjun and Hyun Kim, Tae and Mu Lee, Kyoung},
  booktitle={Proceedings of the IEEE conference on computer vision and pattern recognition},
  pages={3883--3891},
  year={2017}
}

@inproceedings{rim2020real,
  title={Real-world blur dataset for learning and benchmarking deblurring algorithms},
  author={Rim, Jaesung and Lee, Haeyun and Won, Jucheol and Cho, Sunghyun},
  booktitle={European conference on computer vision},
  pages={184--201},
  year={2020},
  organization={Springer}
}

@inproceedings{cai2019toward,
  title={Toward real-world single image super-resolution: A new benchmark and a new model},
  author={Cai, Jianrui and Zeng, Hui and Yong, Hongwei and Cao, Zisheng and Zhang, Lei},
  booktitle={Proceedings of the IEEE/CVF international conference on computer vision},
  pages={3086--3095},
  year={2019}
}

@inproceedings{abuolaim2020defocus,
  title={Defocus deblurring using dual-pixel data},
  author={Abuolaim, Abdullah and Brown, Michael S},
  booktitle={European conference on computer vision},
  pages={111--126},
  year={2020},
  organization={Springer}
}

@inproceedings{abdelhamed2018high,
  title={A high-quality denoising dataset for smartphone cameras},
  author={Abdelhamed, Abdelrahman and Lin, Stephen and Brown, Michael S},
  booktitle={Proceedings of the IEEE conference on computer vision and pattern recognition},
  pages={1692--1700},
  year={2018}
}

@article{niu2025wise,
  title={Wise: A world knowledge-informed semantic evaluation for text-to-image generation},
  author={Niu, Yuwei and Ning, Munan and Zheng, Mengren and Jin, Weiyang and Lin, Bin and Jin, Peng and Liao, Jiaqi and Feng, Chaoran and Ning, Kunpeng and Zhu, Bin and others},
  journal={arXiv preprint arXiv:2503.07265},
  year={2025}
}

@article{liu2025step1x,
  title={Step1x-edit: A practical framework for general image editing},
  author={Liu, Shiyu and Han, Yucheng and Xing, Peng and Yin, Fukun and Wang, Rui and Cheng, Wei and Liao, Jiaqi and Wang, Yingming and Fu, Honghao and Han, Chunrui and others},
  journal={arXiv preprint arXiv:2504.17761},
  year={2025}
}

@inproceedings{sheynin2024emu,
  title={Emu edit: Precise image editing via recognition and generation tasks},
  author={Sheynin, Shelly and Polyak, Adam and Singer, Uriel and Kirstain, Yuval and Zohar, Amit and Ashual, Oron and Parikh, Devi and Taigman, Yaniv},
  booktitle={Proceedings of the IEEE/CVF Conference on Computer Vision and Pattern Recognition},
  pages={8871--8879},
  year={2024}
}

@inproceedings{yang2017deep,
  title={Deep joint rain detection and removal from a single image},
  author={Yang, Wenhan and Tan, Robby T and Feng, Jiashi and Liu, Jiaying and Guo, Zongming and Yan, Shuicheng},
  booktitle={Proceedings of the IEEE conference on computer vision and pattern recognition},
  pages={1357--1366},
  year={2017}
}

@article{DBLP:journals/corr/abs-2503-22941,
  author       = {Yugen Sato and
                  Tomohiro Takagi},
  title        = {Identifying Multi-modal Knowledge Neurons in Pretrained Transformers
                  via Two-stage Filtering},
  journal      = {CoRR},
  volume       = {abs/2503.22941},
  year         = {2025},
  url          = {https://doi.org/10.48550/arXiv.2503.22941},
  doi          = {10.48550/ARXIV.2503.22941},
  eprinttype    = {arXiv},
  eprint       = {2503.22941},
  bibsource    = {dblp computer science bibliography, https://dblp.org}
}

\newpage

\appendix
\section{Appendix}
\label{sec:case_study}

\subsection{Training Data List}

% \paragraph{Open-source image data}

All open-source training data are presented in Tables \ref{tab:image_data}-\ref{tab:text_data}.

\begin{table}[t]
    \centering
    \small
    \renewcommand{\arraystretch}{1.}
    \begin{tabular}{lc}
    \toprule
    Stage & Dataset \\
    \midrule
    \multirow{5}{*}{\textit{Pretrain}} 
        & PixelProse-RedCaps~\citep{pixelprose} \\
        & PixelProse-CommonPool~\citep{pixelprose} \\
        & GRIT~\citep{grit} \\
        & CC3M~\citep{cc3m} \\
        & LLaVA-Pretrain~\citep{llava-pretrain-gen} \\
    \midrule
    \multirow{13}{*}{\textit{\makecell{SFT \\ \& \\ Annealing}}} 
        & Cambrian-10M~\citep{cambrian-10m} \\
        & LLaVA-OneVision~\citep{Llava-onevision} \\
        & Docmatix~\citep{docmatix} \\
        & Pixmo-Docx~\citep{pixmo} \\
        & Pixmo-Points~\citep{pixmo} \\
        & Pixmo-Point-Explanations~\citep{pixmo} \\
        & Pixmo-Ask-Model-Anything~\citep{pixmo} \\
        & Latex-OCR \\
        & Latex-Formulas \\
        & Arxiv-OCR-v0.2 \\
        & MMK12~\citep{mmk12} \\
        & V*~\citep{v*} \\
        & Vision-R1-cold~\citep{vision-r1}  \\
        & Multimodal-Cold-Start~\citep{mmcoldstart} \\
    \bottomrule
    \end{tabular}
    \caption{\centering{The list of image data used during our training.}}
    \label{tab:image_data}
\end{table}

% \paragraph{Open-source video data}

\begin{table}[t]
    \centering
    \small
    \renewcommand{\arraystretch}{1.}
    \begin{tabular}{lc}
    \toprule
    Stage & Dataset \\
    \midrule
    \multirow{3}{*}{\textit{Pretrain}} 
        & Valley-Pretrain~\citep{valley} \\
        & ShareGPT4Video~\citep{chen2024sharegpt4video} \\
        & VideoVista-Event~\citep{li2024videovista} \\
    \midrule
    \multirow{17}{*}{\textit{\makecell{SFT \\ \& \\ Annealing}}} 
        & VideoChat2-IT*~\citep{mvbench} \\
        & LLaVA-Video-178K~\citep{llava-video-178k} \\
        & VideoVista-Train~\citep{li2024videovista} \\
        & VideoGPT-Plus~\citep{videoGPTplus} \\
        & fineVideo~\citep{FineVideo} \\
        & TimeChat-Online~\citep{timechatonline} \\
        & Charades-STA~\citep{charades-sta} \\
        & CinePile~\citep{cinepile} \\
        & SF20K~\citep{sf20k} \\
        & Neptune~\citep{neptune24} \\
        & EgoTaskQA~\citep{jia2022egotaskqa} \\
        & FunQA~\citep{xie2024funqa} \\
        & Vript~\citep{yang2024vript} \\
        & Tarsier2-Recap~\citep{tarsier2} \\
        & InternVideo2-Vid-Text*~\citep{wang2024internvideo2} \\
        & SR-91K*~\citep{ouyang2025sr_91k} \\
        & VideoVista2-CoT~\citep{li2024videovista} \\
    \bottomrule
    \end{tabular}
    \caption{\centering{The list of video data used during our training. * indicates that only a subset of the dataset was employed.}}
    \label{tab:video_data}
\end{table}

% \paragraph{Open-source audio data}

\begin{table}[t]
    \centering
    \small
    \renewcommand{\arraystretch}{1.}
    \begin{tabular}{lc}
    \toprule
    Stage & Dataset \\
    \midrule
    \multirow{8}{*}{\textit{Pretrain}} 
        & Multilingual-LibriSpeech-English~\citep{mls} \\
        & GigaSpeech-L~\citep{giga} \\
        & CommonVoice-English~\citep{commonvoice} \\
        & WavCaps~\citep{wavcaps} \\
        & ClothoV1~\citep{clotho} \\
        & MELD~\citep{meld} \\
        & MusicBench~\citep{musicbench} \\
        & LP-MusicCaps~\citep{lpmusic} \\
    \midrule
    \multirow{34}{*}{\textit{\makecell{SFT \\ \& \\ Annealing}}} 
        & CommonVoice-English~\citep{commonvoice} \\
        & WavCaps~\citep{wavcaps} \\
        & ClothoV1~\citep{clotho} \\
        & MELD~\citep{meld} \\
        & MusicBench~\citep{musicbench} \\
        & LP-MusicCaps~\citep{lpmusic} \\
        & ClothoAQA~\citep{clothoaqa} \\
        & AudioCaps~\citep{audiocaps} \\
        & MELD~\citep{meld} \\
        & ASVP-ESD~\citep{asvp_esd} \\
        & CREMA-D~\citep{crema_d} \\
        & EMOV~\citep{emov} \\
        & ESD~\citep{esd} \\
        & JL-Coprus~\citep{jlcorpus} \\
        & RAVDESS~\citep{ravdess} \\   
        & MusicInstruct~\citep{MusiLingo} \\
        & LibriSpeech-Long~\citep{librispeech} \\
        & RACE-Audio~\citep{race} \\
        & Aishell1~\citep{aishell1} \\
        & Aishell3~\citep{aishell3} \\
        & Mozilla-CommonVoice17~\citep{commonvoice} \\
        & Peoples-Speech~\citep{people} \\
        & GigaSpeech-M~\citep{giga} \\
        & LibriSpeech~\citep{librispeech} \\
        & AVQA~\citep{avqa} \\
        & Music-AVQA~\citep{musicavqa} \\
        & ClothoV2~\citep{clotho} \\
        & Ultra-Chat-Audio~\citep{ultra} \\
        & Belle-Audio~\citep{belle} \\
        & Openhermes-Audio~\citep{OpenHermes} \\
        & moss-Audio~\citep{moss} \\
        & alpaca-Audio~\citep{alpaca} \\
        & Llava-150k-Audio~\citep{llava-pretrain-gen} \\
        & Pixmo-Audio~\citep{pixmo} \\
        & Llava-video-180k-Audio~\citep{llavavid} \\
        & TinyStories-en-Audio~\citep{tinystories} \\
        & TinyStories-zh-Audio~\citep{tinystories} \\
        & VCCM~\citep{vccm} \\
        & SpeechCraft~\citep{speechcraft} \\
        & Stream-Omni-Instruct-Audio~\citep{stream} \\
        & Voice-Assistant-Audio~\citep{miniomni} \\
    \bottomrule
    \end{tabular}
    \caption{\centering{The list of audio data used during our training.}}
    \label{tab:audio_data}
\end{table}

% \paragraph{Open-source text data}

\begin{table}[t]
    \centering
    \small
    \renewcommand{\arraystretch}{1.}
    \begin{tabular}{lc}
    \toprule
    Stage & Dataset \\
    \midrule
    \multirow{5}{*}{\textit{\makecell{SFT \\ \& \\ Annealing}}} 
        & OpenOrca-GPT4~\citep{OpenOrca} \\
        & OpenOrca-Chinese-GPT4~\citep{OpenOrca} \\
        & DAPO-Math~\citep{yu2025dapo} \\
        & Nemotron-Post-Training-Dataset-v1~\citep{Nemotron} \\
        & Mixture-of-Thoughts~\citep{Mixture-of-Thoughts} \\
        
    \bottomrule
    \end{tabular}
    \caption{\centering{The list of video data used during our training.}}
    \label{tab:text_data}
\end{table}

% \paragraph{Open-source image generation and eidition data}

\begin{table}[t]
    \centering
    \small
    \renewcommand{\arraystretch}{1.}
    \begin{tabular}{lc}
    \toprule
    Task & Dataset \\
    \midrule
    \multirow{7}{*}{\textit{\makecell{Image Generation \\ (4.81M)}}} 
        & Flickr30k~\citep{plummer2015flickr30k} \\
        & Coco*~\citep{lin2014microsoft} \\
        & LLaVA-Pretrain~\citep{llava-pretrain-gen} \\
        & JourneyDB*~\citep{sun2023journeydb} \\
        & ReasonGen~\citep{zhang2025reasongen} \\
        & OpenGPT-4o-Image~\citep{chen2025opengpt} \\
        & ScaleCap~\citep{xing2025scalecap} \\
        & TextAtlas5M*~\citep{wang2025textatlas5m} \\
    \midrule
    \multirow{6}{*}{\textit{\makecell{Image Edition \\ (5.68M)}}} 
        & InstructPix2pix*~\citep{brooks2023instructpix2pix} \\
        & MagicBrush~\citep{zhang2023magicbrush} \\
        & HQ-Edit~\citep{hui2024hq} \\
        & UltraEdit*~\citep{zhao2024ultraedit} \\
        & OpenGPT-4o-Image~\citep{chen2025opengpt} \\
        & NHR-Edit~\citep{kuprashevich2025nohumansrequired} \\
    \midrule
    \multirow{1}{*}{\textit{\makecell{Controllable Generation \\ (0.50M)}}} 
        & MultiGen-20M*~\citep{li2024controlnet++} \\
    \\
    \midrule
    \multirow{19}{*}{\textit{\makecell{Image Restoration \\ (0.36M)}}} 
        & DenseHaze~\citep{ancuti2019dense} \\
        & NH-HAZE~\citep{ancuti2020nh} \\
        & Reside-6K~\citep{li2018benchmarking} \\
        & Weather-Stream~\citep{zhang2023weatherstream} \\
        & Outdoor-Rain~\citep{li2019heavy} \\
        & Rain1400~\citep{fu2017removing} \\
        & RainDS~\citep{quan2021removing} \\
        & RainDrop~\citep{qian2018attentive} \\
        & RealSnow~\citep{zhu2023learning} \\
        & Snow100K~\citep{liu2018desnownet} \\
        & LOL-v2~\citep{yang2021sparse} \\
        & CLWD~\citep{liu2021wdnet} \\
        & DIV2K~\citep{agustsson2017ntire} \\
        & Flickr2K~\citep{wang2023ntire} \\
        & GoPro~\citep{nah2017deep} \\
        & RealBlur~\citep{rim2020real} \\
        & RealSR~\citep{cai2019toward} \\
        & DPDD~\citep{abuolaim2020defocus} \\
        & SIDD~\citep{abdelhamed2018high} \\
    \bottomrule
    \end{tabular}
    \caption{\centering{The list of visual generation data used during our training. An asterisk (*) indicates that only a subset of the dataset was employed.}}
    \label{tab:image_gen_data}
\end{table}

\begin{table}[t]
    \centering
    \small
    \renewcommand{\arraystretch}{1.}
    \begin{tabular}{lc}
    \toprule
    Stage & Dataset \\
    \midrule
    \multirow{4}{*}{\textit{\makecell{Cold \\ Start}}} 
    & Mixture-of-Thoughts~\citep{Mixture-of-Thoughts} \\
    & Vision-R1-cold~\citep{vision-r1} \\
    & Multimodal-Cold-Start~\citep{mmcoldstart} \\
    & VideoVista-2-LongCoT~\citep{chen2025videovista_cultural} \\
    \midrule
    \multirow{1}{*}{\textit{GSPO}} 
    & MMPR-Tiny*~\citep{wang2025internvl3}  \\
    \bottomrule
    \end{tabular}
    \caption{\centering{The list of video data used during our training. * indicates that only a subset of the dataset was employed.}}
    \label{tab:rl_data}
\end{table}

\subsection{Evaluation Data List}

\paragraph{Image Understanding}

We evaluate our models across three categories of capabilities of image understanding: General Visual Understanding, STEM Image Understanding (focusing mainly on Science and Mathematics), and OCR \& Document Understanding.

\textbf{General Visual Understanding}. MMBench (EN/CN)~\citep{liu2024mmbench}, MMStar~\citep{mmstar}, RealWorldQA~\citep{real_world_qa}, GQA~\citep{gqa}, MME-RealWorld~\citep{mme-realworld}, and CV-Bench~\citep{cvbench}.

\textbf{STEM Image Reasoning}. AI2D~\citep{ai2d}, MMMU~\citep{mmmu} and MMMU-Pro~\citep{yue2024mmmupro} for science reasoning. MathVista~\citep{mathvista}, MathVision~\citep{mathvision} and LogicVista~\citep{logicvista} for mathematics reasoning.

\textbf{OCR \& Document Understanding}. DocVQA~\citep{docvqa}, ChartQA~\citep{chartqa}, CharXiv (DQ/RQ)~\citep{charxivdq}, and SEED-Bench-2-Plus~\citep{seed2plus}. 

\paragraph{Video Understanding}

We evaluate our models across four categories of capabilities of video understanding: Short Video Understanding, Long Video Understanding, Video Reasoning and Video Temporal Localization.

\textbf{Short Video Understanding}. MVBench~\citep{mvbench}.

\textbf{Long Video Understanding}. Video-MME~\citep{video_mme}, LongVideoBench~\citep{longvideobench} and EgoSchema~\citep{egoschema}.

\textbf{Video Reasoning}. VideoMMMU~\citep{video_mmmu} for video knowledge reasoning, VSI-Bench~\citep{vsi-bench} for video spatial reasoning and TOMATO~\citep{tomato} for video temporal reasoning.

\textbf{Video Temporal Localization}. Charades-STA~\citep{charades-sta}.

\paragraph{Language Capability}

We evaluate our models on two benchmarks focused on complex reasoning and knowledge tasks, including the three versions of GPQA~\citep{gpqa} and the MMLU-Pro~\citep{mmlupro} dataset.

\paragraph{Omni Understanding}

We evaluate our models on two categories of omni benchmarks that require simultaneous understanding of visual and audio information: Video\&Audio and Image\&Audio.

\textbf{Video\&Audio}. WorldSense~\citep{hong2025worldsense}, OmniVideoBench~\citep{omnivideobench} and StreamingBench~\citep{streamingbench}.

\textbf{Image\&Audio}. OmniBench~\citep{omnibench}

\paragraph{Visual Generation}

We evaluate our models across four categories of visual generation tasks: Image Generation, Image Edition, Controllable Generation and Image Restoration.

\textbf{Image Generation}. Wise~\citep{niu2025wise} and Coco30K~\citep{lin2014microsoft}.

\textbf{Image Edition}. GEdit-Bench-EN~\citep{liu2025step1x}, Emu Edit Test~\citep{sheynin2024emu} and MagicBrush~\citep{zhang2023magicbrush}.

\textbf{Controllable Generation}. MultiGen~\citep{li2024controlnet++}.

\textbf{Low-Level Image Restoration}. Rain100L~\citep{yang2017deep} and SIDD~\citep{abdelhamed2018high}.

\paragraph{Audio Understanding and Speech Generation}
We evaluate our models across four categories of capabilities of audio understanding and speech generation: Audio understanding with text reply, text to speech, voice conversation(speech to speech or speech to text), and voice conversation with visual Information(speech and image/video to speech or speech and image/video to text.

\textbf{Audio X → Text}
RACE-audio(middle/high)~\citep{race}, EHSL~\citep{li_unimoe}, MELD~\citep{meld}, MMAU~\citep{mmau}, ClothoAQA~\citep{clothoaqa}, ClothoV1~\citep{clotho}, ClothoV2~\citep{clotho}, AudioCaps~\citep{audiocaps}, and MusicCaps~\citep{lpmusic}.

\textbf{Text → Speech}
LibriTTS~\citep{libritts}, SEED~\citep{seedtts}, and TinyStories~\citep{tinystories}.

\textbf{Speech → Speech/Text}
LlamaQA~\citep{llamaq}, WebQA~\citep{webq}, BigBench Audio~\citep{bigbench}, and MultiChallenge Audio~\citep{multichallenge}.

\textbf{Vision + Speech → Speech/Text}
A-OK-VQA~\citep{aokvqa}, VQAv2~\citep{vqa}, and ActivityNet ~\citep{anet}.

\subsection{Thinking Prompt}

\begin{table}[ht]
\renewcommand{\arraystretch}{0.5}
  \label{tab:prompt_setting}
  \centering
  \begin{tabular}{p{12cm}}
    \toprule
    Prompt For Visual Understanding \\
    \midrule
    \textbf{SYSTEM:} You are Uni-MoE-v2, a helpful multi-modal model. Your role as an assistant involves thoroughly exploring questions through a systematic thinking process before providing the final precise and accurate solutions. This requires engaging in a comprehensive cycle of analysis, summarizing, exploration, reassessment, reflection, backtracing, and iteration to develop well-considered thinking process. Please structure your response into two main sections: Thought and Solution using the specified format: <think> Thought section </think> Solution section. In the Thought section, detail your reasoning process in steps. Each step should include detailed considerations such as analyzing questions, summarizing relevant findings, brainstorming new ideas, verifying the accuracy of the current steps, refining any errors, and revisiting previous steps. In the Solution section, based on various attempts, explorations, and reflections from the Thought section, systematically present the final solution that you deem correct. The Solution section should be logical, accurate, and concise and detail necessary steps needed to reach the conclusion. Now, try to solve the following question through the above guidelines.\\
    \textbf{USER:} Question: \{question\}\\
    \midrule
    Prompt For Visual Generation \\
    \midrule
    \textbf{SYSTEM:} You should first think step by step about how to construct the image, including background, objects, colors, lighting, and style. The reasoning process and answer are enclosed within <think> </think> and <answer> </answer> tags, respectively, i.e., <think>1. Position a man seated indoors, capturing him from the upper chest to just above the chin, focusing on ... </think> <answer> Here is the image: [TASK0][TASK1][TASK2][IMG0][IMG1]......[IMG31] </answer>, which means your output should start with <think> and end with </answer>.\\
    \textbf{USER:} Image generation: \{prompt\}\\
    \bottomrule
  \end{tabular}
  \caption{The prompt setting for thinking models.}
  % \vspace{-16pt}
% \vspace{-12pt}
\end{table}

\subsection{Gradient Estimation Formalization}

\label{sec:gradient_estimation_appendix}

For clarity of exposition, we restrict our discussion to the Top-1 MoE setting, and later describe how the approach can be extended to our Dynamic-Capacity MoE. We first consider the Top-1 MoE layer whose output is given by:
\begin{align}
    \sum_{i=0}^{n-1} \texttt{Softmax}(\vz)_i \cdot \mD_i \cdot Expert(\vx, \vw_i), \mbox{ where } \mD \sim \texttt{Softmax}(\vz). 
    \label{eqn:mixer-layer}
\end{align}
Here, $\vz$ denotes the router logits, $\texttt{Softmax}(\vz)_i$ is the gating probability for expert $i$, $\mD_i$ is a binary mask indicating whether expert $i$ is selected.
% Referring to other parts of the network (including the loss function) as $f$, the training objective is:
Let $f(\cdot)$ denote the remainder of the network, including the loss function. The training objective can then be expressed as:
\begin{align}
   \mathcal{L} 
   &= E_{\mD \sim \texttt{Softmax}(\vz)} \left[f\left(\sum_{i=0}^{n-1} \texttt{Softmax}(\vz)_i \cdot \mD_i \cdot Expert(\vx, \vw_i)\right) \right]  \nonumber\\
   &= \sum_{i=0}^{n-1} f\big(\texttt{Softmax}(\vz)_i \cdot Expert(\vx, \vw_i)\big) \cdot \texttt{Softmax}(\vz)_i. \label{eqn:object}
\end{align}
Equation~\ref{eqn:object} rewrites the expectation over $\mD$ as a weighted sum over experts, where each term is the loss contribution from a single expert multiplied by its routing probability.
% For simplicity, we mark $\texttt{Softmax}(\vz)$ as $\vp$. 
% Then the gradient of $\vz$ is:
For notational simplicity, we denote $\vp = \texttt{Softmax}(\vz)$. The gradient of $\mathcal{L}$ with respect to $\vz$ can be written as:
\begin{align}
    \nabla \vz &= \sum_{i=0}^{n-1} \vp_i \cdot \frac{\partial f\big(\vp_i \cdot Expert(\vx, \vw_i)\big)}{\partial \vz} + f\big(\vp_i \cdot Expert(\vx, \vw_i)\big) \cdot \frac{\partial \vp_i}{\partial \vz} \nonumber \\
    &= \sum_{i=0}^{n-1} \vp_i \cdot \frac{\partial f\big(\vp_i \cdot Expert(\vx, \vw_i)\big)}{\partial \vz} + \Big( f\big(\vp_i \cdot Expert(\vx, \vw_i)\big) - f(0) \Big) \cdot \frac{\partial \vp_i}{\partial \vz}. \label{eqn:baseline-subtract}
\end{align}
The second equality in Equation~\ref{eqn:baseline-subtract} is usually known as baseline subtraction.
% Note that Equation~\ref{eqn:baseline-subtract} is usually known as baseline subtraction.
% In the ODE literature, there are many ways to approximate $f\big(\vp_i \cdot Expert(\vx, \vw_i)\big) - f(0)$. 
% Here, we focus on two approximations:
In the ODE literature, the term $f\big(\vp_i \cdot Expert(\vx, \vw_i)\big) - f(0)$ can be approximated in various ways. We focus on two common numerical schemes:
\begin{itemize}
\item  Euler's method: a first-order ODE solver that approximates $f\big(\vp_i \cdot Expert(\vx, \vw_i)\big) - f(0)$ as $f'\big(\vp_i \cdot Expert(\vx, \vw_i)\big) \cdot \vp_i \cdot Expert(\vx, \vw_i)$.

\item Heun's third-order method: a third-order ODE solver that approximates $f\big(\vp_i \cdot Expert(\vx, \vw_i)\big) - f(0)$ as {\footnotesize $\Big(\frac{1}{4}\cdot f'\big(\vp_i \cdot Expert(\vx, \vw_i)\big) + \frac{3}{4} \cdot f'(\frac{\vp_i \cdot Expert(\vx, \vw_i)}{3})\Big)\cdot \vp_i \cdot Expert(\vx, \vw_i)$}.
\end{itemize}

We next present two alternative approximations of $\nabla_{\vz}$ based on two numerical schemes.

\textbf{First-order (Euler) approximation.}  
Applying Euler's method, the gradient can be expressed as:
\begin{align*}
    \nabla_{\mbox{\scriptsize 1st }} \vz &= \sum_{i=0}^{n-1}  \left( \vp_i \cdot \frac{\partial f\big(\vp_i \cdot Expert(\vx, \vw_i)\big)}{\partial \vz} + f'\big(\vp_i Expert(\vx, \vw_i)\big) \cdot \vp_i Expert(\vx, \vw_i) \cdot \frac{\partial \vp_i}{\partial \vz} \right)  \\
    &= \sum_{i=0}^{n-1} \left( \vp_i \cdot \frac{\partial f\big(\vp_i \cdot Expert(\vx, \vw_i)\big)}{\partial \vz} + \vp_i \cdot \frac{\partial f\big(\vp_i Expert(\vx, \vw_i)\big)}{\partial \vp_i Expert(\vx, \vw_i)} \frac{ \partial \vp_i Expert(\vx, \vw_i)}{\partial \vp_i}  \frac{\partial \vp_i}{\partial \vz} \right)\\
    &= \sum_{i=0}^{n-1} 2 \cdot \vp_i \cdot \frac{\partial f\big(\vp_i \cdot Expert(\vx, \vw_i)\big)}{\partial \vz} \\
    &= E_{\mD \sim \texttt{Softmax}(\vz)} \left[ 2 \cdot \frac{\partial f\big(\vp_\mD \cdot Expert(\vx, \vw_\mD)\big)}{\partial \vz} \right].
\end{align*}

\textbf{Third-order (Heun) approximation.}  
Using Heun's method, which combines multiple derivative evaluations for higher accuracy, we obtain:
\begin{align*}
\nabla_{\mbox{\scriptsize 3rd }} \vz =& \sum_{i=0}^{n-1} \Biggl(
\vp_i \cdot \frac{\partial f\big(\vp_i \cdot Expert(\vx, \vw_i)\big)}{\partial \vz} + \left( \frac{1}{4}\cdot f'\big(\vp_i \cdot Expert(\vx, \vw_i)\big) + \right. \\
& \left. \frac{3}{4} \cdot f'\big(\frac{\vp_i \cdot Expert(\vx, \vw_i)}{3}\big) \right) \cdot \vp_i \cdot Expert(\vx, \vw_i) \cdot \frac{\partial \vp_i}{\partial \vz} \Biggr)\\
=& \sum_{i=0}^{n-1} \big(\frac{5}{4} \cdot \vp_i \cdot \frac{\partial f\big(\vp_i \cdot Expert(\vx, \vw_i)\big)}{\partial \vz} + \frac{9}{4} \cdot \vp_i \cdot \frac{\partial f(\frac{\vp_i \cdot Expert(\vx, \vw_i)}{3})}{\partial \vz}\big)\\
=& E_{\mD \sim \texttt{Softmax}(\vz), \mB \sim \texttt{Bernoulli}(\frac{5}{8})} \left[ (6 - 4\mB) \cdot  \frac{\partial f\big(\frac{1 + 2\mB}{3}\cdot \vp_\mD \cdot Expert(\vx, \vw_\mD)\big)}{\partial \vz} \right].
\end{align*}

\textbf{Hybrid gradient estimator.}  
To balance the stability of router training with the diversity of expert learning, we combine the two estimators above: the first-order gradient is used when the selected expert corresponds to $\arg\max(\vz)$, while the third-order gradient is applied otherwise.  
Let $\delta_\mD$ denote the indicator $\delta(\mD = \arg\max(\vz))$. The combined estimator can be written as:
% Lastly, \underline{we combine} $\nabla_{\mbox{\scriptsize 3rd }}$ and $\nabla_{\mbox{\scriptsize 1st }}$  to balance router learning and expert learning by using $\nabla_{\mbox{\scriptsize 1st }}$ if $\mD = \arg\max(\vz)$ or $\nabla_{\mbox{\scriptsize 3rd }}$ otherwise.
% Particularly, using $\delta_\mD$ to refers to $\delta(\mD = \arg\max(\vz))$, we have:
{
\scriptsize
\begin{align}
    \nabla \vz  = E_{\mD \sim \texttt{\tiny Softmax}(\vz)} [\nabla_{\mD} \vz], \nonumber
\end{align}
\begin{align}  
& \mbox{where}\,\, \nabla_{\mD} \vz & \nonumber\\
= & E_{\mB \sim \texttt{\tiny Bernoulli}(\frac{5}{8})} \biggl[(1 - \delta_\mD) \cdot  (6 - 4\mB) \cdot  \frac{\partial f(\frac{1 + 2\mB}{3}\cdot \vp_\mD \cdot Expert(\vx, \vw_\mD))}{\partial \vz} \biggr] + \delta_\mD \cdot 2 \cdot \frac{\partial f(\vp_\mD \cdot Expert(\vx, \vw_\mD))}{\partial \vz}  & \nonumber\\
= & E_{\mB \sim \texttt{\tiny Bernoulli}(\frac{5}{8})} \left[ 
        \big(6-4\cdot \max(\mB, \delta_\mD)\big) \frac{\partial f\big(\frac{1 + 2\cdot \max(\mB, \delta_\mD)}{3} \cdot \vp_\mD\cdot Expert(\vx, \vw_\mD)\big)}{\partial \vz}
    \right] & \nonumber \\
= & E_{\mB \sim \texttt{\tiny Bernoulli}(\frac{5}{8})} \left[
    2 \cdot f'\big(\frac{1 + 2 \max(\mB, \delta_\mD)}{3} \cdot  \vp_\mD \cdot Expert(\vx, \vw_\mD)\big)  \frac{\partial \vp_\mD \cdot Expert(\vx, \vw_\mD)}{\partial \vz}
    \right]. &
    \label{eqn:sparsemixer-final}
\end{align}
}

\textbf{Gradient Estimation Function.} Following the hybrid gradient estimation in Equation~\ref{eqn:sparsemixer-final}, the computation can be described as follows. 
First, we compute the forward output of the sampled expert $\mD$ weighted by its routing probability:
\[
\vo = \text{Expert}(\vx, \vw_{\mD}) \cdot p_{\mD}.
\]

We then define an indicator variable to check whether $\mD$ is the highest-probability expert:
\[
\delta_{\mD} =
\begin{cases}
1, & \text{if } \mD = \arg\max(\vz), \\
0, & \text{otherwise}.
\end{cases}
\]

Next, we sample a Bernoulli random variable:
\[
B \sim \text{Bernoulli}\left(\frac{5}{8}\right),
\]
Finally, the hybrid scaling factor applied to $h$ is:
\[
\vo^{est} = 2 \cdot \vo + \texttt{detach}\left( \max\left( \delta_{\mD}, \frac{1 + 2B}{3} \right) \cdot \vo - 2 \cdot \vo \right),
\]
where $\texttt{detach}(\cdot)$ returns a copy of its argument without gradient flow. 
When $\delta_{\mD} = 1$, the scaling factor is fixed at $2$ (first-order approximation); otherwise, it is given by $\frac{1 + 2B}{3}$ (third-order approximation). 

To extend it to our Top-$P$ strategy, we apply the same computation sequentially to each activated expert, sampling without replacement from the routing distribution until the cumulative probability exceeds $P$. 
In this way, the gradient estimation naturally generalizes to multiple experts while remaining consistent with the selection process in Algorithm~\ref{algo:dynamic_moe}, ensuring that the hybrid scaling mechanism is fully compatible with our dynamic-capacity MoE framework.

\end{document}